\documentclass[a4paper,fleqn]{cas-dc}

\usepackage[authoryear]{natbib}
\usepackage{bbm}
\usepackage{bm}
\usepackage{amsfonts}
\usepackage{multirow}
\usepackage{caption}
\usepackage{subcaption}
\usepackage{array}
\usepackage{threeparttable}
\usepackage{newtxtext}
\usepackage{float}
\usepackage{nicefrac}
\usepackage{enumitem}
\usepackage[utf8]{inputenc}
\usepackage{stfloats}

\def\tsc#1{\csdef{#1}{\textsc{\lowercase{#1}}\xspace}}
\tsc{WGM}
\tsc{QE}
\begin{document}
\let\WriteBookmarks\relax
\def\floatpagepagefraction{1}
\def\textpagefraction{.001}
\newcommand{\rulesep}{\unskip\ \vrule\ }

\shorttitle{Generalization in neural networks}    

\shortauthors{C. Rohlfs}  

\title [mode = title]{Generalization in neural networks: a broad survey}  

% \tnotemark[<tnote number>] 



\author{Chris Rohlfs}[orcid=0000-0001-7714-9231]

% \cormark[<corr mark no>]

% \fnmark[<footnote mark no>]

\ead{car2228@columbia.edu}

% \ead[url]{<URL>}

% \credit{<Credit authorship details>}

\affiliation{organization={Columbia University Department of Electrical Engineering},
            addressline={Mudd 1310, 500 West $120^{th}$ Street}, 
            city={New York},
%          citysep={}, % Uncomment if no comma needed between city and postcode
            postcode={10027-6623}, 
            state={NY},
            country={USA}}


\begin{abstract}
This paper reviews concepts, modeling approaches, and recent findings along a spectrum of different levels of abstraction of neural network models including generalization across (1) Samples, (2) Distributions, (3) Domains, (4) Tasks, (5) Modalities, and (6) Scopes. Strategies for (1) sample generalization from training to test data are discussed, with suggestive evidence presented that, at least for the ImageNet dataset, popular classification models show substantial overfitting. An empirical example and perspectives from statistics highlight how models' (2) distribution generalization can benefit from consideration of causal relationships and counterfactual scenarios. Transfer learning approaches and results for (3) domain generalization are summarized, as is the wealth of domain generalization benchmark datasets available. Recent breakthroughs surveyed in (4) task generalization include few-shot meta-learning approaches and the emergence of transformer-based foundation models such as those used for language processing. Studies performing (5) modality generalization are reviewed, including those that integrate image and text data and that apply a biologically-inspired network across olfactory, visual, and auditory modalities. Higher-level (6) scope generalization results are surveyed, including graph-based approaches to represent symbolic knowledge in networks and attribution strategies for improving networks' explainability. Additionally, concepts from neuroscience are discussed on the modular architecture of brains and the steps by which dopamine-driven conditioning leads to abstract thinking.
\end{abstract}

% Keywords

\maketitle

\section{Introduction} \label{generalization introduction}

Generalization is a fundamental objective of deep learning, and recent achievements in the field have expanded the ability of neural network models to consolidate relationships among variables into patterns that apply in other situations. Some such innovations improve the stability and consistency of model performance in a given domain---aims that researchers have emphasized relate directly to a model's ability to generalize \citep{Bousquet2002}---while other work specifically designs models to adapt to different domains or tasks.

This paper provides a broad overview of different forms of generalization in neural networks. Existing surveys describe theoretical bounds on learning generalizable functions \citep{Mohri2018}, ``regularization'' techniques for ensuring stability of model forecasts across samples \citep{Bejani2021, Kukacka2018, Shorten2019, Qian2022, Tian2022} and populations \citep{Liu2022b,Lust2021,Liu2023}, causal learning to identify counterfactual scenarios \citep{Athey2019,Guo2020}, methods such as those to adjust when the function of interest is changing \citep{Bayram2022, Xiang2023, Yuan2022}, to adapt a model to use an existing skill in a new context \citep{Gulrajani2020,Niu2020,Wang2021,Wang2023,Zhou2022,Zhuang2021}, to leverage prior learning to acquire new skills \citep{Li2021,Vandenhende2021,Wang2019,Zhang2022}, the recent emergence of highly versatile transformer-based ``foundation'' models \citep{Bommasani2021,Chaudhari2021}, and the representation and extraction of symbolic knowledge in deep learning \citep{Bader2005,Besold2017,Davis2017,Ji2022,Townsend2020}. Studies in the neuroscience literature also discuss methods of abstraction used by biological brains \citep{Devineni2022,Mansouri2020,Meunier2009}. Due to the breadth of topics covered here, this article is in places a review of reviews. It does not provide a comprehensive outline or comparison of state-of-the-art methodologies for any one of these areas but instead offers a high-level view of the diversity of ideas about generalization and approaches to the important question of how and where a neural network can be applied. Its primary aim is to supplement these detailed surveys with an intuitive and concept-driven description of how these types of generalization relate to one another, what type of work is being done in these areas and in some cases, how well the strategies are working, and potential next steps toward achieving in machines the levels of abstraction accomplished by humans and animals. 

\begin{figure*}
\centering
\includegraphics[width=0.8\textwidth]{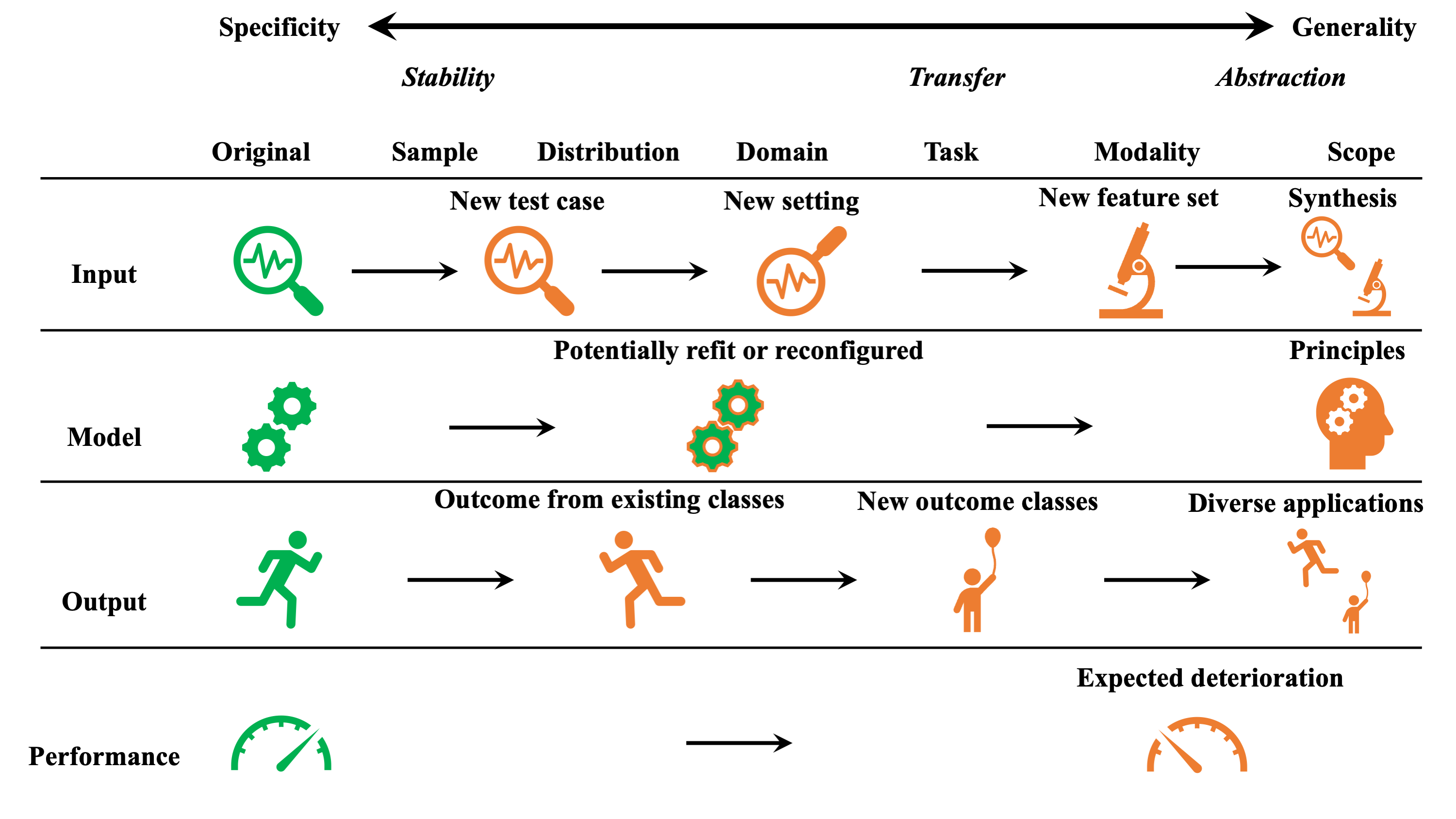}
\caption{Types of Generalization}
\label{fig:generality}
\end{figure*}

Figure \ref{fig:generality} presents a schematic of different ways in which a model can be generalized. Plotted from left to right is the degree of abstraction involved in the generalization of that model. The left-hand side of the diagram, under the heading ``Original,'' corresponds to the deployment of a model on its original training data. Moving to the right, the degree of specificity of the application declines, and the level of generality increases. These forms of generalization describe the application of a model across:
\begin{enumerate}
\item \emph{Samples}: Out-of-sample test cases drawn from the same population as in training
\item \emph{Distributions}: Out-of-distribution test cases drawn from new populations
\item \emph{Domains}: Out-of-domain contexts where the target function differs from that seen in training
\item \emph{Tasks}: New questions or output classes in the same sphere of knowledge with generally similar input data format
\item \emph{Modalities}: Questions involving fundamentally new input data formats or spheres of knowledge 
\item \emph{Scopes}: Complex applications requiring semantic understanding
\end{enumerate}

\noindent Authors have used some of these terms in differing and overlapping ways in the literature. Nevertheless, in the absence of a universally accepted lexicon, the seven categories presented in Figure \ref{fig:generality} offer a concise taxonomy that captures many of the important ways in which generalization is performed. Discussing them in increasing order of abstraction also helps to highlight ways similarities across these different types of generalization problems.

In the top line of the diagram, just below the two-sided arrow, these six types of generalization are consolidated into three broad categories. The first, stability, includes sample and distribution generalization and describes the ability of models to generalize across minor input perturbations. The ground truth target function that drives the input-output relationships is the same as in training. The only changes are that the specific observations were not seen in training and might differ substantially from those that were. The second, knowledge transfer, encompasses generalization across domain, task, and modality. It describes situations in which the model was trained to learn one target function, and it is reconfigured, retrained, or fine-tuned so that it can be applied to learn a new but related input-output relationship that potentially uses drastically different data. The third broad category includes the furthest-reaching forms of generalization and is placed under the heading of abstraction. In this category, the information gleaned from the model is distilled into concepts with potentially wide applicability.

Each of the four rows of icons in Figure \ref{fig:generality} a key element involved in the process of applying a model to a new situation. The icons illustrate the extent to which changes in the level of generality affect the input data, the mathematical tools used for learning and prediction, the outputs that these tools produce, and the degree to which those outputs are accurate. For the baseline labeled ``Original'' in the left-hand column, where no generalization is required, the model was designed and calibrated to maximize the exact metric on which it is being evaluated, and we can expect its level of performance to be high.

Moving to the right, the model is deployed in situations that are increasingly different from this baseline. On the left-hand side of the first row, the input data differ only slightly from training. At greater levels of generality, they eventually take on a completely different form. As the second row shows, the model itself is initially unchanged. At moderate levels of generality, the essential components of the model---what is being generalized---are fixed. While some refitting or reconfiguration may occur, it is the model that is intended to remain stable and to perform with some degree of consistency in the face of changing input feeds and output requirements. At the highest level of abstraction depicted in the diagram, however, the precise model structure is discarded in favor of general principles. As the third row illustrates, the output is expected to differ somewhat as the model is applied to new and different types of observations. At small and moderate levels of generality, these exact predictions may differ, but they are drawn from the same set of possible classes. Higher levels of abstraction involve changes in the set of possible output classes and eventually in the overall structure of the problem. The icons on the bottom row of the diagram indicate that applying models to situations that are increasingly different from training is expected to degrade performance.

Sample generalization, discussed in Section \ref{Sample}, focuses on the well-known problem of overfitting in deep learning, whereby a model predicts outcomes more accurately in training than on out-of-sample test cases. Neural networks' strong performance out-of-sample is compared to effectively uninformative theoretical worst-case guarantees, and regularization strategies for maintaining robust performance out-of-sample are discussed. The accuracies of several popular network-based visual object classifiers are compared between ImageNet-V2 and a separately compiled test set. Suggestive evidence indicates that, while training performance has been steadily improving, the amount of overfitting has been large and stable, consistently accounting for misclassification of 20\% or more of test cases. Data augmentation strategies are proposed as potential solutions, but this large training-test gap in a well-known classification problem is an area for future research.

Next, Section \ref{Distribution} examines generalization to other distributions. It requires not only effectively forecasting out-of-sample but to cases that may look very different from those in training. It is thus a more difficult problem than sample generalization. Distribution generalization is known to cause difficulties for data-driven models, because the new test cases may have rare combinations of features or sensitivities whose impacts were not learned precisely in training. As with sample generalization, theoretical worst-case guarantees on out-of-distribution learnability are mostly uninformative. Techniques from machine learning and from classical statistics and econometrics are discussed for one important form of out-of-distribution generalization---to counterfactual cases for causal inference. As a motivating example, a counterfactual image is presented of a lion on a city block. While the true class of the image is readily apparent to humans, some popular models are found to misclassify it. Effective generalization to counterfactual cases is highlighted as a promising area for future deep learning research that draws considerable interest from applied practitioners. 

Out-of-domain generalization, covered in Section \ref{Domain}, is in a sense an extreme version of out-of-distribution generalization. The model is applied to a new situation with modified rules for mapping input features to output classes---for instance, identifying objects from cartoons rather than photos or classifying documents written in Icelandic rather than English. A cartoon or an Icelandic text would be a rare or impossible case in the training data of photographs or English texts. The relationship between inputs and output classes is distorted so much from that seen in training that it is most practical to model it as its own distinct target function. The section covers two forms of generalization across domains. First, the problem of ``concept drift,'' or a gradual change in domain, is reviewed, along with established strategies for addressing it. Case studies are discussed, including the failure of Google Flu Trends and Completely Automated Public Turing tests to tell Computers and Humans Apart (CAPTCHA). Second, datasets, approaches, and results are discussed for sharp and observable shifts in domain. The Photo-Art-Cartoon-Sketch (PACS) visual classification exercise---in which the network is trained to recognize the same objects in different visual vernaculars (four different domains)---is considered as a motivating example. Domain generalization begins with an established ``foundation model'' that is pretrained on a well-known dataset of images and used as a ``backbone'' and whose parameters are held fixed. New network layers are added to perform the classification task at hand, some of which are domain-specific, and strategies known as \emph{transfer learning} are used to enable the system to classify test cases from a domain that was not seen during training.

Section \ref{Task} considers recent developments in the ability to generalize deep learning models to new tasks with different outcome classes than in the original problem. The distribution of cases and the domain (for instance photographs or text) may be unchanged from training. Thus, unlike in distribution or domain generalization, the processing of data features in task generalization is potentially unchanged. Unlike with distribution or domain generalization, however, no prediction is possible without reconfiguring the model, because the structure of the outputs has changed. Early efforts built task flexibility into the network training process and worked to counteract the phenomenon of \emph{catastrophic forgetting} through which learning a new task would degrade the model's performance on previously tasks. Recent progress in computer vision is discussed that builds upon the transfer learning approach. A pre-trained image classification backbone is extended with additional layers or steps to perform \emph{few-shot learning}. These additional elements enable the model to recognize new classes of images after seeing just one or five example cases. Next, \emph{transformers} are explored in the context of natural language. Networks with this innovative architecture are pre-trained on extensive corpora of text-based data. Like the backbones of image classifiers, these pre-trained transformers serve as foundation models for a variety of tasks. Models built on this common structure are found to perform at or above human level in batteries of language tests. The recent rise of highly versatile transformer-based foundation models is briefly discussed.

In cross-modality generalization, patterns learned from one form of data---for instance, images, text, or audio---are applied to make inferences relevant to another of those areas. Section \ref{Modality} reviews some recent innovative studies in the literature that generalize models across widely different problems and data types. As with task generalization, it is necessary in modality generalization to reconfigure the model in order for it to produce forecasts. Generalization across modalities can be thought of as an extreme change in domain, so that not only has the target function changed, but there has been a complete change in the structure of the input data, the output format, or both. One area in which cross-modality generalization has been particularly influential is the intersection of vision and language. Image classifiers trained on caption-enhanced images are found to identify objects more effectively than those trained on images alone.

The most abstract form of generalization, covered in Section \ref{Scope}, is across scopes. It is less about running a new model and more about the interaction between machine and human logic. Through it, models learn and communicate semantic knowledge and general principles, including those that might be represented through decision trees or symbolic logic. This section examines how human knowledge is codified into graphs and incorporated into network-based architectures. An overview is presented of competitions that evaluate the degree to which the resulting graph-enhanced neural networks can reason through open-ended questions. The topic of knowledge extraction is also covered, with example cases illustrating ways of analyzing the reasoning behind deep learning systems predictions. Both forms of generalization---the representation of semantic knowledge in neural networks and the distillation of learned knowledge into understandable principles---are promising areas for further exploration.

Next, Section \ref{Nature} reviews neuroscience research on the modular and dopamine-driven means by which abstraction occurs in animal brains. This biological perspective provides insights into modular elements of existing artificial neural networks as well as potential directions in which artificial systems might generalize more effectively.

Section \ref{Conclusion} summarizes, discusses potential future directions for network-based abstraction, and concludes.

\section{Sample Generalization} \label{Sample}

\subsection{Bias and Variance} \label{Bias}

\begin{figure}
  \centering
  % include first image
  \includegraphics[width=0.45\textwidth]{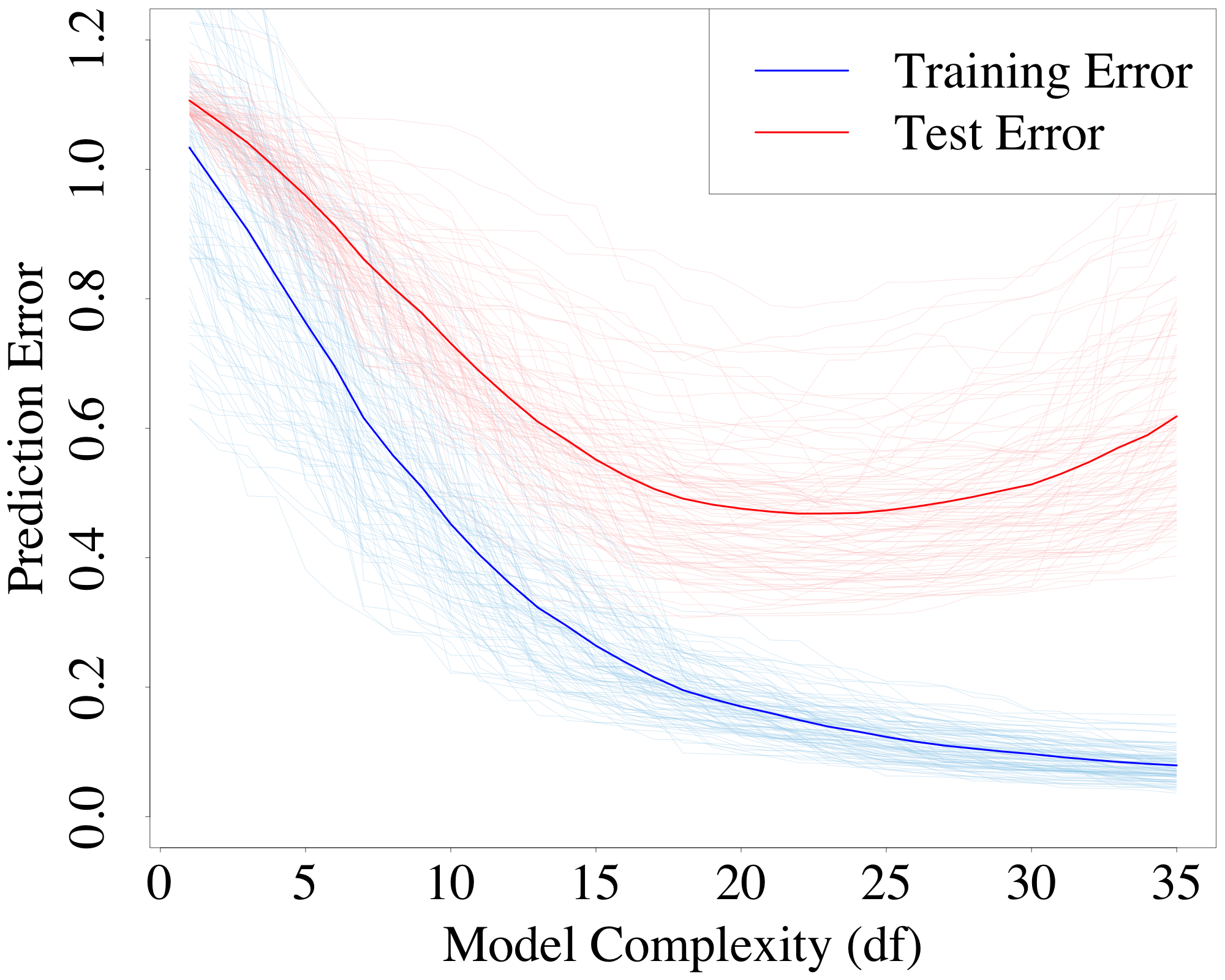}
\caption{Bias-Variance Tradeoff in Simulated Data}
\label{fig:hastie}
\end{figure}

One key variant of generalization that is used with nearly all deep learning models is the application of a trained model to new test cases. Across many areas of empirical research, models are known to forecast more accurately among \emph{in-sample} observations used in the fitting process than on previously unseen \emph{out-of-sample} observations, and the issue is particularly severe among complex models with large numbers of estimated parameters. This phenomenon is illustrated in Figure \ref{fig:hastie}, which is modified slightly from \cite{HastieTibshiraniFriedman2009} and shows training error in blue and test error in red, both as functions of model complexity. The image, constructed for illustrative purposes, was constructed by applying LASSO regression to simulated data.

On the left-hand side, the algorithms exhibit a bias toward overly smooth and simple functions that are ``underfitted'' and omit key predictors and nonlinearities, resulting in high errors in both the training and test samples. This effect is sometimes called ``inductive bias,'' which refers to the modeling assumptions necessary to make out-of-sample inferences \citep{Mitchell1997}.

The problem of underfitting, through which an overly simple model does not effectively respond to data, is analogous to the biological phenomenon of \emph{fixed action patterns} observed in insects and birds. \cite{Davies1989} note that the parenting behaviors of some ants and birds are evolutionarily preprogrammed and thus fixed over the animals' lifetimes. Their inability to adjust in response to outside stimuli like the characteristics of their offspring leaves them vulnerable to brood parasites. They are consequently tricked into expending resources raising the young of other species.

Mammals, with their greater cognitive resources, do not generally rely upon elaborate instinctive programs of behaviors that are locked down at birth. Instead, each generation must acquires skills like nesting through dopamine-driven learning \citep{Dewsbury1978}. While costly, this ability to learn and to respond to external events helps them to avoid repeat mistakes.

It is this form of learning that artificial neural networks seek to imitate. By incorporating parameters into the model that are sensitive to situation-specific factors---the input data---the model is trained to behave differently depending upon what features a given out-of-sample test case exhibits.

Moving to the right on the graph, the number of model parameters rises, and this source of error declines. While training error declines consistently with model complexity, test error falls initially and then rises due to variance caused by \emph{overfitting}---that is, some spurious predictors reduce training error, but their effects do not generalize out-of-sample. Due to this second source of error, which increases with model complexity, a model that is selected to minimize in-sample error is likely to be overly complex relative to a model that minimizes the error from out-of-sample forecasting.

\subsection{Conceptual Framework} \label{Conceptual Sample}

Despite these risks of under- and overfitting, theoretical work in machine learning provides conditions for achieving a certain type of generalization across samples known as \emph{probably approximately correct} (PAC) learning. A learning algorithm is PAC for some thresholds $\delta$ and $\epsilon$ if fewer than $100*\delta\%$ of unseen cases have a probability greater than $\epsilon$ of being misclassified. In this framework, $1-\delta$ is called the confidence level and $\epsilon$ is the error rate. For the examples presented in this section, $\delta = 0.01$ and $\epsilon = 0.05$. This definition allows for some cases with extremely high probability of misclassification, provided that they make up less than $1\%$ of the population. For the remaining $99\%$ of cases, the likelihood of misclassification is contained at $5\%$.

The number of training cases required to achieve this PAC standard of generalizability depends upon the function being learned, the distribution of cases in the population, and the algorithm. \cite{Vapnik1995} and \cite{Mohri2018} review bounds that have been derived for the minimum sample sizes required for PAC learning, and Table \ref{tab:paclearning} presents some example values. The ``Lower Bound'' column shows the lowest sample size at which PAC learning is possible for some target function, data distribution, and algorithm. It is calculated as the algorithm's \emph{Vapnik-Chervonekis (VC) dimension} divided by $320\epsilon^2$.\footnote{This is the ``unrealizable'' case in which the hypothesis set does not necessarily include the true target function but may include a sufficiently close approximation.} The ``Upper Bound'' column shows the sample size at which PAC learning is guaranteed for a broad set of possible target functions, data distributions, and algorithms. It is calculated as the sample size $m$ that numerically solves the nonlinear equation $m\epsilon^2/8 = d \log (2em/d) + \log (4/\delta)$, where $d<m$ is the VC dimension and the mathematical constant $e$ appears in the equation and as the base for the log. While \cite{Kawaguchi2022} derive tighter bounds for specific types of neural networks, the more general and well-known bounds are considered here.

The VC dimension is a measure of model complexity that is the machine learning analogue to degrees of freedom in classical statistics. It is the maximum number of training cases that an algorithm can ``shatter,'' that is, perfectly fit any set of labels applied to those cases. Imagine, for instance, a spam filter with possible labels of ``spam'' and ``not spam,'' with a training sample of 50 emails. There are $2^{50} \approx$ one quadrillion possible ways that these emails could be sorted into the two bins, each reflecting some hypothetical user's view of which emails should be kept and which should be filtered out. A learning algorithm with a VC dimension of 50 could perfectly fit this 50-observation training set, regardless of which of these quadrillion users labeled the emails. Given 51 emails, however, the algorithm lacks the expressive power to exactly match at least one of the $2^{51} \approx$ two quadrillion hypothetical users' labeling schemes.

While neural networks' VC dimensions are typically not known, the popular models all have some training error, which demonstrates that they cannot shatter sample sizes as large as their own training sets. The influential LeNet-5 model for classifiying handwritten digits \citep{Lecun1998} uses 60,000 training cases and has some training error when only 15,000 training cases are used. Thus, its VC dimension must fall below 15,000. Additionally, all known classifiers of the ImageNet-V2 \citep{Russakovsky2015} fail to shatter its 1.28 million-observation training sample and must have VC dimensions lower than that amount. \cite{Zhang2017, Zhang2021c} also show that some well-known neural networks have high but not perfect training accuracy on ImageNet-V2, even when the labels are randomly assigned.\footnote{For certain types of neural networks, \cite{Bartlett2019} provide tighter bounds on the VC dimension based upon models' numbers of parameters.} On the lower end, classifiers are unlikely to have VC dimensions lower than the number of classes (10 for LeNet-5 and 1,000 for ImageNet-V2). Additionally, while they do not establish shattering in the sense of capturing every possible labeling, \cite{Zhang2017, Zhang2021c} find that some sophisticated networks can perfectly predict random labels applied to the 50,000-observation CIFAR-10 training set \citep{Krizhevsky2014}.

\begin{table}[h]
\renewcommand{\arraystretch}{1}
\centering
\resizebox{0.45\textwidth}{!}{
\begin{small}
\begin{tabular}{p{0.12\textwidth}p{0.1\textwidth}p{0.1\textwidth}} \\
VC Dimension & Lower Bound & Upper Bound  \\
\hline \hline
$10$ & $13$ & 13.1 million \\
$100$ & $125$ & 126 million \\
$1,000$ & $1,250$ & 1.26 billion \\
$10,000$ & $12,500$ & 12.6 billion \\
$100,000$ & $125,000$ & 126 billion \\
one million & 1.25 million & 1.26 trillion \\
\hline
\end{tabular}
\end{small}}
\caption{Upper and Lower Bound Sample Sizes for PAC Learning when $\delta = 0.01$ and $\epsilon = 0.05$}
    \label{tab:paclearning}
\end{table}

As Table \ref{tab:paclearning} shows, the upper bound sample sizes at which PAC learnability is guaranteed are roughly a million times as large as the lower bound sample sizes at which some functions are PAC learnable. Thus, even if a network's VC dimension were known exactly, there is still a wide range of possibilities for the number of training observations required to learn a given empirical problem. Target functions that are easier to learn, given the algorithm and data distribution, require sample sizes near the lower end of the range, and functions that are harder to learn require sample sizes near the higher end.

One might reasonably expect common machine learning problems, which many humans can solve with little effort, to be easier to learn than random or worst-case groupings of elements, as noted and observed empirically by \cite{Kawaguchi2022}. Hence, we might expect them to require sample sizes in the lower part of the range. Indeed, LeNet-5's error rate on unseen test cases is below 1\%, even though its 60,000-observation training set is considerably lower than the millions or billions that the upper bound would require.

\subsection{Empirical Strategies} \label{Empirical}

Because highly complex neural networks generalize so well relative to the upper bound for PAC learnability, deep learning research makes extensive use of models that are ``overparameterized''---that is, those whose model complexity far exceeds the number of training observations. As in the the right hand side of Figure \ref{fig:hastie}, the training error is particularly low, but the model may suffer from substantial degradation in performance when generalizing out-of-sample due to the variance that arises from overfitting.

The strategies that neural network researchers use to mitigate this risk are described as ``regularization'' and are described in detail in surveys by \cite{Bejani2021,Bishop2006}, \cite{GoodfellowBengioCourville2017}, \cite{DosSantos2023}, \cite{Kukacka2018}, and \cite{Moradi2020}. The approaches can be grouped into three categories: simplifying the model, coarsening the fit, and augmenting the data. The list below describes popular techniques within each of these categories:

\begin{enumerate}[leftmargin=*,labelindent=0em,label=(\arabic*)]
  \item \textbf{Simplify Model}. Useful for improving generalizability while also reducing computational cost and increasing explainability.
  \begin{itemize}[leftmargin=0em]
    \item \emph{Multi-objective Optimization.} Algorithm modified to balance training error and model complexity, often measured as magnitude of parameter weights. Frequently implemented through constrained optimization. Examples in \cite{Albuquerque2000, Rocha2020} and \cite{Torres2022} with a helpful survey by \cite{Tian2022}.
    \item \emph{Weight Sharing.} Reuse weight values across the network to reduce parameters and enhance generalization. Common in convolutional \citep{Lecun1998} and recurrent networks \citep{Bengio1994}, language translation \citep{Press2017, Inan2017} and autoencoders \citep{Vincent2010}.
    \item \emph{Pruning.} Eliminate neurons or connections with little influence post-fit \citep{Cheng2023, Li2023, Marino2023}.
    \item \emph{Knowledge Distillation.} Train a smaller model to mimic a larger one \citep{Hinton2015, Gou2021}.
  \end{itemize}
  \item \textbf{Coarsen Fit}. Helps to concentrate training on broad patterns, which can be most reliably and robustly learned, rather than on small fluctuations.
  \begin{itemize}[leftmargin=0em]
    \item \emph{Noise.} Add random values to gradients or parameters to reduce relative importance of minor perturbations \citep{Neelakantan2015, Blundell2015}.
    \item \emph{Dropout.} Temporarily drop some neurons during each training iteration to reduce overreliance on individual neurons \citep{Hinton2012, Srivastava2014}.
    \item \emph{Batch and Layer Normalization.} Demean and scale inputs within minibatches and across layers to limit influence of artificial sampling variation on parameter updates \citep{Ioffe2015}.
    \item \emph{Early Stopping.} Stop training once validation error increases, even if training error continues to decline \citep{Prechelt2012}.
    \item \emph{Double Descent.} Overparameterize models to achieve better generalization, with empirical evidence supporting improved out-of-sample performance at very high levels of model complexity \citep{Belkin2019, Kawaguchi2022, Nakkiran2021, Neyshabur2017}.
  \end{itemize}
  \item \textbf{Augment Data.} Useful for addressing specific known forms of overfitting but increases training cost.
  \begin{itemize}[leftmargin=0em]
    \item \emph{Input Adjustment.} Add noisy, shifted, or transformed versions of training data to enhance robustness against common classification errors \citep{Bishop2006, Shorten2019}.
    \item \emph{Adversarial Learning.} Perturb inputs in worst-case directions to create challenging training cases \citep{Akhtar2019, Chakraborty2018, Goyal2023, Qian2022}.
    \item \emph{Multi-task Learning.} Train the model on more than one related task. For instance, one task might involve identifying when an object or recognizable sound is in frame, while another would determine which object or sound it is, with network elements shared for steps like edge detection that both tasks require \citep{Vandenhende2021, Zhang2022}.
  \end{itemize}
\end{enumerate}

These three categories of regularization approaches tackle the problem of overfitting in very different ways. The first, model simplification, has a straightforward analogue to classical statistics, in which overfitting is directly related to the number of free parameters. It applies the intuition from Figure \ref{fig:hastie}, reducing the number of parameters---or the magnitudes of those parameters---as a means of improving generalizability.

The second category, by contrast, is specific to the mechanics of complex machine learning systems without unique or exact solutions. Approaches to coarsen the fit are born out of experimentation with those systems.

The final category, data augmentation, most closely resembles strategies used in biological learning. As \cite{Sinz2019} notes, data collection in biological systems is curiosity-driven, focusing on training cases that deviate from expectation or fill specific gaps in understanding. By nudging inputs in worst-case directions, adversarial learning employs a similar strategy in the artificial network setting. \cite{Sinz2019} also observes that biological brains divide into modules, and each component of the system is informed by observations from a variety of applications in which it is used. Because of this design feature, biological learning is relatively insensitive to aberrations in any one domain of knowledge. The regularization strategy of multi-task learning applies a similar insight to artificial networks.

\subsection{Application to ImageNet} \label{ImageNet}

\begin{figure}
\begin{subfigure}{0.45\textwidth}
  \centering
  % include first image
  \includegraphics[width=\textwidth]{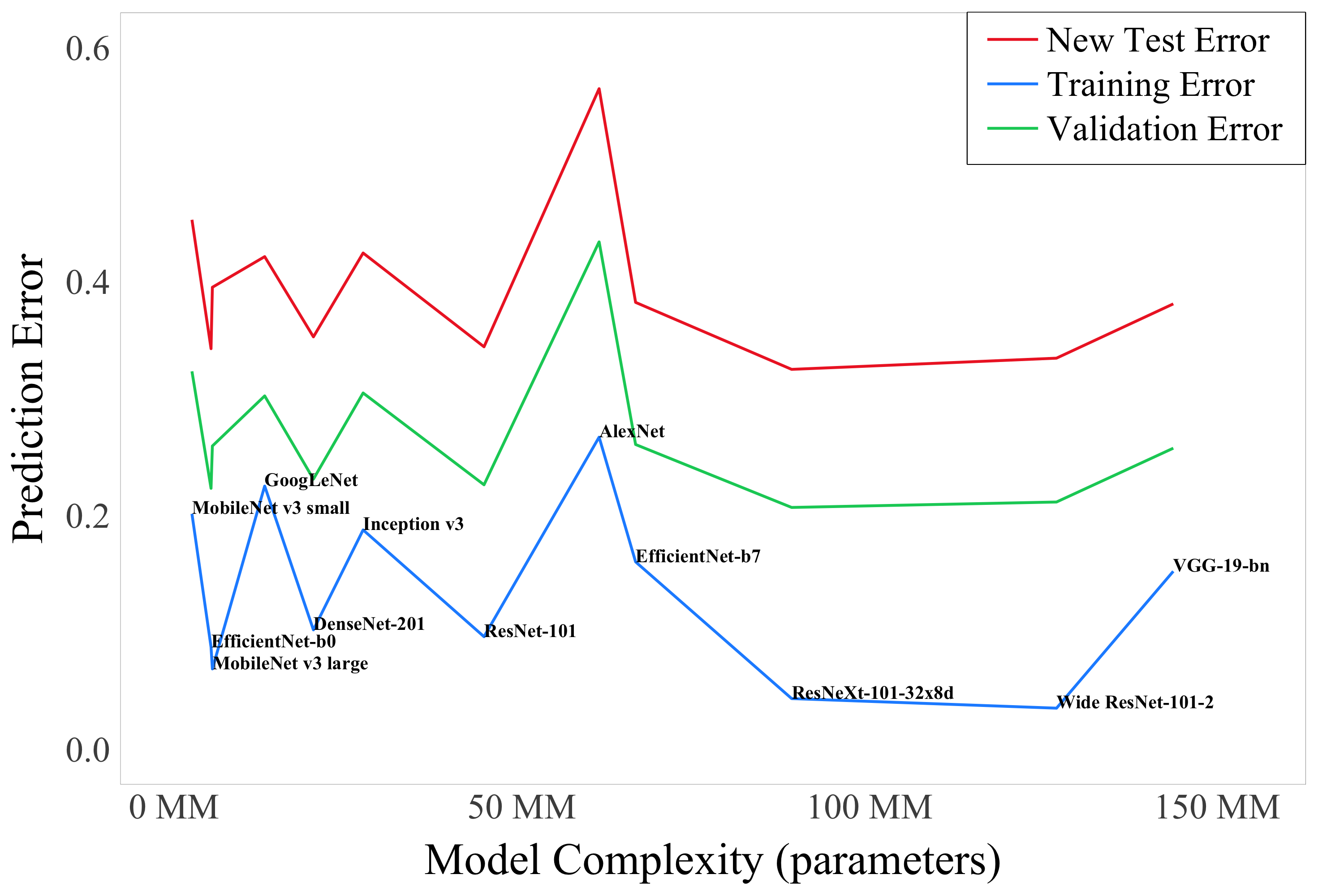}
  \caption{Complexity is Number of Parameters}
  \label{fig:overfit_nn}
\end{subfigure}
\begin{subfigure}{0.45\textwidth}
  \centering
  % include first image
  \includegraphics[width=\textwidth]{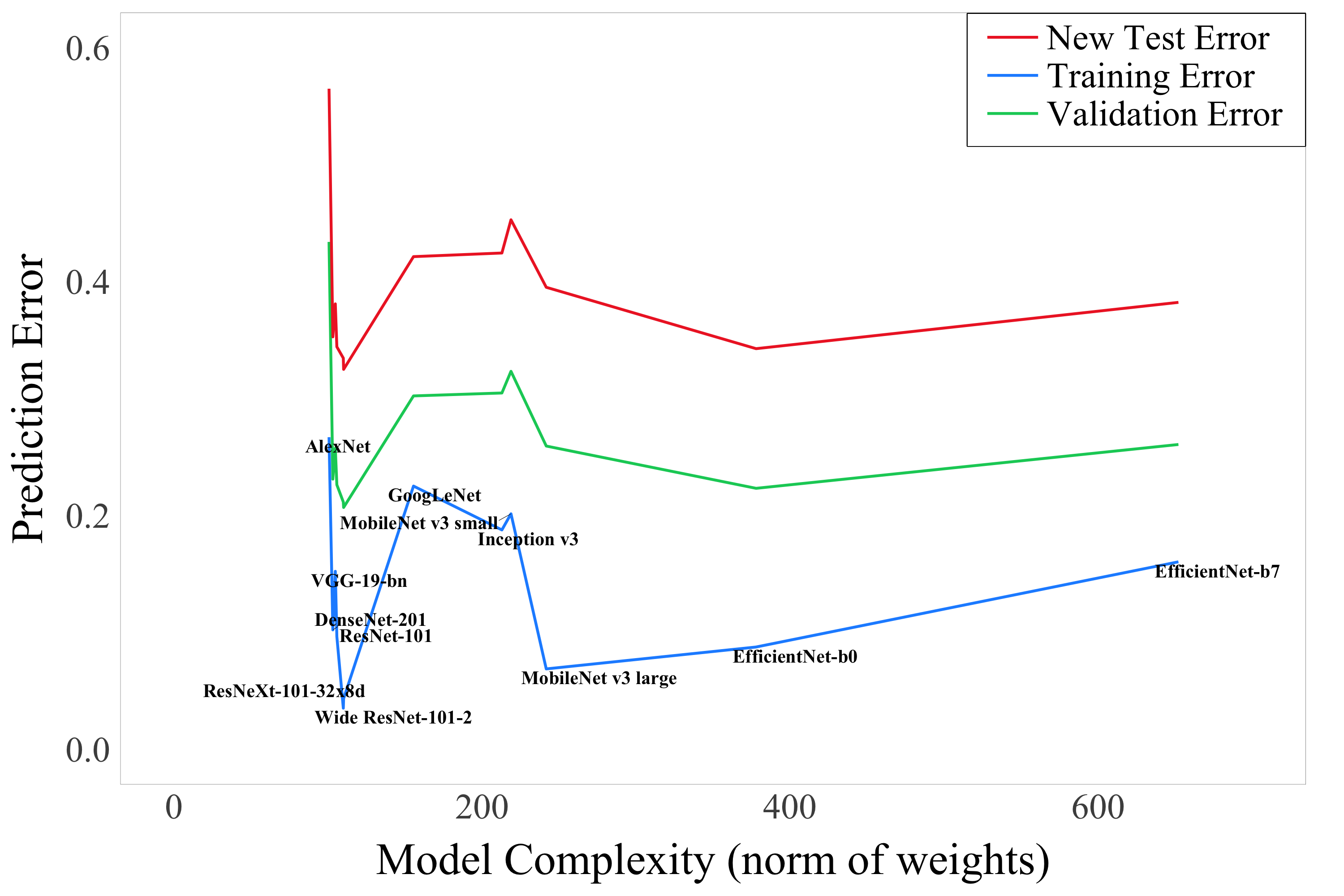}
  \caption{Complexity is Norm of Weights}
  \label{fig:overfit_norm}
\end{subfigure}
\begin{subfigure}{0.45\textwidth}
  \centering
  % include first image
  \includegraphics[width=\textwidth]{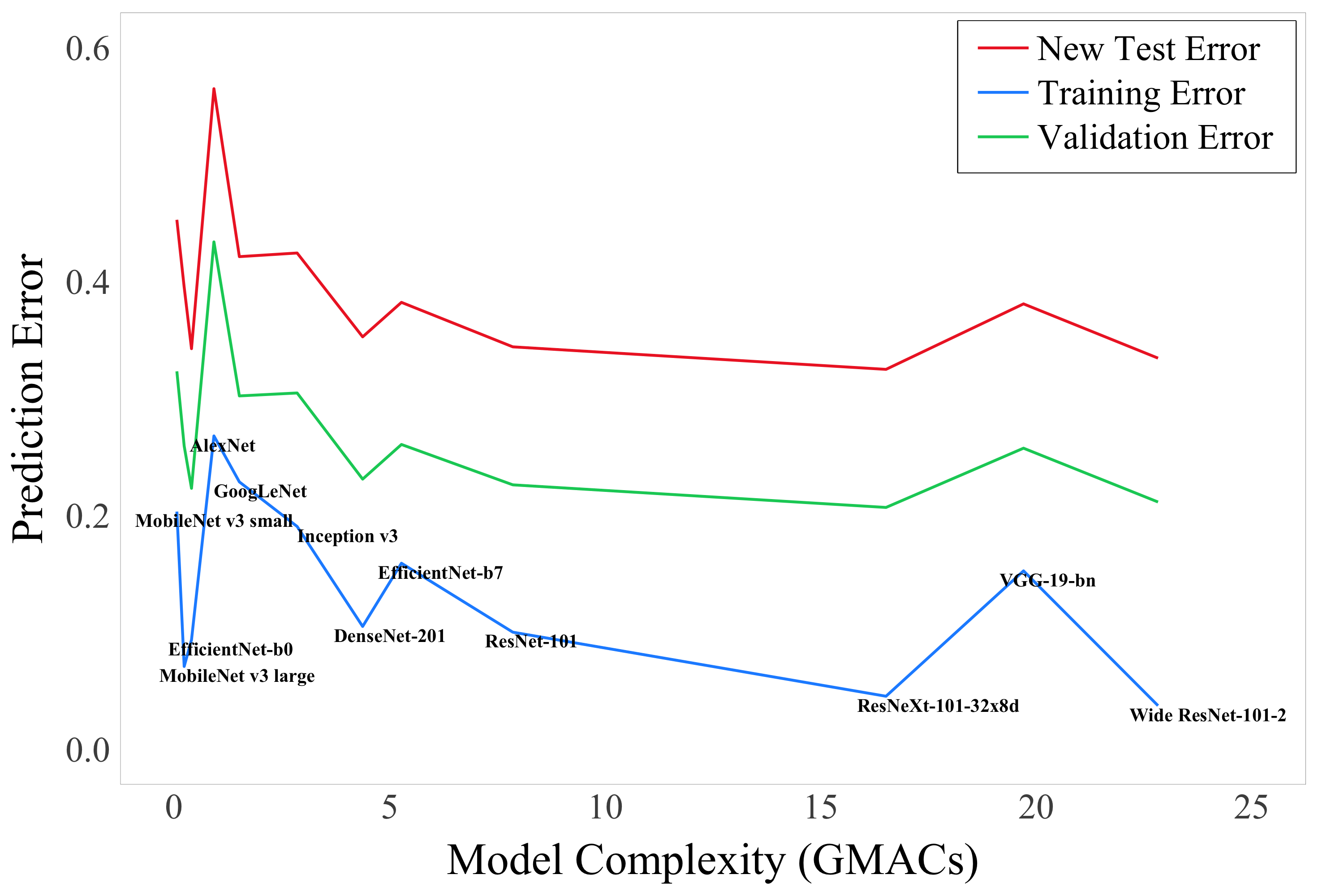}
  \caption{Complexity is GMACs}
  \label{fig:overfit_gmac}
\end{subfigure}
\caption{Error Rates versus Complexity on the ImageNet Data}
\label{fig:overfitting}
\end{figure}

\begin{table*}[ht!]
\renewcommand{\arraystretch}{1.25}
\centering
\resizebox{\textwidth}{!}{
\begin{tabular}{p{0.16\textwidth}p{0.16\textwidth} c c c p{0.29\textwidth} c c c c}
\multirow{2}{*}{Model} & \multirow{2}{*}{Source} & \# Parame- & Norm of & GMACs & \multirow{2}{*}{Key Design Elements} & \multicolumn{3}{c}{\% Accuracy of Top-1 Selection} & \multirow{2}{*}{\shortstack{Training\\-New Test}} \\
 & & ters (MM) & Weights & per Case & & Training & Validation & New Test & \\
\hline
EfficientNet-b0 & \cite{Tan2020} & 5.3 & 377.9 & 0.40 & Coordinated scaling of width, & 90.6\% & 77.7\% & 65.7\% & 25.5\% \\
EfficientNet-b7 &  & 66 & 652.2 & 5.27 & depth, and resolution & 84.1\% & 73.9\% & 61.8\% & 22.2\% \\
MobileNet v3 large & \cite{Howard2019} & 5.5 & 241.6 & 0.23 & Network architecture search & 92.9\% & 74.1\% & 60.5\% & 32.6\% \\
MobileNet v3 small & & 2.5 & 218.7 & 0.06 & & 79.7\% & 67.7\% & 54.7\% & 25.1\% \\
ResNeXt-101-32x8d & \cite{Xie2017} & 89 & 110.0 & 16.51 & Modular structure with repeated use of common transformation architecture & 95.5\% & 79.3\% & 67.5\% & 28.2\% \\
DenseNet-201 & \cite{Huang2017} & 20 & 103.0 & 4.37 & Dense connectivity across non-adjacent layers & 89.5\% & 76.9\% & 64.7\% & 25.1\% \\
Wide ResNet 101-2 & \cite{Zagoruyko2017} & 127 & 109.8 & 22.82 & Feedback, width & 96.3\% & 78.8\% & 66.5\% & 29.9\% \\
ResNet-101 & \cite{He2016} & 45 & 105.6 & 7.85 & Feedback, depth & 90.0\% & 77.4\% & 65.6\% & 24.8\% \\
Inception v3 & \cite{Szegedy2016} & 27 & 212.8 & 2.85 & Parallelism, expanded convolution and regularization & 80.9\% & 69.5\% & 57.6\% & 23.7\% \\
GoogLeNet & \cite{Szegedy2015a} & 13 & 155.4 & 1.51 & Parallelism, increased depth and width & 77.1\% & 69.8\% & 57.9\% & 19.6\% \\
VGG-19-bn & \cite{Simonyan2015} & 144 & 104.6 & 19.70 & Increased depth with small convolutional filters & 84.7\% & 74.2\% & 61.9\% & 22.9\% \\
AlexNet & \cite{Krizhevsky2014} & 61 & 100.5 & 0.92 & Computational parallelism & 73.2\% & 56.6\% & 43.5\% & 29.8\% \\
\hline
\end{tabular}}
\caption{Sample Generalization of Commonly Used Classifiers on the ImageNet Dataset}
    \label{tab:sample}
\end{table*}

Next, to better understand the empirical relevance of overfitting, we turn to an empirical application from the field of computer vision. Specifically, we compare the training, validation, and test performance of twelve established neural networks on the ImageNet-V2 task of identifying objects from a set of 1,000 possible categories. Many of the original studies do not report training accuracy. For the purposes of this discussion and following convention in the literature, overfitting is measured here as the difference between training accuracy and test accuracy \citep{GoodfellowBengioCourville2017, HastieTibshiraniFriedman2009}. Training and validation accuracy were consequently reproduced for all twelve models by applying the pre-trained model parameters from PyTorch \citep{Paszke2019} to the 1,281,167-observation training set and the 50,000-observation validation set from the original ImageNet-V2 data \citep{Russakovsky2015}. Additionally, a 10,000-observation supplemental dataset from \cite{Recht2019} was compiled using the same conventions as the original ImageNet samples but was not available to the modelers during training.\footnote{Of the three new test datasets provided by \cite{Recht2019}, the ``matched frequency'' one is used so that the proportions in each category are the same as in the original data. Following the PyTorch documentation, a standard set of transformations is applied to the images: each is first resized to 256, then center-cropped to 224, then normalized with RGB channel-specific means of 0.45, 0.456, and 0.406 and standard deviations of 0.229, 0.224, and 0.225.} That sample is used to measure out-of-sample performance in a similar exercise as in \cite{Recht2019}. Those results, are labeled here as ``New Test,'' are used here to assess test accuracy in the overfitting measure.

The 12 neural networks considered all use many of the regularization techniques described in the previous subsection. All use multi-objective optimization in the form of norm-based penalties for complexity, and all manage model complexity further by employing convolutions and other strategies for weight-sharing. Additionally, all 12 perform dropout and batch normalization to coarsen the fit. Some also adjust the learning rate across training iterations in a way that resembles early stopping, and some achieve the high levels of complexity that has been found to produce the ``double descent'' phenomenon. Additionally, all augment the training data with transformed versions of the images, including flipping, shifting, cropping, and modifying the colors.

To match the layout of Figure \ref{fig:hastie}, Figures \ref{fig:overfit_nn}, \ref{fig:overfit_norm}, and \ref{fig:overfit_gmac} present these performance statistics as the fraction that are incorrectly classified, labeled as ``Prediction Error.'' If $A$ is the number of correct predictions divided by the total, then $1-A$ is the value plotted along the vertical axes. As with Figure \ref{fig:hastie}, this error is plotted against model complexity on the horizontal axis. In Figure \ref{fig:overfit_nn}, complexity is measured as the number of parameters. Following \cite{Bartlett1997}, Figure \ref{fig:overfit_norm} measures complexity as the magnitude of the estimated weights, computed here as the Euclidean norm. In Figure \ref{fig:overfit_gmac}, required computations, measured as Giga-Multiply-Accumulate (GMAC) is used as the measure, which is a variation on Floating Point Operations (FLOPs) that is calculated using the fvcore profiler in Python \citep{Yang2020c}.\footnote{While such profiler-based metrics have their limitations, \cite{Rohlfs2023c} obtains similar complexity numbers by timing calculations on a personal computer.}

The model names appear next to their levels of performance in the training data.

The same findings from Figure \ref{fig:overfitting} also appear in Table \ref{tab:sample}. The models are listed in table in decreasing order of publication date. For each of these models, the table presents the source, the values for both measures of complexity, a brief description of the novel elements introduced by that network, and the performance data expressed in terms of percent accurate, or $100*A$. The key measure of overfitting, the difference between training and test accuracy, is shown in the right-hand column.

The key finding from Figures \ref{fig:overfit_nn} and \ref{fig:overfit_norm} and Table \ref{tab:sample} are that, despite the many regularization approaches used, the amount of overfitting, is substantial. The magnitude of this gap, shown in the right-hand column of the table and illustrated graphically as the difference between the red and blue lines, ranges from 19.6 to 32.6 percentage points. In 11 of the 12 models considered here, this gap from overfitting is larger than the training error, GoogLeNet being the one exception.

Across these 12 neural networks, there are suggestive patterns of increased training accuracy and increased norm-based complexity over time, with values of both variables generally higher for entries at the top of the table than at the bottom. No such pattern can be seen for overfitting, however. Thus, the evidence from these 12 models suggests that progress in ImageNet accuracy over this period has been effectively independent of overfitting. While training error has declined considerably, the preliminary findings from this comparison indicate that overfitting has been stubbornly persistent as a sizeable problem, and there is not a clearly discernible relationship between the amount of overfitting and model complexity.

These models already employ many of the regularization strategies typically applied to avoid overfitting. While it is unclear exactly why the amount of test error is so consistently large, one possible explanation is that the training data are insufficiently diverse. Augmenting the data with a wider variety of cases---particularly unusual-looking and counterfactual examples that would help with out-of-distribution generalization---is one possible avenue that is worth considering for addressing this issue.

Another pattern visible from Figures \ref{fig:overfit_nn} and \ref{fig:overfit_norm} and Table \ref{tab:sample} is the large and stable difference between validation and test error. This difference, seen in the graphs as the gap between the green and red lines, ranges from 11.8 to 13.6 percentage points. This evidence indicates that the overfitting seen in popular ImageNet classifiers is only partially attributable to overfitting of the model weights in the training data. This preliminary evidence is consistent with a roughly equal amount of variance coming from network architecture, hyperparameter selection, and other broad judgments made during the design and validation of the model.

\section{Distribution Generalization} \label{Distribution}

\subsection{Broadening Out-of-Sample Performance} \label{Broadening}

\begin{figure*}[b!]
\begin{subfigure}{0.45\textwidth}
  \centering
  % include first image
  \includegraphics[width=\textwidth]{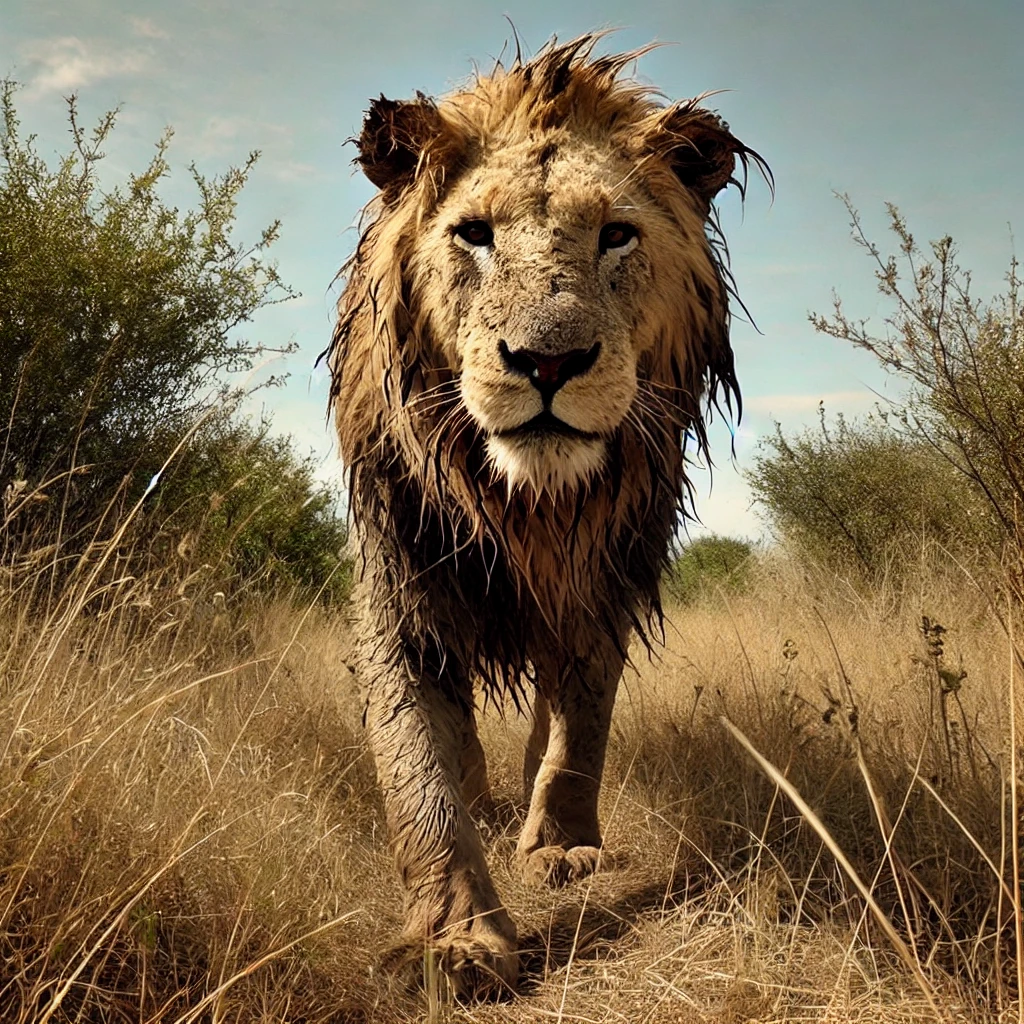}
  \caption{Lion in Natural Habitat}
  \label{fig:lionhabitat}
\end{subfigure}
\hfill
\begin{subfigure}{0.45\textwidth}
  \centering
  % include first image
  \includegraphics[width=\textwidth]{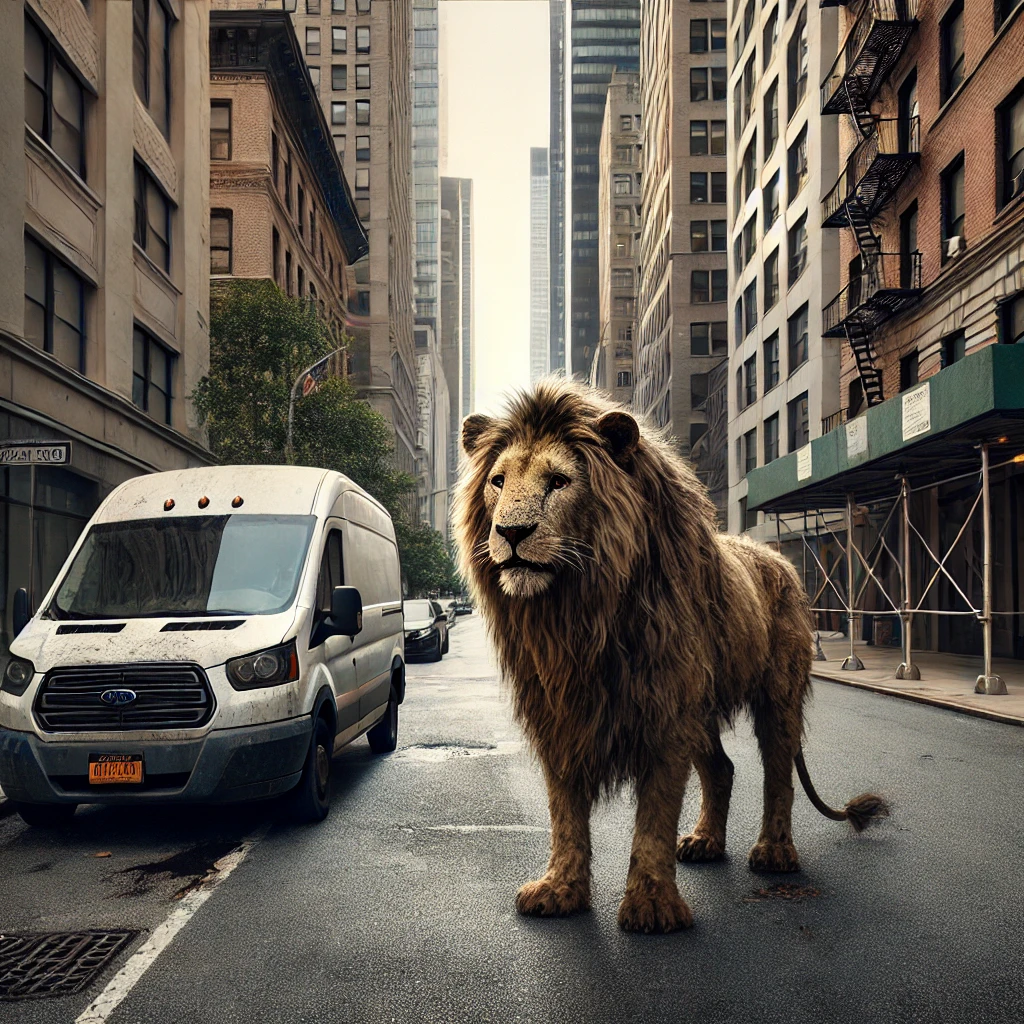}
  \caption{Lion in the City}
  \label{fig:lioncity}
\end{subfigure}
\caption{Example In- and Out-of-Distribution Test Cases}
\label{fig:lion}
\end{figure*}

In many situations in practice, researchers wish to generalize models more extensively than simply to randomly selected test samples. In medical settings, it is often desirable to use a single model to diagnose multiple types of patients and to maintain a high standard of accuracy not just for patients on average but separately for each type. Market researchers may want to forecast consumers' responses to a previously unseen product or advertisement. Educators might try to predict how expanding a pilot program to additional classrooms or schools would affect student outcomes. Applications of the model to subpopulations or to rare or hypothetical cases in this way are forms of \emph{out-of-distribution generalization}.

In all of these contexts, the target function $f$ that maps features to labels remains unchanged. One of those specific or hypothetical test cases could in principle appear in the training data and would have the correct label applied to it. A random test sample could possibly include it as well, but that case would be one of many and would appear with low probability. What differs from the problem of sample generalization is that there is a particular set of test cases that are of interest, and they are not randomly drawn from the same population as the training sample is.

If the target function $f$ and all the relevant predictors were known with absolute certainty, the researcher could accurately classify any test cases, regardless of the distribution from which they are drawn. A key difficulty arises in learning because the function itself is an approximation that is sensitive to the proportions in which different types of cases appear. The definition of PAC learning allows for hypotheses that have blind spots in the sense that a PAC-learned classifier might perform terribly on some fraction $\delta$ of test cases. No performance guarantee would apply to a test sample that consists solely of this subset of hard-to-classify cases.

To illustrate a key problem that this form of generalization poses, Figure \ref{fig:lion} presents example in- and out-of-distribution cases that could be presented to a object recognition classifier. Both images were artificially generated using DALL-E from OpenAI \citep{Openai2024a, Openai2024b}. Both clearly depict lions, but the first shows one in its natural habitat, while the other presents one in an urban setting. Moving the animal from the savannah does not change the ground truth class to which it belongs, and any human looking at the images would recognize both as lions. In practice, however, the ImageNet training images of lions come almost exclusively from natural and semi-natural settings like wildlife reserves and zoos. Many artificial network-based classifiers rely upon environmental cues like grass and dirt in the surroundings to inform their predictions, and the change in background can throw them off.

Table \ref{tab:lion} presents the labels that each of the 12 artificial neural network models from Table \ref{tab:sample} apply to the two images in Figure \ref{fig:lion}, along with the level of confidence to which they assign those labels. As the table shows, all 12 of the artificial neural networks listed in Table \ref{tab:sample} correctly identify Figure \ref{fig:lionhabitat} as a lion, and they generally do so with high confidence. Only half, however, correctly label the image in Figure \ref{fig:lioncity} as a lion. The other six models are evenly split between labeling the animal as a chow chow or a Tibetan mastiff, two dog breeds that are much more likely than a lion to be seen on a city street. While not shown in the table, nearly all of the models that misclassify Figure \ref{fig:lioncity} do include lion in their top five guesses. The sole exception is MobileNet v3 small, whose top five are all dog breeds.

\begin{table}[h]
\renewcommand{\arraystretch}{1}
\centering
\resizebox{0.50\textwidth}{!}{
\begin{small}
\begin{tabular}{c c c c c}
\multirow{2}{*}{Model} & \multicolumn{2}{c}{Natural Habitat} & \multicolumn{2}{c}{City Street} \\
 & Label & Confidence & Label & Confidence \\
\hline
EfficientNet-b0 & lion & 79.6\% & chow chow & 57.0\% \\
EfficientNet-b7 & lion & 85.9\% & lion & 67.3\% \\
MobileNet v3 large & lion & 99.7\% & chow chow & 89.5\% \\
MobileNet v3 small & lion & 86.8\% & Tibetan mastiff & 20.5\% \\
ResNeXt-101-32x8d & lion & 94.5\% & lion & 99.4\% \\
DenseNet-201 & lion & 97.6\% & chow chow & 89.0\% \\
Wide ResNet 101-2 & lion & 99.7\% & lion & 96.8\% \\
ResNet-101 & lion & 99.8\% & lion & 96.0\% \\
Inception v3 & lion & 100.0\% & lion & 99.3\% \\
GoogLeNet & lion & 49.7\% & Tibetan mastiff & 14.0\% \\
VGG-19-bn & lion & 96.4\% & lion & 66.2\% \\
AlexNet & lion & 95.4\% & Tibetan mastiff & 78.4\% \\
\hline
\end{tabular}
\end{small}}
\caption{Model Predictions for Two Lion Photos}
    \label{tab:lion}
\end{table}

In related work using a Recurrent Neural Network (RNN) to classify handwritten digits, \cite{Horoi2020} find a high degree of sensitivity to slight modifications in test cases. The authors find that, when rows are deleted from a test image, when the last rows are blacked out, or when extra blank rows are added to the sequence, the model projections are inaccurate and unstable. The authors additionally find that the intermediate outputs generated from successive iterations of the recurrence exhibit volatile behavior and do not follow intuitive patterns from one iteration to the next.

In addition to examining specific ways of manipulating test cases, as in Table \ref{tab:lion} and the digit-classification exercise, some work takes a systematic approach to the problem of generalization across distributions. A survey by \cite{Lust2021} also highlights the use of adversarial examples as a promising means of identifying the limits of a model to generalize out-of-distribution. Such studies generate different types of adversarial cases to trace out the ``generalization envelope'' describing the scope of distributions over which the model can be expected to perform consistently.

To assess the ability of learned systems to generalize across distributions, another area of research applies reweighting and subsampling to the training data to simulate potential distributions of test cases. This approach might not help in the classification of hypothetical test cases like Figure \ref{fig:lion} or a never-before-seen ad or product, which have negligible likelihood of appearing in training data. They may, however, help to understand performance across subpopulations, as in the medical diagnosis example described above. In one approach known as Distributionally Robust Optimization (DRO), researchers train model parameters to ensure a minimum level of performance in worst-case scenarios. \cite{Liu2023} note in a useful summary that explicit distribution-aware optimization strategies of this form have been found to be impractical due to their high computational demands and sensitivity to underlying assumptions about the potential test sample distributions.

Beyond these strategies for assessing a model's applicability across potential out-of-distribution test cases, researchers in some cases have explored the nature of the errors that arise in such contexts. Next, we will consider ways in which the data and model structure can be built out to improve deep learning models' abilities to produce accurate projections across different distributions of test cases.

\subsection{Counterfactuals and Causality} \label{Causality}

The generalizability of neural network models to out-of-distribution cases is not unlike the identification of counterfactuals in classical statistics. Consider, for instance, a deep learning model that uses data on patients' medical histories to project their likelihoods of heart disease. If sufficiently many training cases are used, learning theory guarantees that the predictions of heart disease will be PAC for test cases drawn at random from that same population. What if, however, a doctor is considering advising patients to stop smoking? The doctor's advice effectively envisions counterfactual out-of-distribution test cases. The hypothetical patients that the doctor imagines have histories of smoking, and they exhibit whatever behaviors and genetic, socioeconomic, and psychological factors are correlated with smoking in the data. Due to the doctor's intervention, however, they do not smoke.

A model that captures general patterns in the data might overstate or understate the importance of smoking relative to other risk factors like cholesterol intake and sedentary lifestyle---and the direction and magnitude of that error is likely to vary across patients in ways that are hard to predict. Such imprecision will not cause major problems in conventional out-of-sample forecasting, because smoking tends to appear along with those other risk factors in the general population. For in-distribution purposes, smoking, cholesterol, and lifestyle are good enough proxies for one another. As noted in Section \ref{Sample}, coarse fitting procedures that focus on general patterns tend to generalize well in-distribution. When the researcher wishes to apply the model to out-of-distribution counterfactual cases, however, this failure to correctly isolate each predictor's contribution becomes important.

At the heart of this issue are \emph{confounding variables}, features that co-move with the predictor of interest and also have their own impacts on the model outcome. The difficulty in inference that these variables cause is known in statistics and econometrics as \emph{omitted variables bias}. It is this same problem that prevents the 12 popular neural networks from recognizing a lion in the city. If a photo of a lion is drawn from the same distribution as was used for training, it will include the natural-looking surrounding environment that one would expect from the training data. The pattern in which lions always appear together with grass and dirt is only broken by the appearance of a counterfactual, out-of-distribution test case.

As the lion in the city example shows, being able to observe all the relevant data does not necessarily solve the problem of identifying counterfactuals. In that situation, the direct inputs to the model are pixel values, and the confounding variables---descriptors of the background environment---are complex combinations of those pixel values. Those confounding features could potentially be learned in the deep layers of the model. Training data do not, however, typically include cases that shift those features independently of the objects being recognized. For this reason, we should not expect models to learn how to disentangle deep confounders from the ground truth objects they seek to identify.

Even if it is observed directly, as might be the case with a predictor like cholesterol intake in the smoking example, a confounding variable could still get in the way of forecasting counterfactual outcomes. Due to the coarseness of many neural networks' fits and the lack of independent variation in the predictor of interest---in this case smoking cessation---a deep learning model may fail to accurately separate the effect of that variable from that of the confounder. Additionally, many of the confounding variables, like the many genetic, socioeconomic, and psychological determinants of smoking, can only be measured with rough proxies. For cases like health outcomes in which key drivers are unobservable, this problem of confounding variables and a lack of out-of-distribution generalizability is inevitable.

Given the practical value of counterfactuals for guiding policy decisions, business strategies, and other real-world interventions, some researchers have begun to explore the identification of causal relationships learning-based approaches. In addition to having these direct applications for evaluating potential courses of action, causal links are often tied to physical, biological, or economic processes. Many argue that such fundamental relationships are more stable and robust than statistical associations, which can be especially sensitive to differences across populations and over time---and thus more reliable for generalization across populations and distributions \citep{Pearl2009, Rubin1974, Scholkopf2021}.

Outside of machine learning, statisticians and econometricians have developed a variety of approaches for identifying counterfactual scenarios and causal relationships. \emph{Randomized Controlled Trials} (RCTs) are widely agreed to be the most reliable. By directly choosing who receives a treatment, the researchers can ensure that it is independent of any potential confounding factors. RCTs are often costly and raise ethical issues, and so another popular strategy involves ``natural experiments'' or ``quasi-experiments'': events that caused some essentially random set of subjects to receive a treatment, so that the experimental environment can be simulated using observational data. This essentially random driver of treatment is called an \emph{instrumental variable}. In the example of quitting smoking, regulations such as cigarette taxes are sometimes used as instrumental variables. \cite{DeCicca2022} provides a review. For situations in which RCTs and quasi-experimental strategies are not viable but key confounding variables are observed directly, researchers sometimes use regressions or matched-pair comparisons to separate their effects from those of the predictor of interest. The use of ``control variables'' in this way is referred to by \cite{Pearl2009} as a ``back-door'' adjustment. Popular early references for RCTs, quasi-experiments, matching, and other methods for causal inference include \cite{Campbell1963, CookCampbell1979}, and \cite{Rubin1974}. \cite{AngristPischke2009} and \cite{Pearl2009} provide more recent overviews.

Research on causality and counterfactuals is not as well developed for learning-based settings as it is in classical statistics. One area that has been explored in the learning-based estimation of causal relationships is ``structural'' graph-based models. Popular strategies of this form are built around \emph{Directed Acylic Graphs} (DAGs). Predictors are ordered into tiers based upon their relative positions in the causal chain, which is then used to inform the network structure. Orderings are obtained from domain knowledge or through adversarial learning, based on the premise that relationships that are robust to adversarial attacks are likely to be causal. \cite{Kaddour2022, Peters2017, Scholkopf2021, Vowels2022}, and \cite{Zanga2022} provide useful surveys. Such structures may help to rule out some highly implausible effects, such as tomorrow's outcomes impacting events today. They do not, however, address the far more serious challenge of unobserved confounding variables, and they are generally not tested for out-of-distribution generalization.

Counterfactuals and deep learning both spark intense interest among practitioners, and the intersection of the two is a ripe area for future research. In addition to graph-based strategies, some promising recent studies exist combine established methods for identifying counterfactuals with learning-based approaches for capturing complex, nonlinear relationships with rich sets of interaction effects. Summaries can be found in \cite{Guo2020, Athey2019} and \cite{Chernozhukov2024}. While few combine neural networks with experimental or quasi-experimental designs, one notable exception is \cite{Hartford2017}. That paper applies an deep learning instrumental variables approach to simulated data and to a real world application of online ads. In the application, the researchers find that the network-based approach produces similar findings as an earlier parametric study but with less manual effort due to automated feature selection. In another interesting study, \cite{vanderLaan2011} describe an ensemble-based ``targeted learning'' approach through which shallow or deep learning models can help to refine the estimated treatment effects obtained from RCTs and increase the level of precision with which impacts can be estimated. \cite{Pirracchio2020} presents an application that combines shallow learning models with RCTs to assess a corticosteroid treatment.

Another area of interest in causal machine learning is matching. \cite{Chernozhukov2018} and \cite{Wager2018}, for instance, combine shallow learning with propensity score matching. Some other researchers use rich sets of covariates and the learned structure from neural networks to predict outcomes for counterfactual test cases and estimate causal impacts at the individual level \citep{Hassanpour2019, Johansson2016, Louizos2017, Shalit2017}. These deep matching approaches are applied to the data from well-known RCTs, and their average treatment effects are generally close to those obtained from the randomized experiments.

As deep learning researchers aim to improve models' generalizability across distributions and populations, data is increasingly becoming at least as important as methodology. In the controlled setting of the game of Go, the use of challenging self-generated training cases dramatically improved model performance \citep{Silver2016, Silver2017}. As the lion in the city example illustrates, confounding variables can be deep and difficult to learn, and thinking about counterfactuals has the potential to identify models' blind spots. The adversarial examples that researchers often create through slight perturbations to training data help to improve models' robustness, in part by simulating the curiosity-driven learning of biological systems that \cite{Sinz2019} describes. Augmenting training data with more severe and stranger modifications has the potential to do even more.

\subsection{Interaction Effects} \label{Interactions}

Given the many constraints associated with running large RCTs or finding credible quasi-experiments, the extent to which an estimated causal impact can generalize across populations has long been a focus of the statistics literature and is known to statisticians as \emph{external validity} \citep{Campbell1963,CookCampbell1979,ImbensAngrist1994,Card2001,HeckmanVytlacil2005}. It is the need to capture cross-subject variation in impacts---and to better generalize to other populations---that has led to increased interest in interaction effects and the learning-based models that can capture them.

One of deep learning's key contributions to measurement and inference is the ability to identify complex and deep features that arise through multilayered interactions across many input variables. Future research could apply neural networks to existing data from a wide range of RCTs and natural experiments to shed light on how those impacts vary across subjects and how the effects can be effectively generalized to other populations. Easy-to-use tools that enable applied researchers to leverage network-based approaches could be particularly valuable.

This importance of interactions in generalization is paradoxical. The bias-variance tradeoff shown in Figure \ref{fig:hastie} suggests that the ability to generalize should increase as parameters are subtracted from the model. Taking account of additional interaction effects across predictors would push model complexity in the opposite direction. Estimating deep interactions requires large numbers of parameters.

Findings from the neuroscience literature may help to resolve this apparent contradiction. In reviewing the determinants of mental and behavioral flexibility in \emph{Drosophila}, \cite{Devineni2022} describe a particular interaction effect that provides a helpful motivating example. Fruit flies have an innate aversion to the smell of carbon dioxide, which is a warning signal emitted by other flies. In situations of extreme hunger, however, the brain releases chemical signals that cause this response to shift. Flies seek out the odor, because it raises a possibility of food that outweighs the risk. Hence, the modulation of priorities across different contexts is handled by a system that operates at a global level and is distinct from the mechanisms that process innate and learned responses. In this case, the modulation is straightforward. Food has become a higher priority, and the chemical signals tilt decision-making to reflect this preference.

While the effects of context are not always so straightforward, this example from fruit flies highlights one way in which priorities can shift in response to context while keeping the essential structure of a decision-making process intact. Along similar lines, deep learning models could incorporate dramatic changes in the population of test cases through relatively simple responses. For instance, hyperparameters such as activation thresholds could be designed to depend upon case- or distribution-specific factors.

\section{Domain Generalization} \label{Domain}

In some situations, the nature of test cases encountered---or the behavior of the target function in response to those cases---differs so much from that experienced in training that the model must be modified in order to accommodate it. When generalizing from one distribution to another, the focus is on developing flexible models that can perform well across the diversity of cases observed in the population. When the changes are sufficiently extreme, however, it is no longer helpful to think about one large model that does everything. What we call generalization happens at a higher level of abstraction, where there is different version of the model for each domain.

The types of changes in domain that researchers consider can be divided into two broad categories. In the first, domain varies continuously, sometimes in ways that are not immediately apparent. The second form of domain change involves a shift to a distinct set of cases in a new environment. 

\subsection{Concept Drift} \label{Drift}

In the gradual form of domain change known as ``concept drift,'' It is not only the distribution of covariates that shifts, but there is some change in the ground truth target function mapping those covariates to labels.\footnote{As many authors note, varying definitions of concept drift are used in the literature. For the purposes of this discussion, a definition of gradual and potentially imperceptible domain change will be used.} Examples of this change include the distortion of images or sounds and the evolution of words' meanings in texts. Surveys of this form of domain change and strategies to address it can be found in \cite{Bayram2022, Li2019, Lu2019, Wang2018}, and \cite{Yang2021}.

A well-known early example of concept drift is Google Flu Trends. The project, first announced in 2008, produced a highly effective predictor of flu patterns in the United States based upon Google users' search activity. The performance of the model declined, however, in successive flu seasons as search activity changed, partly due to the rollout of a new ``autocomplete'' feature in Google Search that changed the likelihoods of different searches \citep{Lazer2014}.

\cite{Henriksson2021} note that, for some applications, performance degradation of this form poses safety concerns. The authors test a framework in which a supervisor monitors the classification probabilities generated by a trained neural network and alerts the user if they indicate an increase in the rate of outliers or a change in the distribution of test cases being evaluated. The authors find that the supervisor's ability to identify distributional changes generally improves with the training accuracy of the model but is inconsistent, varying widely across training epochs. Similarly, \cite{Rohlfs2023c} finds among a sampling of neural networks for object recognition that the models' self-assessments of classification confidence are effective proxies for accuracy among in-distribution test cases. When given low-resolution variants of those images, however, the classifiers' performance declines, and the models are overconfident in their classifications.

Seen from the lens of the bias-variance tradeoff in Subsection \ref{Bias}, this phenomenon through which the target function changes but parameters are fixed is a form of underfitting. Ants and birds whose patterns of behavior are locked down by evolution are insensitive to data they receive about their situation. Similarly, the model's training parameters have been locked down. The model's forecasts are still responsive to changes in the features associated with new test cases. It is not just the covariates that are changing, however. Like in the Google Flu case, the target function that maps covariates to output classes is changing. Because the model's parameters do not change, the model is unresponsive and therefore underfit with respect to this change in the target function.

In shallow learning contexts, \cite{Wang2018} find that, when domain change occurs in a continuous fashion, much of the degradation in performance reflects changes in the likelihoods of different classes and can effectively be addressed by monitoring those probabilities and adjusting the corresponding thresholds. Threshold-based adjustment of this form is not unlike the adjustment seen in fruit flies, discussed earlier, in which starvation causes the insect to modify its behaviors---for instance, the amount of stimulus required for a given action---to increase the likelihood of obtaining food.

\begin{figure}
\centering
\includegraphics[width=0.45\textwidth]{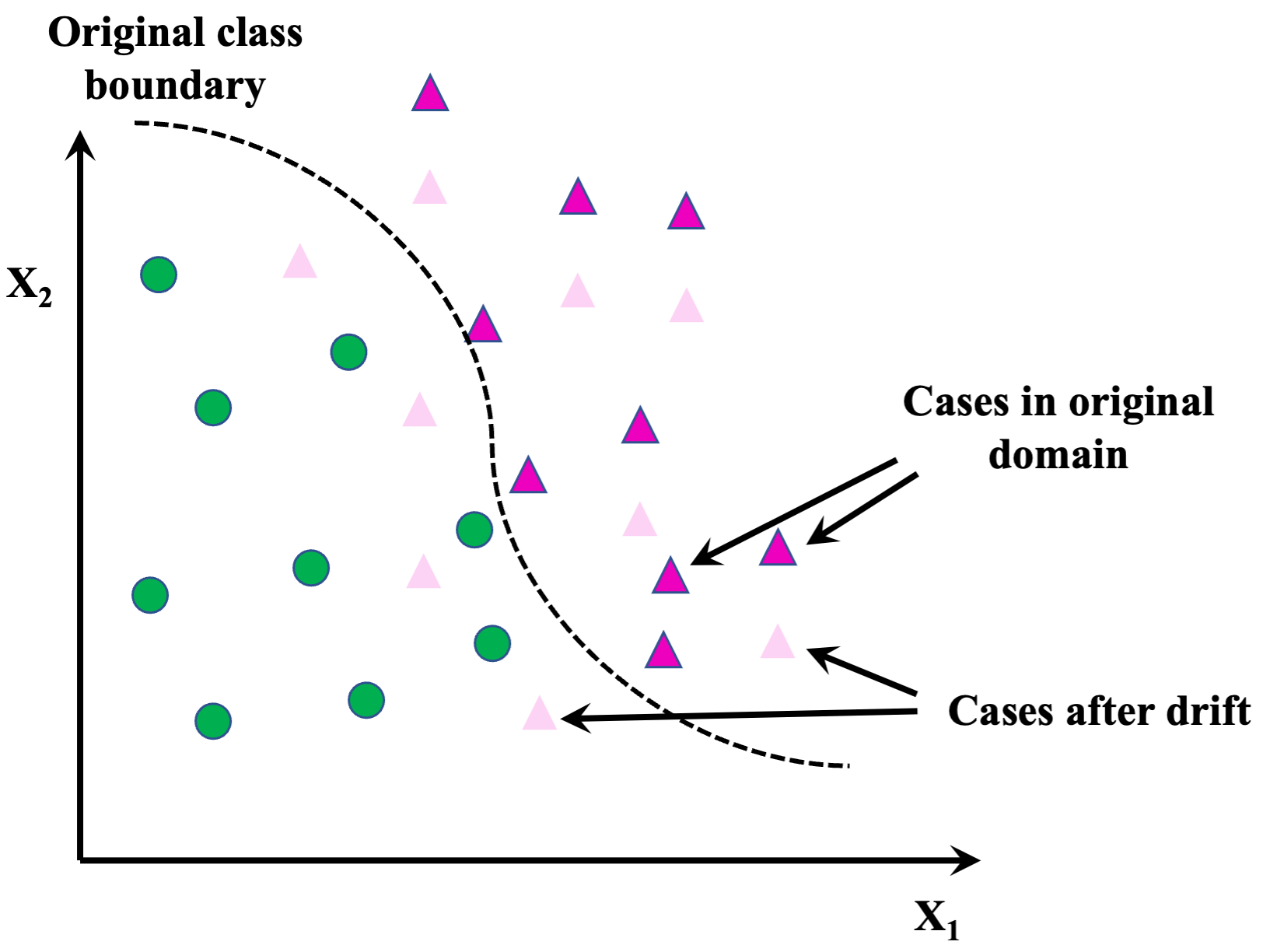}
\caption{Concept Drift}
\label{fig:drift}
\end{figure}

The phenomenon of changing thresholds is illustrated in Figure \ref{fig:drift}, which is adapted from \cite{Bayram2022}, who provide a useful review of studies of concept drift and its varying definitions in the literature. In the example, there is some boundary between the green circles and the violet triangles that was learned from training data. Drift occurs in that, among new test cases from the triangle class, the values of the second predictor $X_2$ tend to be lower than seen in training. In some cases of concept drift, researchers aim to determine when a substantive change in domain has occurred so that the model should be reconfigured before further use. Other drift-based approaches aim to develop classifiers that are robust to such continuous domain variation. \cite{Bayram2022} note that, when attempting to measure whether drift has occurred, performance-based tests of changes in distribution and domain have a tendency to raise false alarms, because performance may degrade for reasons such as data quality that would not be solved through reconfigurations to the model.\footnote{In a related survey, \cite{Yang2021} discuss the problem of detecting if a case falls outside of the trained distribution. Similarly to \cite{Bayram2022}, the authors find a diversity of terminology and propose a unified framework for considering such situations.}

\begin{figure}
\centering
% include first image
\includegraphics[width=0.3\textwidth]{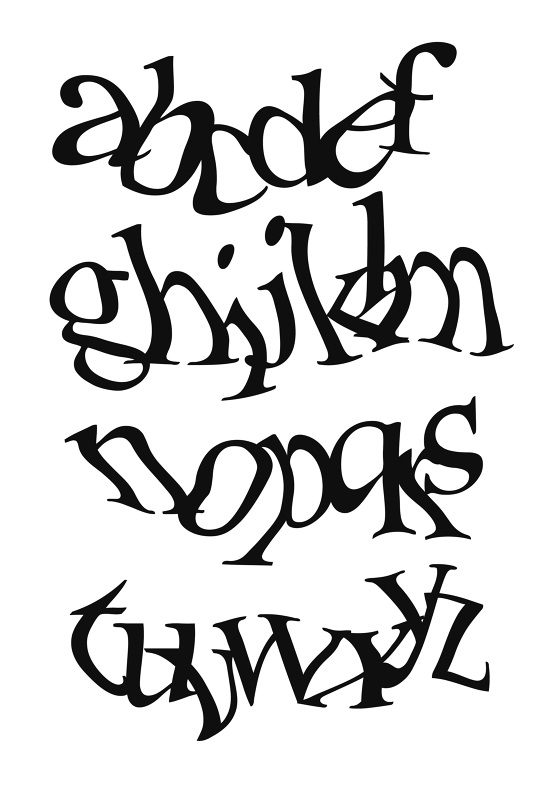}
\caption{CAPTCHA-type Characters}
\label{fig:captcha}
\end{figure}

\cite{Xiang2023} and \cite{Yuan2022} provide useful surveys of strategies through which neural networks have been adapted to accommodate changes in concept definitions. In time-series contexts, some researchers partially or fully refit model parameters on a regular basis. This process of putting greater weight on recent observations and sometimes discarding earlier cases helps to prevent learned systems from using stale concepts. As \cite{Yuan2022} also note, researchers have also found ways to adjust structural elements to accommodate concept drift, adjusting the width \cite{Kauschke2019} and depth \cite{Pratama2019} in response to rising rates of misclassification.

In visual object-recognition problems, rotated or distorted images are examples of concept drift. The changes in features could be subtle enough that they do not impact the labels that a human would apply to the cases---so that the true input-output relationship would be unaffected. As with generalization across distributions, however, a classifier trained on undistorted data would have difficulty accurately classifying the new cases.

Completely Automated Public Turing test to tell Computers and Humans Apart (CAPTCHA) problems constitute one everyday setting to which researchers have applied tools to recognize and adapt to concept drift. These tests check to ensure that a computer user is human by present the user with a text or image recognition problem that would cause difficulties for an automated system. CAPTCHA tests that are used to secure computer systems against automated attacks constitute a compelling category of problems---and one that has been studied extensively---in which context and feature distributions vary continuously. Many CAPTCHAs rely upon the recognition of abstract features---\emph{e.g.}, distinguishing letters that are deformed, overlapping, or have distracting elements such as noise or ancillary lines. A set of characters written in this deformed and overlapping style is illustrated in Figure \ref{fig:captcha}. The image is taken from a CAPTCHA-inspired font by \cite{Rogers2014}. Optical character recognition (OCR) approaches typically distinguish text in a sequential manner, first determining boundaries to isolate individual characters and then determining which letter or number each of those characters represents. While it does not cause great difficulty for humans, the overlapping of characters in the image confounds the automated approach.

The wide variety of CAPTCHA-style problems available provides the deep learning community with an extensive amount of human-labeled data to illustrate what features define different classes of objects and what features can be altered without causing their labels to change. \cite{Xu2020} review a variety of deep learning approaches to construct and to decrypt such problems. The decryption strategies---which the authors refer to as ``attacks'' on CAPTCHA problems---generally employ human assistance to reframe specific CAPTCHA tasks into formats that are tractable by existing classification models. For example, \cite{George2017} replace the standard sequential approach to OCR with an iterative one in which character boundaries and what letters or numbers they represent are determined in a coordinated fashion. The feature space is expanded to include contours of the entire sequence of letters rather than individual characters, and having access to the contours enables the classifier to decipher sequences of text are overlapping or distorted.

\subsection{Observable Shift} \label{Observable}

\begin{table*}[b!]
\renewcommand{\arraystretch}{1.25}
\centering
\resizebox{\textwidth}{!}{
\begin{tabular}{p{0.06\textwidth}p{0.36\textwidth}p{0.43\textwidth}p{0.43\textwidth}}
Medium & Dataset & Classes & Domains \\
\hline
& DomainNet \citep{Peng2019} & 345 categories of everyday items & Clipart, infographic, painting, quick drawing, photo, and sketch \\
& PACS \citep{Li2017a} & Dog, elephant, giraffe, guitar, house, horse, person & Art painting, cartoon, photo, and sketch \\
& VLCS \citep{Fang2013} & Bird, car, chair, dog, person & Four different source datasets of photos \\
& ImageCLEF-DA \citep{Caputo2014} & Plane, bike, bird, boat, wine, bus, car, dog, horse, motorcycle, computer, person & Three different source datasets of photos \\
Images & VisDA2017 \citep{Peng2017} & Plane, bike, bus, car, horse, knife, motorcycle, person, plant, skateboard, train, truck & Synthetic and real-world images \\
& Office-Home \citep{Venkateswara2017} & 65 classes of everyday items & Art, clipart, product, and real world \\
& Office-31 \citep{Saenko2010} & 31 classes of office objects & Amazon product pictures, Webcam photos, DLSR photos \\
& Caltech Camera Traps \citep{Beery2018} & 17 types of animals, car, or no object & 140 fixed camera locations \\
\hline
 & Reuters-21578 \citep{Lewis1997} & Newswire article subjects & Categories of article subjects: Exchanges, Organizations, People, Places, Other \\
Texts & Enron Emails \citep{Klimt2004} & User-defined folders & Email users \\
& Multi-domain Sentiment Dataset \citep{Blitzer2007} & Amazon product ratings based on review texts & Books, Kitchen, Electronics, and DVDs \\
\hline
\end{tabular}}
\caption{Datasets commonly used for Learning Domain Generalization for Classification Tasks}
\label{tab:adaptation}
\end{table*}

In the second form of domain change, when there is a dramatic and observable shift, many features of the input data distribution as well as the target function are dramatically altered. Sufficiently many key elements of the input-output relationship are retained, however, for generalization from the old domain to be possible. Examples could include a change in the identity of the speaker for a speech recognition model or the application of a text processing tool to a foreign language. Importantly, the researcher does not have access to labeled training data for the new domain. To classify the new test cases, it is therefore necessary for the model to learn how to generalize across domains.

\begin{figure}
\centering
\includegraphics[width=0.45\textwidth]{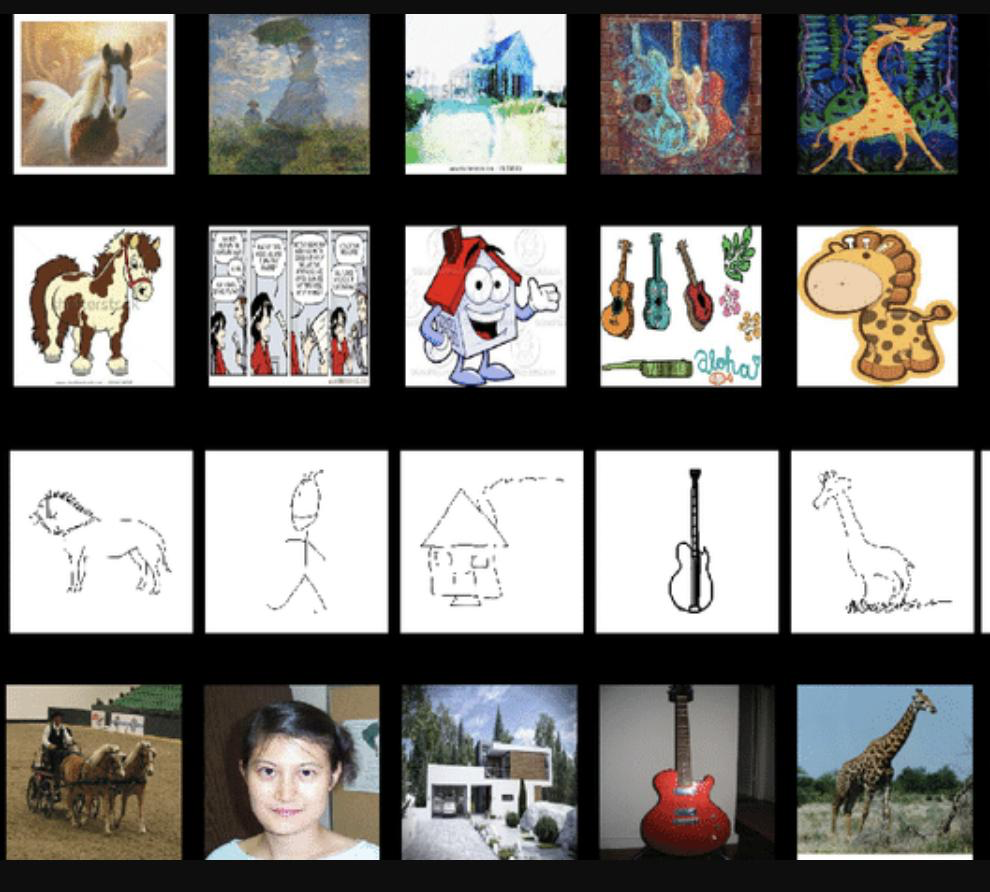}
\caption{Illustration of images in the PACS dataset for Domain Generalization}
\label{fig:pacs}
\end{figure}

Figure \ref{fig:pacs}, reproduced from \cite{Xu2019}, illustrates one compelling dataset that researchers have used to train neural networks to perform this type of generalization. The images in the PACS data represent objects from seven classes including horse, person, house, guitar, and giraffe, as depicted in the figure, as well as dog and elephant. The four rows of images illustrate the four domains in which these different classes of objects are represented: art paintings, cartoons, photos, and sketches. As the example images illustrate, each class has some distinctive traits that appear with some consistency across domains, such as the spots and neck on a giraffe and the A-shaped roof of a house. The exact ways in which those traits appear on the images varies considerably, however, across the domains. Each domain has its own complex, nonlinear relationship through which features map to these seven classes. For the purposes of learning domain generalization, a neural network is trained to identify these different object types using data from three of the four domains, so that it has experience not only with the classes but in how their definitions tend to vary from one domain to the next. It is then tested on its ability to perform the same task in the new domain.

Table \ref{tab:adaptation} describes a sampling of this variety of datasets used for developing domain generalizing models. The four columns describe the input type (images or texts), the dataset name and source, the classes that are learned within a domain, and the different domains in which these classes are defined. The first eight datasets describe object identification problems that are similar in structure to that posed in PACS. The numbers of classes range from five to 345. Some of the datasets like PACS have diverse sets of domains to learn, while in other cases, the domains are relatively similar, consisting of photographs taken from different image libraries. The text-based datasets present tasks of identifying the subjects of newswire articles (with domains consisting of different broad categories of subjects), sorting emails into folders for different users, and projecting the consumer rating based upon the review text for different types of Amazon products.

\subsubsection{Transfer Learning} \label{Transfer}

\begin{figure*}
\begin{subfigure}{0.40\textwidth}
  \centering
  % include first image
  \includegraphics[width=\textwidth]{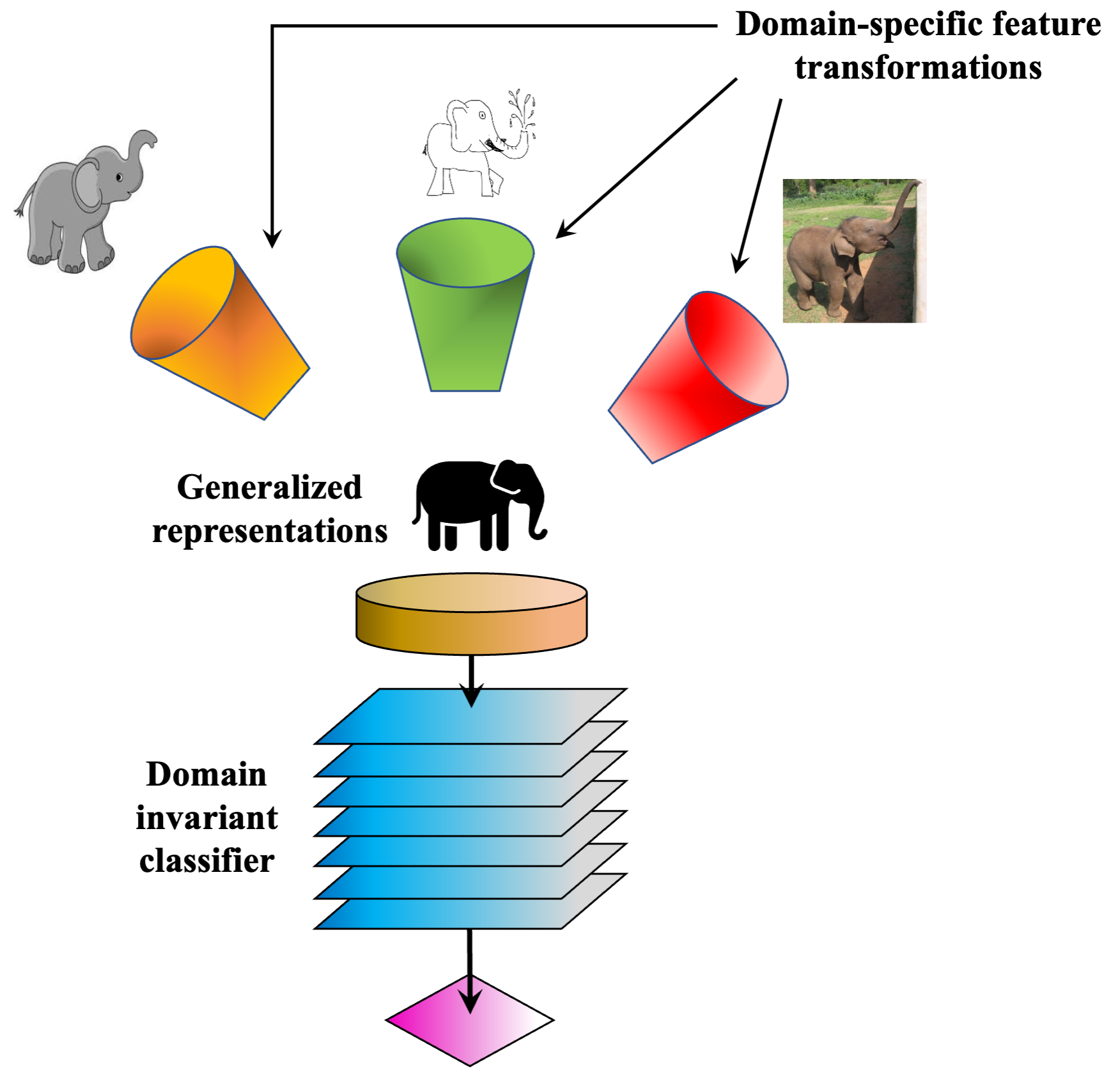}
  \vspace{0.25 in}
  \caption{Feature Mapping}
  \label{fig:features}
\end{subfigure}
\hfill
\begin{subfigure}{0.50\textwidth}
  \centering
  % include first image
  \includegraphics[width=\textwidth]{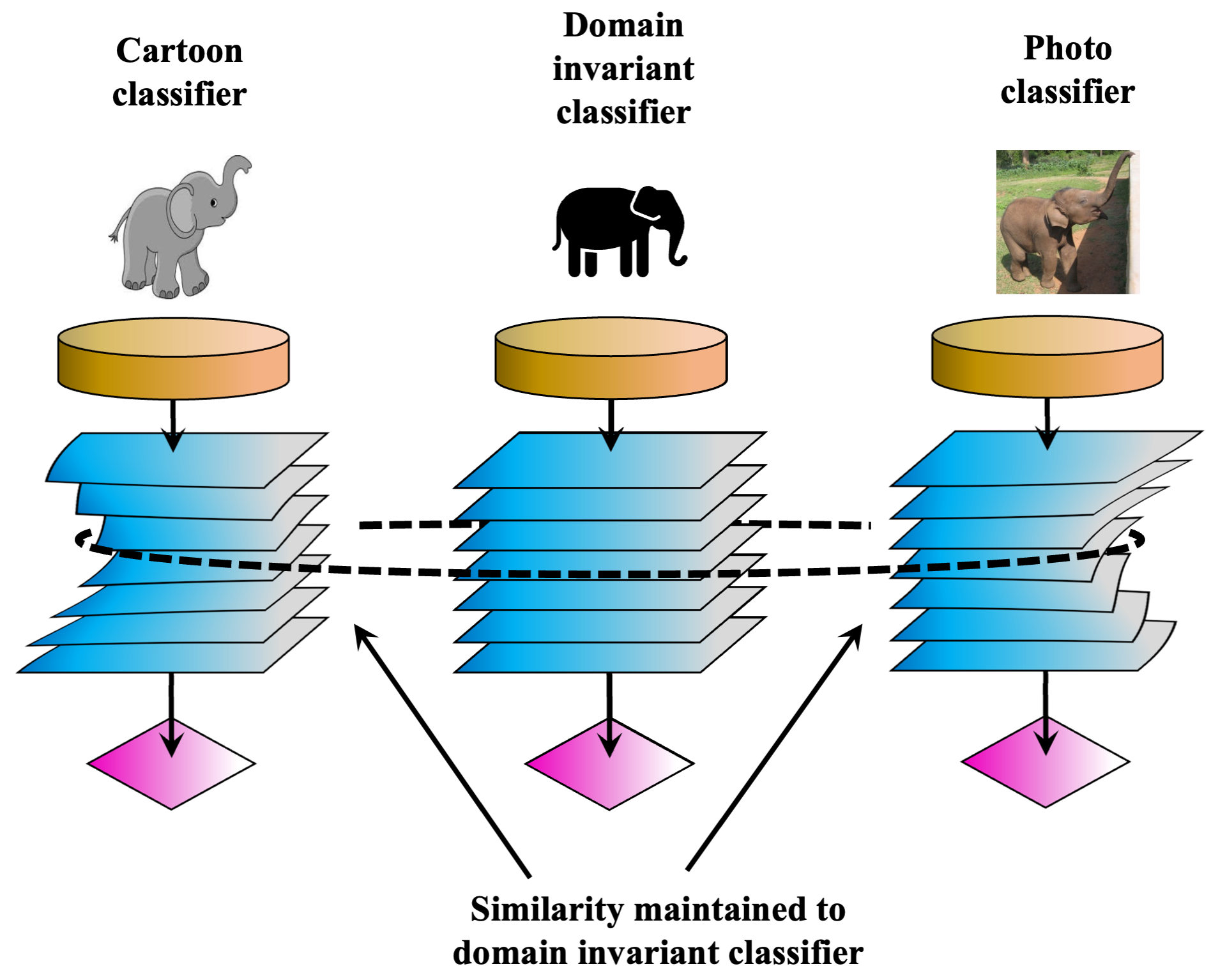}
  \caption{Parameter Sharing}
  \label{fig:sharing}
\end{subfigure}
\caption{Illustration of Forms of Transfer Learning}
\label{fig:inductive}
\end{figure*}

One popular and effective strategy for developing models to generalize across discrete changes in domain is \emph{transfer learning}. Through this approach, a neural network is designed to \emph{adapt} knowledge acquired in one domain and apply it to another. Table \ref{tab:transfer} summarizes the variety of transfer learning strategies that have been used in the literature. For greater detail on these approaches and the variety of problems to which they have been applied, \cite{Gulrajani2020,Niu2020,Wang2021,Wang2023,Zhou2022}, and \cite{Zhuang2021} provide extensive and useful reviews.\footnote{Transfer learning is discussed in this section on domain generalization because that is the form of generalization with which the method is most closely associated. Even so, any of the differences across distributions discussed in Section \ref{Distribution} could alternatively be thought of as changes in domain, so that the strategies described here could also be used for distribution generalization. Additionally, some of the task generalization approaches described in Section \ref{Task} are variations on transfer learning.}

Transferring network-based learning across distinct domains takes one of two conceptual approaches. In \emph{transductive} domain adaptation, there is explicit translation from one domain to another. Such strategies require some knowledge of the new domain at the time that the model is developed. Typically, there is a dataset of unlabeled cases from the new domain that is used to determine how the patterns of features differ between the old and the new setting. \emph{Inductive} models, on the other hand, are developed for use in any arbitrary domain. The researcher supposes that there is a core, domain-invariant model that represents the archetypal versions of the classes being modeled. Domain-specific training data are used to learn both about the particular patterns exhibited in that domain and also about the core model. Inductive transfer learning can use unlabeled feature data from the new domain during model development, but it is not required. \cite{Gulrajani2020} provide a brief survey of approaches and datasets for this form of domain generalization, such as those used by the original \cite{Li2017a} on PACS, and the authors provide a set of associated algorithms and datasets in their DomainBed package for use with PyTorch \citep{Paszke2019}.

\begin{table}[ht!]
\renewcommand{\arraystretch}{1.25}
\centering
\resizebox{0.45\textwidth}{!}{
\begin{tabular}{p{0.16\textwidth}p{0.32\textwidth}}
Strategy & Description \\
\hline
\hline
Feature Mapping & Input features transformed to feed into domain-shifted or domain-agnostic model \\
Parameter Sharing & Domain-specific models with sharing or restrictions across parameters \\
Ensemble Weighting & Aggregation of model outputs across domains \\
\hline
\end{tabular}}
\caption{Transfer Learning Strategies}
\label{tab:transfer}
\end{table}

The first of the strategies from Table \ref{tab:transfer}, feature mapping, can transductive or inductive. An inductive variant is illustrated in Figure \ref{fig:features}. In feature mapping approaches, there is a learned transformation function that processes input features from a given domain so that they can be consumed by a classifier from another domain or a domain invariant classifier, as in the diagram. The top of the figure shows pictures of elephants from three domains: cartoons, sketches, and photographs, all from the PACS data. Each domain has its own feature transformation function that converts the input into the format expected by the classifier. The transformation often expands the input space, so that the generalized representation shown in the diagram is not necessarily viewable as an image. All domain-specific computation is performed on the features, and the remaining components of the classifier are shared across two or more domains. In some cases, researchers use feature transformation to supplement the training process for the domain-specific classifiers. \cite{Shankar2018}, for instance, employ generative adversarial learning in which transformed cases from other domains are used to create adversarial examples to refine the classification and domain-identification components of the network.

\begin{table*}[b!]
\renewcommand{\arraystretch}{1.25}
\centering
\resizebox{\textwidth}{!}{
\begin{tabular}{p{0.18\textwidth}p{0.17\textwidth}p{0.18\textwidth}p{0.16\textwidth}p{0.09\textwidth}p{0.07\textwidth}p{0.07\textwidth}p{0.07\textwidth}p{0.07\textwidth}}
\multirow{2}{*}{Model} & \multirow{2}{*}{Source} & \multirow{2}{*}{Backbone} & \multirow{2}{*}{Strategy Type} & \multicolumn{5}{c}{Unseen Domain} \\
 & & & & Art Painting & Cartoon & Photo & Sketch & Average \\
\hline
\hline
SIMPLE+ & \cite{Li2022b} & Multiple & Feature Mapping & & & & & 99.0\% \\
PromptStyler & \cite{Cho2023} & CLIP, ViT-L/14 & Feature Mapping & & & & & 98.6\% \\
GMDG & \cite{Tan2024} & RegNetY-16GF, SWAD & Parameter Sharing & & & & & 97.9\% \\
D-Triplet & \cite{Guo2024} & RegNetY-16GF & Feature Mapping & 98.3\% & 98.5\% & 99.9\% & 93.8\% & 97.6\% \\
Mixture-of-Adapters & \cite{Lee2023} & OpenCLIP, ViT-B/16 & Parameter Sharing & & & & & 97.4\% \\
MIRO & \cite{Cha2022} & RegNetY-16GF, SWAD & Feature Mapping & & & & & 96.8\% \\
CAR-FT & \cite{Mao2022a} & CLIP, ViT-B/16 & Parameter Sharing & & & & & 96.8\% \\
VL2V-SD & \cite{Addepalli2024} & CLIP, ViT-B/16 & Feature Mapping & & & & & 96.7\% \\
UniDG & \cite{Zhang2023} & CLIP, ViT-B/16 & Feature \& Parameter & & & & & 96.7\% \\
SLEDGE+ & \cite{Li2022b} & Multiple & Feature Mapping & & & & & 96.1\% \\
Ensemble of Averages & \cite{Arpit2022} & RegNetY-16GF & Ensemble Weighting & 94.1\% & 96.3\% & 99.5\% & 93.3\% & 95.8\% \\
UniDG & \cite{Zhang2023} & ConvNeXt & Feature \& Parameter & 98.2\% & 90.2\% & 99.9\% & 92.9\% & 95.3\% \\
PCL & \cite{Yao2022} & ResNet-50, SWAD & Feature Mapping & 90.2\% & 83.9\% & 98.1\% & 82.6\% & 88.7\% \\
TF & \cite{Li2017a} & Alexnet & Feature Mapping & 62.9\% & 67.0\% & 89.5\% & 57.5\% & 69.2\% \\
\hline
\end{tabular}}
\caption{Performance of Models on PACS Domain Generalization}
\label{tab:pacs}
\end{table*}

The next domain generalization strategy listed in Table \ref{tab:transfer}, parameter sharing, is illustrated in Figure \ref{fig:sharing}. As with Figure \ref{fig:features}, the example presented in the graph involves inductive transfer. As Figure \ref{fig:sharing} illustrates, the parameter sharing approach involves a separate network-based classifier for each domain. Optimization proceeds in a simultaneous fashion, with restrictions applied to the parameters to ensure that the domain-specific classifiers are sufficiently similar to the domain invariant one. This approach can also be applied transductively, in which case there is no domain invariant model, but parameter restrictions are applied to maintain similarity between the domain-specific models.

In the final generalization approach from Table \ref{tab:transfer}, ensemble weighting, multiple neural networks are developed during the training process: one for each of the domains that is available during training, with no domain invariant model. When test cases are observed from the unseen domain, they are run through each of these trained models, some decision rule such as majority rule or averaging of probabilities is used to aggregate these domain-specific projections into a final classification.

\subsubsection{Empirical Results} \label{Empirical Transfer}

Next, we examine how some of these strategies for generalization have performed in the domain generalization exercise from the PACS data. Table \ref{tab:pacs} lists the names, attributes, and performance statistics for several recent models together with the original PACS study from \cite{Li2017a}. The best-performing specification is presented for each model, with the model names, specification details, and performance statistics matching those presented in the website \cite{PapersWithCode2024}. Each specification performs four domain generalization exercises in which one of art painting, cartoon, photo, or sketch is the unseen target domain, and the other three are the source domains. Because there are multiple source domains, each of the specifications here uses inductive transfer and a domain-invariant classifier from which to generalize to the unseen domain. For all of the models, the average accuracy across the four unseen domains is shown in the table. For papers that report the results of the four separate generalization exercises separately, those breakouts are also shown. The two separate entries for the UniDG model include the best-performing specification overall as well as the best-performing specification for which this breakout is reported.

Each of these specifications shown in Table \ref{tab:pacs} has a model name as well as a ``backbone'' presented. The backbone refers to an established \emph{foundation model} whose architecture and trained parameters are used as a starting point. Some of these models, including AlexNet \cite{Krizhevsky2014}, ConvNeXt \cite{Liu2022}, RegNetY-16GF \cite{Radosavovic2020}, and ResNet-50 \cite{He2016} are trained using ImageNet-V2 \citep{Russakovsky2015}, the popular object-recognition data discussed in Subsection \ref{ImageNet}. Those for which ``Multiple'' is listed as a backbone use more than one foundational model built off of those data. For some of those backbones, the Stochastic Weight Averaging Densely (SWAD) regularization technique from \cite{Cha2021} is used. Some other of these models use as a foundation model the multimodal Contrastive Language-Image Pre-Training (CLIP) approach of \cite{Radford2021} and its OpenCLIP variant. Both supplement image data with text in order to improve classification. The underlying network architectures used for these models are variants of the Vision Transformer (ViT) set of models from \cite{Dosovitskiy2020}.

In addition to backbone, a ``strategy type'' is listed for each model. While each of study introduces its own model-specific innovations, this field in the table indicates which of the general categories of transfer learning approaches from Table \ref{tab:transfer} each one use. UniDG is labeled ``Feature and Parameter'' as that model combines elements of both feature mapping and parameter sharing. Due to the variety of foundational models and datasets used, the results presented in the table should not be construed as a direct comparison of the effectiveness of different generalization strategies. They should instead be regarded as an illustration of the different approaches used and how improvements in many dimensions have enhanced neural networks' ability to generalize across domains.

As the last column from Table \ref{tab:pacs} indicates, the ability to generalize across visual domains has improved considerably over the time that researchers have been working on the problem. While the initial \cite{Li2017a} study only achieved 69.2\% accuracy on average, recent approaches have reached as high as 99.0\%. Among the three categories of approaches, feature mapping is the most commonly represented among the high-performing models, but strong performance can be seen from parameter sharing and ensemble weighting as well.

For the models that report breakdowns for their results, generalization is the consistently the strongest when ``photo'' is the unseen domain. This finding is relatively unsurprising, given that the foundation models are built on photographic data. Performance tends to be worst on the ``sketch'' category, whose images tend to be the most abstract and to provide the least visual information.

\section{Task Generalization} \label{Task}

Moving further to the right in the specificity-generality continuum in Figure \ref{fig:generality}, the next higher level of abstraction is generalization across tasks. The distributions and domains may still be changing, but that is not necessary. The defining element of this form of generalization is the output classes or specific question being answered is different from in the original model. It is thus absolutely necessary to reconfigure the model, because the format of the outputs has changed.

This section begins with a discussion of early work to train a neural network to perform a new set of skills while retaining previously acquired knowledge. It then discusses advances in the use of backbone models that, having learned how to perform a general set of tasks, can learn new variations with a small number of training examples, a process known as \emph{few-shot learning}. We then explore recent innovations in text-based deep learning, where the \emph{transformer} architecture, another backbone-driven strategy, has helped researchers to produce language-based knowledge that is highly portable across tasks. While few-shot learning only expanded the set of outcome classes, these language tasks are diverse, including different types of questions and output formats. 
We then discuss how this growing ability to generalize knowledge across tasks has led to the rise of widely-used \emph{transformer-based foundation models}, with increased consolidation and centralization of network-based learning for core functions.

\subsection{Catastrophic Forgetting} \label{Catastrophic}

In biological systems, learning is continuous and online, so that encountering new data can lead decision rules to change \citep{Sinz2019}. At the same time, stale information is steadily discarded through the pruning of unused synapses, but it is done slowly to maintain stability and manage risk \citep{Akers2014,Richards2017}. Researchers working with artificial systems face this same challenge of balancing new and existing skills.

The most common form of task generalization considered in the literature is to train a classifier to identify one or more new categories. Early work approached the problem from the perspective of expanding the possible outcomes for an existing trained classifier. Researchers observed a phenomenon of \emph{catastrophic forgetting}, whereby acquiring a new skill reduced the system's ability at previously learned tasks. \cite{Vandenhende2021} and \cite{Zhang2022} survey a variety of approaches for adapting networks' structures and other design elements to handle multiple tasks without such difficulties. A few are discussed here.

Strategies to avoid catastrophic forgetting have mainly involved modifications to the optimization problem used to fit the parameters of the neural network, including Progressive Learning (PL), elastic weight consolidation (EWC), and synaptic intelligence (SI), all of which have been effective in some applications but come with costs. In PL, a subset of the learned architecture from prior cases is maintained and repeatedly consulted for comparison when new cases are encountered \citep{Rusu2016a, Rusu2016b}. This approach has proven to be effective in the completion of mazes and video game tasks. It lacks scalability, however, as it expands the size of the network and the amount of time and resources required with each new training observation. \cite{Fayek2020} partially address this limitation with a pruning procedure that regularly removes the previously trained weights whose estimated magnitudes fall below some threshold.

EWC and SI both vary the speed at which parameter updates occur depending upon their relevance to previously learned tasks. EWC does so through Bayesian updating with task-specific constraints added at a global level to the optimization problem to account for different parameters' impacts on performance in other tasks. A more localized modification of the loss function is performed in SI. Associated with each connection is both a weight and a degree of importance of that parameter in previously learned tasks, where higher importance increases the cost of updating the weight. Researchers have found both approaches to reduce the problem of catastrophic forgetting while maintaining high performance at multiple object identification tasks or video games \citep{Kirkpatrick2017, Liu2020, Zenke2017}. Like PL, however, the strategies introduce new problems related to cost and scalability. Under both EWC and SI, as the system learns additional tasks, it becomes increasingly constrained and slow to learn \citep{Schwarz2018}. All three methodologies address the problem of forgetting skills, and work along these lines has contributed to progress in knowledge distillation \citep{Gou2021}, but the networks are difficult to scale due to their continually growing demands for computation and storage.

\begin{figure*}
\begin{subfigure}{0.20\textwidth}
  \centering
  % include first image
  \includegraphics[width=\textwidth]{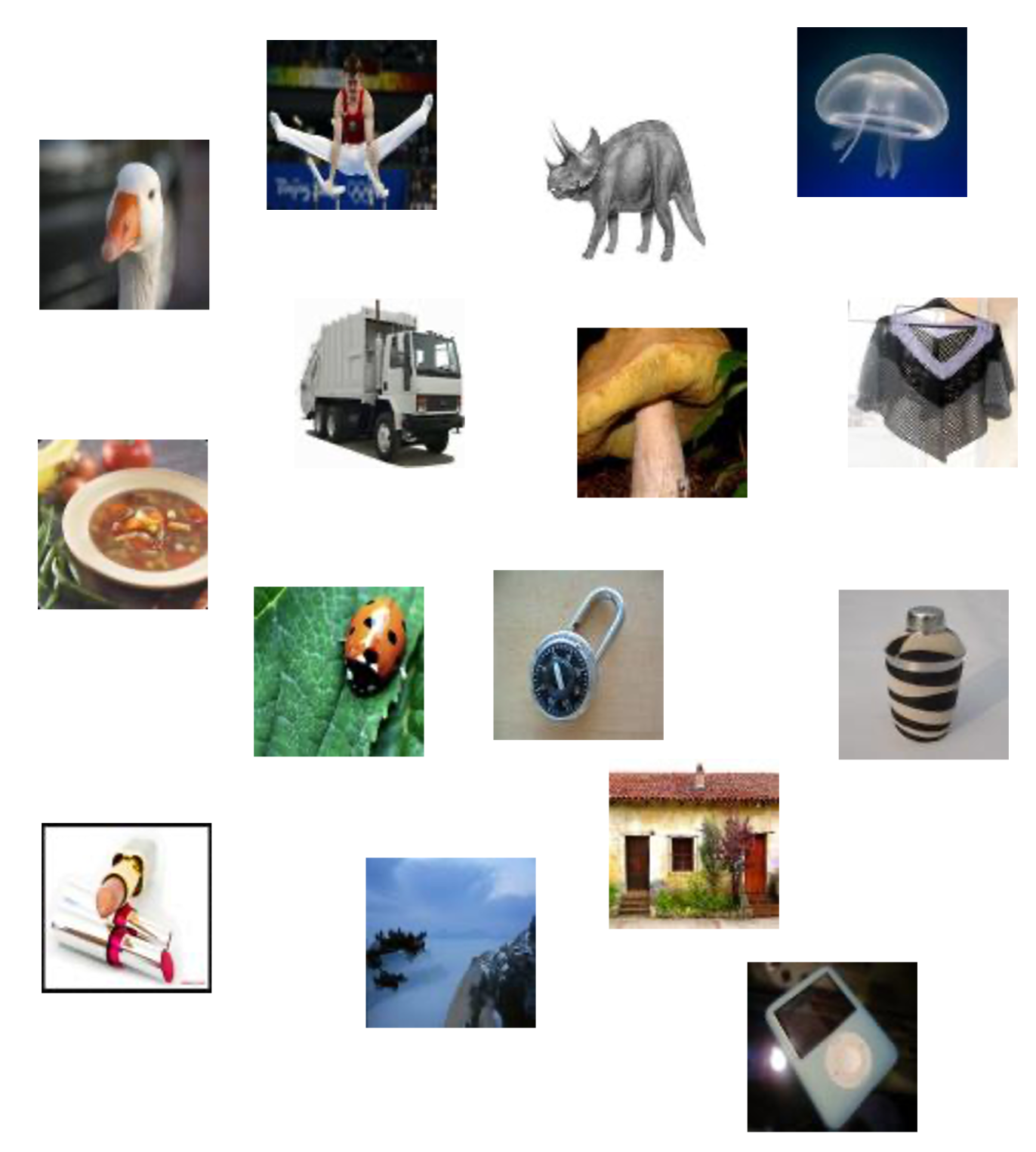}
  \caption{Training Data}
  \label{fig:training}
\end{subfigure}
\rulesep
\begin{subfigure}{0.38\textwidth}
  \centering
  % include first image
  \includegraphics[width=\textwidth]{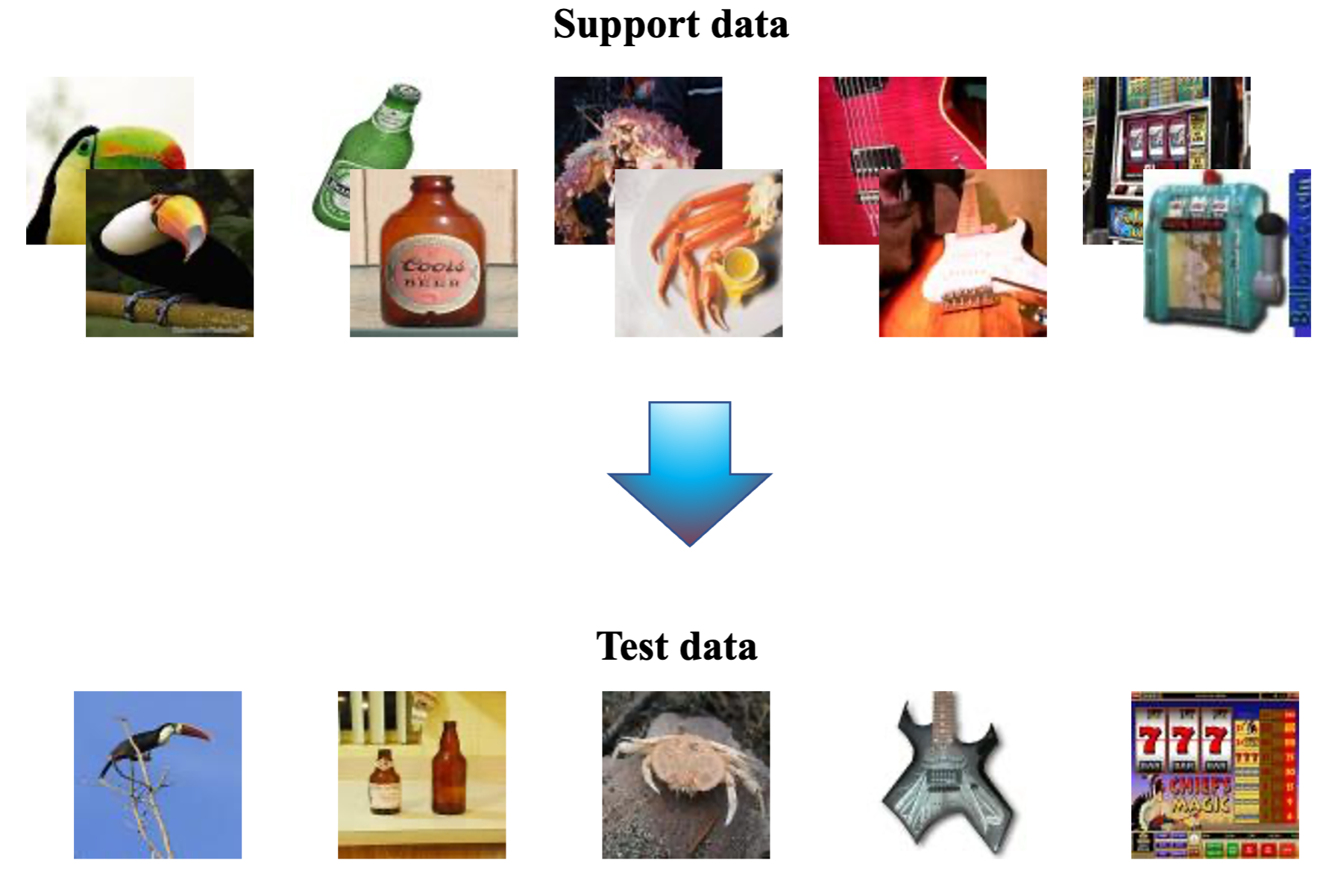}
  \caption{Easy Task}
  \label{fig:easy}
\end{subfigure}
\rulesep
\begin{subfigure}{0.38\textwidth}
  \centering
  % include first image
  \includegraphics[width=\textwidth]{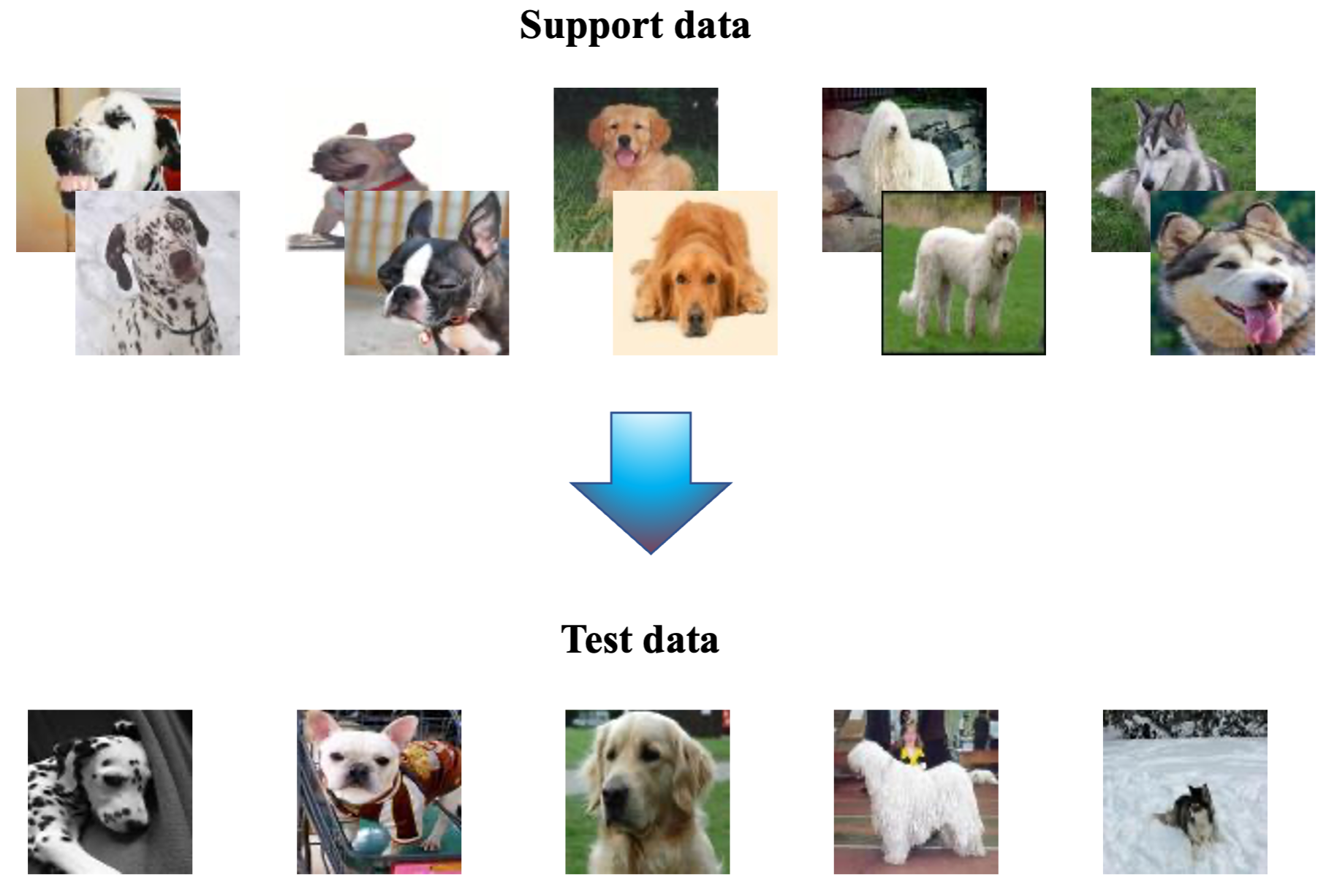}
  \caption{Hard Task}
  \label{fig:hard}
\end{subfigure}
\caption{Few-Shot Learning Problems}
\label{fig:fewshot}
\end{figure*}

Some recent progress toward stabilizing deep learning networks aims to maintain scalability by addressing catastrophic forgetting without repeatedly reviewing training cases. A biologically-inspired approach by \cite{Pandit2020} is patterned after the ``neurogenesis'' that regulates memory and forgetting in animal brains \citep{Akers2014}. The approach, like SI, introduces local modifications to the network's loss function and updating rules. The neural network that the authors design handles memory and storage through the gradual replication of commonly used neurons and connections and the termination of unnecessary ones. In a Bayesian-style approach, \cite{Teh2017} train a modular RL network whose components are coordinated by a central policy. Worker modules are assigned distinct tasks and solve their optimization problems separately, but each acts in reference to the common policy and considers closeness to this policy as one of its objectives. Each task's optimal solution in turn contributes to the determination of the general policy. The authors test this approach on multi-room maze and laser tag problems and find that it performs effectively and that its solutions are robust to small and moderate-sized changes in hyperparameter values.

As \cite{Vandenhende2021} and \cite{Zhang2022} discuss, some fruitful strategies for generalizing across tasks employ approaches similar to the policy coordination approach of \cite{Teh2017} through the use of transfer learning. In a feature-mapping variant, task-specific models dedicated to the same task share a common core of network layers that serve as a ``trunk,'' potentially separate feature transformers (if their inputs vary), and task-specific output layers or ``heads.'' In cases of computer vision, separate heads might exist for object detection, semantic segmentation of images, and object recognition. Alternate parameter-sharing transfer-based strategies allow each task-specific model to have its own trunk but with constraints that maintain similarity across the heads. These transfer approaches can be seen as variations on the transfer strategies described in Figure \ref{fig:inductive} for domain generalization, but with distinct output heads for different tasks. The growth of transfer-based strategies such as these, based upon foundation models as backbones, has emerged as a key means through which existing models can be trained to perform new tasks without degrading earlier performance. The next two subsections examine powerful variations on this approach---first to recognize objects with very few examples and then to learn varied natural language tasks.

\subsection{Few-Shot Learning} \label{Few-Shot}

\begin{figure*}
\begin{subfigure}{0.30\textwidth}
  \centering
  % include first image
  \includegraphics[width=\textwidth]{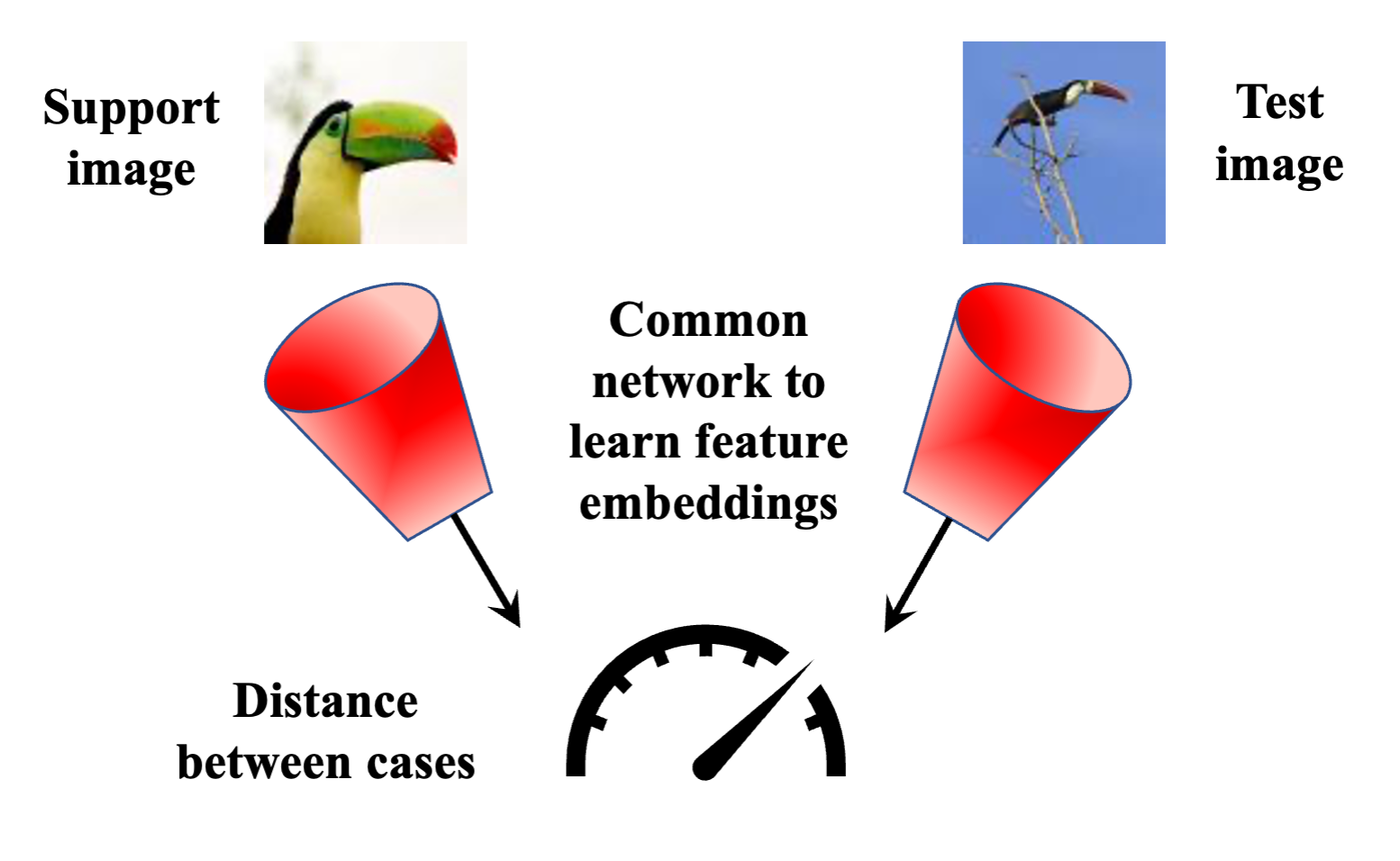}
  \vspace{0.25 in}
  \caption{Distance to Embeddings}
  \label{fig:distance}
\end{subfigure}
\rulesep
\begin{subfigure}{0.6\textwidth}
  \centering
  % include first image
  \includegraphics[width=\textwidth]{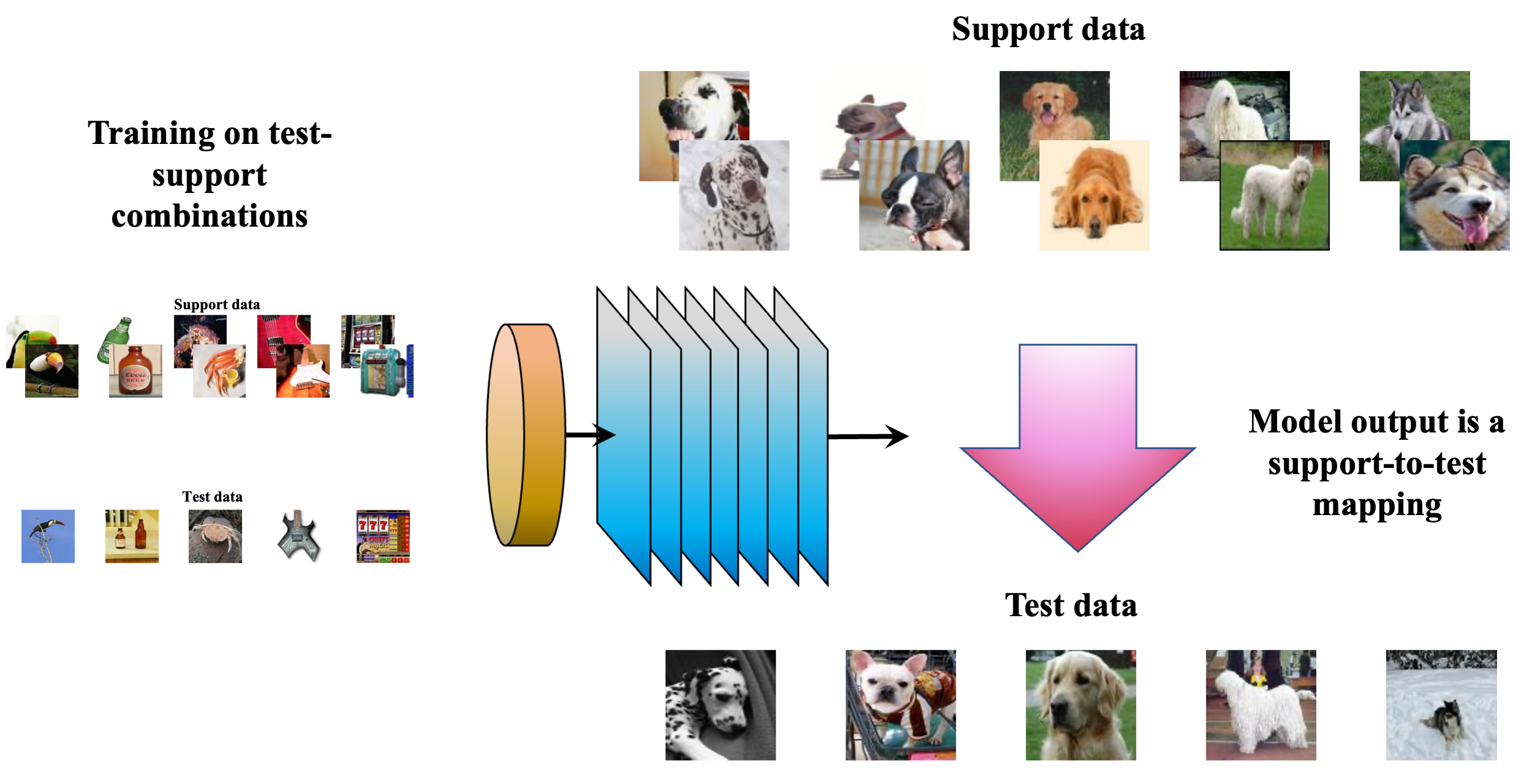}
  \caption{Meta-Learning}
  \label{fig:meta-learning}
\end{subfigure}
\caption{Models for Few-Shot Learning}
\label{fig:fewshot-models}
\end{figure*}

Exciting progress has been seen recently in a particular form of task generalization known as \emph{few-shot learning}, a setup that is illustrated in Figure \ref{fig:fewshot}. The images in the figure are taken from the mini-ImageNet dataset and the structure of the two problems aligns with tests performed by \cite{Vinyals2016}. The network has access to a large number of training cases of different types. Examples are presented in Figure \ref{fig:training}. The images shown in the figure are all from different classes, but the data include multiple examples within each category. During the training period, there is a holdout set of classes, and the model is not allowed to view any of the cases from those classes. In testing, the model is given a set of new classes to distinguish. Figure \ref{fig:easy} shows an example of a relatively easy five-way discrimination task in which the five classes are chosen from the dataset at large. The model is exposed to a small number of examples from each class (two in the example) through a process described as \emph{tuning}, and it is then evaluated in testing on its ability to distinguish images from the five categories. Figure \ref{fig:hard} shows a more difficult task in which all five classes are descended from the broader set of dog breeds, so that object identification requires a finer level of distinction.

Some approaches to few-shot learning transfer skills from networks trained to perform more traditional image classification problems. \citep{Vinyals2016}, for instance, learn feature embeddings from ImageNet to enable distance-type comparisons of test versus support images. In these models, much of the network-based learning is of embeddings to represent the features from individual images. Discrimination across classes is performed by applying a metric with which to compute the distances between each of the test images and the support images. Features of the distance metric may be learned in the training process, but the training images do not include elements from the support classes considered in testing. This distance-to-embedding approach is illustrated in Figure \ref{fig:distance}. A common network is used to transform image features into the embeddings consumed by a pairwise distance function which is used to evaluate image similarity between support and test cases. 

Another popular strategy of \emph{meta-learning}, as in the Model-Agnostic Meta-Learning (MAML) approach of \cite{Finn2017}, which builds on the approach of \cite{Andrychowicz2016}, replicates in training the tuning and evaluation processes performed in testing using a variety of support and test sets constructed from a training set of images. What is learned by the network is a generalization approach, as illustrated by the pink arrow. That approach itself consumes labeled support images as inputs and produces labels for test images as outputs. Competing generalization approaches are evaluated on the accuracy of their label assignments for the test cases. In testing and deployment, labeled support images and unlabeled test images are supplied as inputs, and the output is a mapping that labels the test images. As Figure \ref{fig:meta-learning} illustrates, training cases for this model consist of support-test combinations, and the output is a learner that can be tuned on new support cases. Approaches to few-shot learning are described in greater detail in surveys by \cite{Li2021,Wang2019}, and meta-learning is reviewed more broadly in \cite{Hospedales2022}.

\begin{table}[hb!]
\renewcommand{\arraystretch}{1.25}
\centering
\resizebox{0.48\textwidth}{!}{
\begin{tabular}{p{0.25\textwidth}p{0.12\textwidth}p{0.05\textwidth}p{0.05\textwidth}p{0.05\textwidth}p{0.05\textwidth}}
\multirow{2}{*}{Model} & \multirow{2}{*}{Backbone} &\multicolumn{2}{c}{CIFAR-FS} &  \multicolumn{2}{c}{Mini-ImageNet} \\
& & 1-shot & 5-shot & 1-shot & 5-shot \\
\hline
P$>$M$>$F \citep{Hu2022a} & DINO-ViT-base & 84.3\% & 92.2\% & 95.3\% & 98.4\% \\
SOT \citep{Shalam2022} & ResNet-12 & 89.9\% & 90.7\% & 92.8\% & 88.8\% \\
PEMnE-BMS \citep{Hu2022b} & WRN & 88.4\% & 91.9\% & 85.5\% & 91.5\% \\
\hline
\end{tabular}}
\caption{Performance of Few-Shot Classifiers on CIFAR-FS and Mini-ImageNet}
\label{tab:fewshot}
\end{table}

The levels of performance of three state-of-the-art few-shot learners are presented in Table \ref{tab:fewshot}. Two common datasets are used: the CIFAR-FS \citep{Bertinetto2019} and Mini-ImageNet \citep{Vinyals2016}. Both have structures similar to that illustrated in Figure \ref{fig:fewshot}, and performance is evaluated here on the easy type of problem of discriminating across randomly selected holdout classes. These datasets are subsets of the CIFAR-100 \citep{Krizhevsky2009} and ImageNet ILSVRC-2012 \citep{Russakovsky2015} sets of images that have been configured for few-shot learning. Results are shown in all cases for 5-way few-shot estimation. Performance is evaluated by introducing the model to five new classes of images, giving it one or five examples of each, and testing its ability to accurately discriminate among test images from those five classes. The P$>$M$>$F model of \cite{Hu2022a} involves three steps of pre-training an image classification model, meta-learning the task generalization process, and fine-tuning to the support data for testing. Its pretraining employs as a backbone the DINO \citep{Caron2021} self-supervised variant of the ViT \citep{Dosovitskiy2020} model, and it uses the ProtoNet algorithm \citep{Snell2017} for its meta-learning step. Like \cite{Vinyals2016}, the Self-Optimal Transport (SOT) approach of \cite{Shalam2022} involves the learning of feature embeddings for distance-based comparisons, using the ProtoNet algorithm \citep{Snell2017} on a small, customized convolutional net with a ResNet-12 backbone for fine-tuning \citep{He2016}. The PEMnE-BMS strategy of \cite{Hu2022b} supplements feature transformation-based transfer learning with additional steps including boosting, with a Wide ResNet (WRN) \citep{Zagoruyko2017} backbone following the specification of \cite{Mangla2020}. As the table shows, these recent top-performing models execute this task with an rate of accuracy of 84.3\% to 95.3\% for one-shot and 88.8\% to 98.4\% for five-shot learning.

\subsection{Transformers} \label{Transformers}

An additional area of task generalization in which exciting progress has been made in recent years is the \emph{transformer} architecture introduced and popularized by \cite{Vaswani2017}. Transformers build upon the autoencoder architecture for capturing deep features to reduce the dimensionality of complex data, popularized by \cite{Hinton2006a} and \cite{Masci2011}. The transformer architecture builds upon this encoding of deep structures by increasing the richness of the relationships among them. Like Recurrent Neural Networks (RNNs), transformers operate on sequential data like words in a sentence. Importantly, however, the transformer structure's ``self-attention'' mechanism allows for higher levels of connectedness and dependency across elements that are far away from one another than was achieved through RNNs.

\begin{figure*}
\begin{subfigure}{0.45\textwidth}
  \centering
  % include first image
  \includegraphics[width=\textwidth]{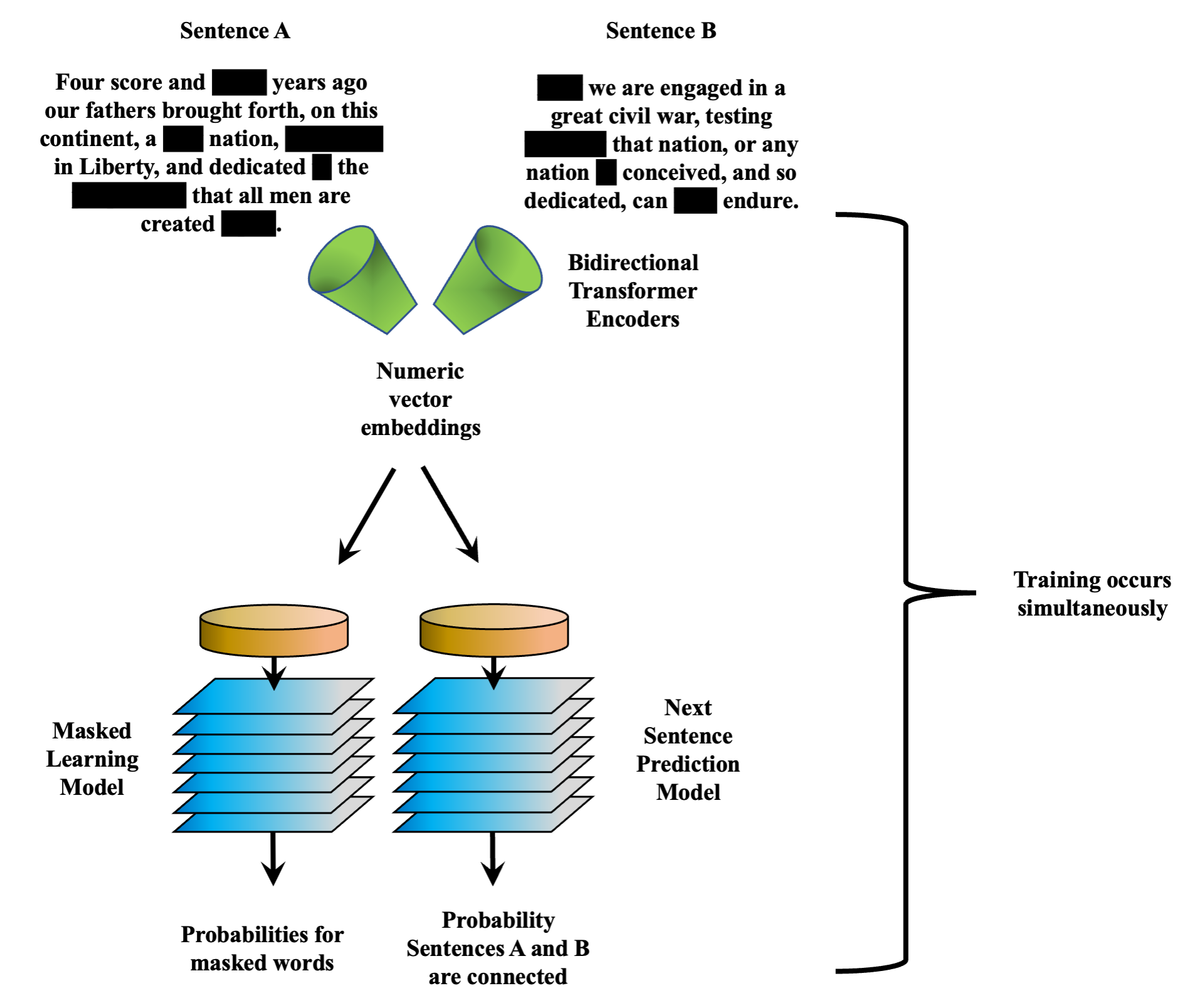}
  \caption{Training}
  \label{fig:bert-train}
\end{subfigure}
\begin{subfigure}{0.45\textwidth}
  \centering
  % include first image
  \includegraphics[width=\textwidth]{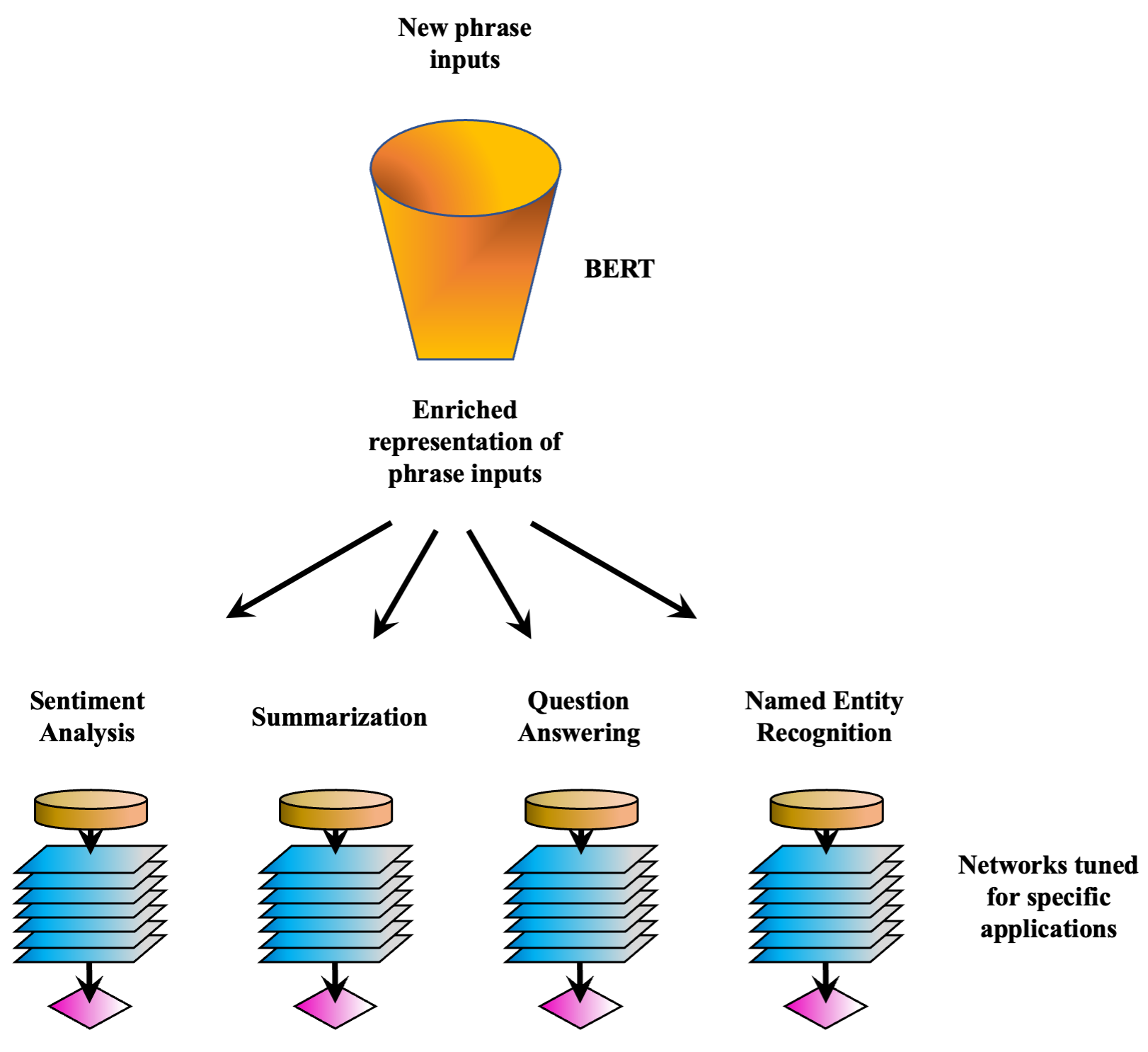}
  \caption{Fine-Tuning}
  \label{fig:bert-tune}
\end{subfigure}
\caption{Bidirectional Encoder Representation from Transformers (BERT)}
\label{fig:bert}
\end{figure*}

The transformer architecture has proven to be highly effective at improving model performance in a variety of tasks, particularly those related to vision and language. \cite{Chaudhari2021} reviews the architecture and a variety of applications. Key innovations in vision are discussed in \cite{Dosovitskiy2020} and a survey by \cite{Khan2022}.

Transformers' ability to understand relationships among distant elements of a sequence has proven to be critical in the processing of natural language data, in which a word like ``this'' or ``he'' can reference an entity mentioned far earlier or later in a sentence or phrase. One highly influential application of transformers in the NLP setting is the Bidirectional Encoder Representations from Transformers (BERT) approach for language comprehension introduced by \cite{Devlin2019}.

The overall structure of the BERT model is depicted in Figure \ref{fig:bert}. Figure \ref{fig:bert-train} illustrates the training process. The inputs are pairs of potentially related phrases with some words redacted. In the example, they are the first two sentences of the Gettysburg Address \citep{Lincoln1863}. The features from these sentences are transformed into vector representations. For each word, these embeddings capture context by characterizing distributions of words that tend to appear close to that word. As ``bidirectional'' in the model name implies, such nearby words are identified through forward and backward traversal of the phrase and consequently include words appearing both before and after the word being described. The embeddings are subsequently fed as inputs into two classifiers. The first determines the identity of the masked words. The second determines if the two sentences are related in the sense of occurring one after the other in a text. The entire process is trained simultaneously.

\begin{table}[ht!]
\renewcommand{\arraystretch}{1.35}
\centering
\resizebox{0.50\textwidth}{!}{
\begin{tabular}{p{0.21\textwidth}p{0.21\textwidth}p{0.08\textwidth}}
Model & Test & Performance \\
\hline
Human Benchmark & GLUE (9 tests) & 80\% \\
DeBERTa \citep{He2021} & \citep{Wang2019b} & 90.0\% \\
ALBERT \citep{Lan2020} & & 89.4\% \\
Human Benchmark & SuperGLUE (8 tests) & 89.8\% \\
DeBERTa \citep{He2021} & \citep{Wang2019c} & 90.3\% \\
\hline
\end{tabular}}
\caption{Performance of BERT-based Approaches on Language Tests}
\label{tab:bert}
\end{table}

Tuning and deployment of the BERT model is shown in Figure \ref{fig:bert-tune}. In testing and deployment, the embeddings, the distributions of likely words, and the classifications of sentence relatedness are combined into a broader set of embeddings that describe pairs of phrases. Subsets of those embeddings are used as inputs for task-specific classifiers. The training of those task-specific classifiers is described as tuning or fine-tuning.\footnote{A single BERT is shown in Figure \ref{fig:bert-tune} for simplicity. In practice, BERT is often tuned differently for different tasks, so that the version used for each one would be slightly different.} The model itself becomes a feature transformer, consuming phrases or pairs of phrases and outputting the intermediate embeddings, the probabilities for the masked words, and the sentence similarity measure. Subsets of this complete set of embeddings are used as inputs for specific applications such as sentiment analysis or rating the correctness of potential answers to supplied questions. Each of the applications is trained separately through a process described as fine-tuning.

Table \ref{tab:bert} presents the results of two recent BERT-based networks, DeBERTa and ALBERT, on two batteries of phrase comprehension tests that are commonly used to evaluate NLP engines. The General Language Understanding Evaluation (GLUE) and SuperGLUE measures are averages across batteries of nine and eight tests, respectively, and NLP engines are assessed on their average performance. Typical human performance is provided as a benchmark. As the table illustrates, for both tests, state-of-the-art BERT-based methods outperform humans.

\subsubsection{Transformers as Foundation Models} \label{Foundation}

\begin{table*}[ht!]
\renewcommand{\arraystretch}{1.35}
\centering
\resizebox{0.85\textwidth}{!}{
\begin{tabular}{p{0.25\textwidth}p{0.30\textwidth}p{0.40\textwidth}}
Modality & Representative Tasks & Models \\
\hline \hline
Vision & Object detection \& recognition, image segmentation, resolution enhancement, captioning, \& generation. & ViT \citep{Dosovitskiy2020}, Swin Transformer \citep{Liu2021c} \\
Language & Text classification, summarization, generation, \& translation and question answering. & BERT \citep{Devlin2019}, LLaMA \citep{Touvron2023}, GPT \citep{Brown2020} \\
Multimodal (Vision \& Language) & Test-based image search, description, \& generation. & CLIP \citep{Radford2021}, DALL-E \citep{Ramesh2021}, ALIGN \citep{Jia2021} \\
\hline
\end{tabular}}
\caption{Representative Transformer-based Foundation Models}
    \label{tab:foundation}
\end{table*}

Because transformer-based architectures are so versatile and portable, researchers are increasingly relying upon pre-trained \emph{foundation models}. The use of such models mirrors the strategies used in domain generalization and few-shot learning, through which a model trained on a large dataset can then be applied to slightly different problems with relatively little domain- or task-specific training. Established neural networks trained on ImageNet, such as those described in Subsection \ref{ImageNet} on sample generalization, are examples of foundation models. The popularity of transformers has led to the application of this strategy on a larger scale. Popular toolkits for using foundation models include \cite{Wu2019} in the vision setting and \cite{Wolf2020} for NLP. \cite{Bommasani2021} discuss the patterns and implications of this industry-wide shift.

Table \ref{tab:foundation} summarizes some transformer-based foundation models that are popular now and some tasks to which they are commonly applied. In addition to those dedicated to vision-related tasks and those specific to NLP, some use data from both modalities to inform one another, a form of generalization that we discuss in greater detail in the next section.

\section{Modality Generalization} \label{Modality}

Artificial neural networks are typically specific to a datatype, considering only images, only text, or only audio, for instance. Evidence from the neuroscience literature suggests that, by contrast, some important elements of learning and memory operate at a more abstract level in which information across different sensory sources is consolidated. In conditioning experiments on fruit flies, for instance, \cite{Ueno2017} finds that some stimuli only lead to learned responses when researcher-generated odors and lights are simultaneously present. The authors propose that this phenomenon may arise due to stability-enhancing dual-gated learning pathways. Additionally, evidence from \cite{Quiroga2005} suggests that human brains may have cells dedicated to specific individuals like celebrities, so that seeing a picture of American actress Jennifer Aniston or just her name would activate the same location in the brain.

In the same way that biological brains abstract learn and remember entities and phenomena in a way that is distinct from the information source, some research in artificial neural networks transfers learning across modalities---that is, across different data formats or spheres of knowledge from those of the original model. It is a broader form of generalization than distribution, domain, and task and is in a sense a drastic change in all of these at once. The literature in this area is not extensive, but it is growing. Some recent influential transformer-based models, for instance, achieve high levels of performance at vision-related tasks by exploiting deep and extensive cross-model information. The CLIP model \citep{Radford2021}, used as a backbone in some of the domain generalization models described in Subsection \ref{Transfer}, is one example. Such models incorporate information on how different types of images have been described with text as well as learned skills on how text-based concepts relate to one another. This combined sourcing of datatypes enables the model to make connections and identify related concepts more effectively than is achieved with visual data alone.

\begin{table*}[ht!]
\renewcommand{\arraystretch}{1.35}
\centering
\resizebox{\textwidth}{!}{
\begin{tabular}{p{0.30\textwidth}p{0.30\textwidth}p{0.30\textwidth}p{0.10\textwidth}}
\multicolumn{4}{c}{Panel A: Text-Supplemented Image Classification} \\
Model & Modality & Classification Task & Performance \\
\hline
Deep Transfer Networks & Images alone & Classify web images & 70.7\% \\
\citep{Shu2015} & Images with text embeddings & Classify web images & 80.7\% \\
CLIP \citep{Radford2021} & Images with coarse captions & Classify ImageNet images & 82.9\% \\
\citep{Vasu2024} & Images with detailed synthetic captions & Classify ImageNet images & 86.4\% \\
\\
\multicolumn{4}{c}{Panel B: Mammalian Olfactory System Classifier} \\
\hline
Learning in the Wild & Olfaction & Odorants from gas sensor & 96.0\% \\
\citep{Borthakur2019} & Olfaction & Odorants from gas sensor with drift & 91.1\% \\
& Visual & Type of Japanese forest & 88.4\% \\
& Auditory & Species of frog or toad & 93.3\% \\
\hline
\end{tabular}}
\caption{Performance of Cross-Modality Transfer Models}
    \label{tab:modality}
\end{table*}

Table \ref{tab:modality} summarizes the findings from three innovative studies that help to illustrate possibilities in this area. The first two lines of Panel A present results from the Deep Transfer Networks (DTN) approach of \cite{Shu2015}, which builds upon a related non-network based classification approach by \cite{Qi2011}. In that model, a feature transformation function is trained based upon web-based images together with accompanying text descriptions. A classifier is then trained to label the images based upon these text-enriched embeddings of the image data. In deployment, the transformer generates embeddings for images that do not contain text descriptions---essentially describing text that tends to accompany images similar to the current input---and those embeddings are fed as inputs into the classifier. While newer transformer-based approaches improve upon the text-to-image generalization ability, this study is instructive because of its comparison of performance with and without the supplemental text-based data. The first line of the table presents the results of a baseline classifier using a Stacked Auto-Encoder that only includes data on images. The second line shows the classification results for the DTN model that incorporates the feature transformer trained with text descriptions. Results are shown from the authors' tests that include 10 training images per class. As the results show, the approach using embeddings substantially improves classification performance over the benchmark.\footnote{While not strictly modality generalization as described in this survey, some recent intriguing work combines image and text data to perform a variety of complex tasks such as labeling multiple objects in an image (\emph{cf.} \citep{Xu2022}), and \cite{Baevski2022} propose an approach for embedding data features that is designed to be common across speech, image, and text applications.}

The next two lines of panel A present results from \cite{Vasu2024}. That study compares the ability to classify ImageNet images between two versions of CLIP \citep{Radford2021} using different levels of data quality. The baseline version uses web-scraped data from OpenAI with imprecise and coarsely-written captions. The enhanced dataset used by the authors uses synthetic and more descriptive captions that were machine-generated separately for each image using a variation of the captioning model by \cite{Yu2022} and developed as part of the DataComp competition for improved CLIP performance \citep{Gadre2023}.\footnote{OpenAI-WIT compared with DataComp-DR from \cite{Vasu2024}. ViT-B/16 architecture \citep{Dosovitskiy2020} and input resolution of 224 pixels used in both.} As the results show, the use of greater detail and precision in the captions---even the synthetically generated kind that themselves derive from language-based transformer models---lead to a noticeable improvement in image classification performance.

Panel B of Table \ref{tab:modality} describes a modality generalization approach that is inspired by the structure of animalian brains. \cite{Borthakur2019} begin with a biologically-inspired classifier with numbers of nodes, levels of connectivity, and other design features based upon that of the mammalian olfactory system. They then examine the extent to which a set of hyperparameters developed for one problem of classifying odorants can be applied to different problems with comparable effectiveness. They develop their Spiking Neural Network (SNN) to classify odorants from gas sensor data in an online learning context, and they then re-train the classifier on data from the same sensor after it has aged three years. In the language of Figure \ref{fig:generality}, this first test can be viewed as something between distribution and domain generalization, as in the case of concept drift. The authors then apply the same network structure with the same hyperparameters to the classification of different types of Japanese forests from satellite images and to the classification of frog and toad species based upon auditory information. They find that their SNN setup retains a high degree of accuracy in these widely different problems. While changing the context and problem degrades the performance somewhat, they find that similar accuracy as the original classifier can be achieved without hyperparameter re-tuning by increasing the number of training cases.

\section{Scope Generalization} \label{Scope}

\begin{figure*}
  \centering
  \begin{subfigure}{0.45\textwidth}
    \centering
    % include first image
    \includegraphics[width=\textwidth]{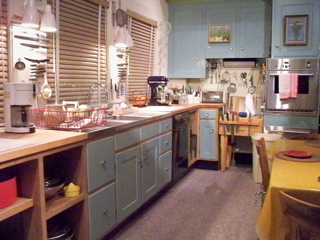}
    \caption{Original Photograph}
    \label{fig:julia_unlabeled}
  \end{subfigure}%
  \hfill
  \begin{subfigure}{0.45\textwidth}
    \centering
    % include second image
    \includegraphics[width=\textwidth]{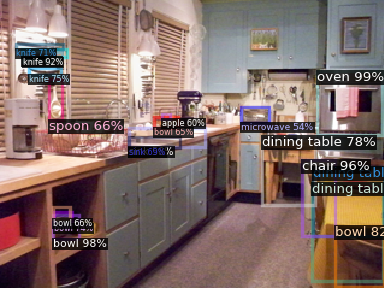}
    \caption{Objects Labeled}
    \label{fig:julia_labeled}
  \end{subfigure}
\caption{Julia Child's Kitchen}
\label{fig:julia}
\end{figure*}

In the mid-$20^{th}$ century, much of the work on artificial intelligence (AI) focused on codifying and automating logical reasoning. Early advances in this area, known as \emph{symbolic AI} and referred to by \cite{Haugeland1989} as ``good old-fashioned AI,'' engendered a sense of optimism with bold projections of the rate at which the field would develop \cite{Simon1958}. The decades that followed, however, are widely regarded as an ``AI winter'' of disappointing progress. When, in the late $20^{th}$ and early $21^{st}$ centuries, rapid advances in AI began again to surpass expectations, it was through the emergence of the powerful but less transparent technologies of machine learning and artificial neural networks \citep{Russell2020}.

In the process of exploring and refining these newer approaches, some research has begun to revisit AI's origins in symbolic reasoning. Generalization across scopes---that is, generalization at the level of abstract principles---and the field of \emph{neurosymbolic AI} involve two ways of translating between human and machine understanding. The first, knowledge representation, searches for ways in which logic and human insights can be incorporated into deep learning strategies to enhance their performance. The second, knowledge extraction, seeks to translate the outputs and reasoning from artificial systems so that they can be better understood by humans.

\subsection{Knowledge Representation} \label{Representation}

Some of the greatest challenges for modern AI involve the distillation of facts into principles and the application of those principles to new areas of knowledge. \cite{Davis2015} highlight the amount of semantic knowledge that humans regard as commonsense by considering the picture shown in Figure \ref{fig:julia_unlabeled} of Julia Child's kitchen.\footnote{Taken by Matthew Bisanz and presented as in \cite{Davis2015} to illustrate the many types of inferences that humans draw from stimuli.} A human viewer would naturally infer that four legs are holding up the tabletop and some sort of nail, hook, or adhesive is keeping the pictures on the cabinet doors---and that without them, they would all fall to the floor due to the force of gravity. While these thoughts might not occur to someone consciously, violations of such principles would be immediately noticeable. This example of the kitchen photograph helps to illustrate the many inferences that humans draw from stimuli and the depth of understanding required to replicate that process artificially.

Biological systems process information in a variety of ways. Some components of brains perform pattern recognition like deep learning does. Other less-understood components perform logical reasoning. No data-driven machine-based system yet exists that learns logical and symbolic principles effectively from scratch. It is not clear how such a learning system would be designed or what data it would use. In the meantime, however, researchers have pursued a hybrid approach in which humans compile their own learned principles and logical relationships and then train networks to apply them.

Substantial progress has been made to establish a base of facts and relationships among them as well as the tools necessary to expand upon that set of information. As \cite{Bader2005}, \cite{Besold2017}, and \cite{Townsend2020} note in surveys on neurosymbolic AI, some research has worked to incorporate human sense and domain expertise into deep learning systems. In some neural-symbolic cycles, the machine could be used to identify objects in an image, as in Figure \ref{fig:julia_labeled}, in which objects have been identified using the Detectron2 package from Meta \citep{Wu2019}.\footnote{Faster R-CNN architecture \citep{Ren2017} with RexNeXt-1010-32x8d backbone \citep{Xie2017} and a Feature Pyramid Network (FPN), traied for three times the default number of iterations (3x schedule) and optimized for object detection on the COCO dataset \citep{Lin2014}.} Researchers could include additional ``neural gates'' into the architecture that would capture logical conclusions. For instance, a human-designed gate might fire only if more than one cooking appliance is identified in a photo. Such logical components could help artificial systems to more quickly recognize that the room being pictured is a kitchen. This symbolic logic that would be used for deduction is often causal and related to counterfactuals. It is therefore important that object identification and other machine-based inferences that are made from the data are based upon neural networks whose projections are robust to this form of out-of-distribution generalization.

\begin{figure}[t]
\centering
% include first image
\includegraphics[width=0.45\textwidth]{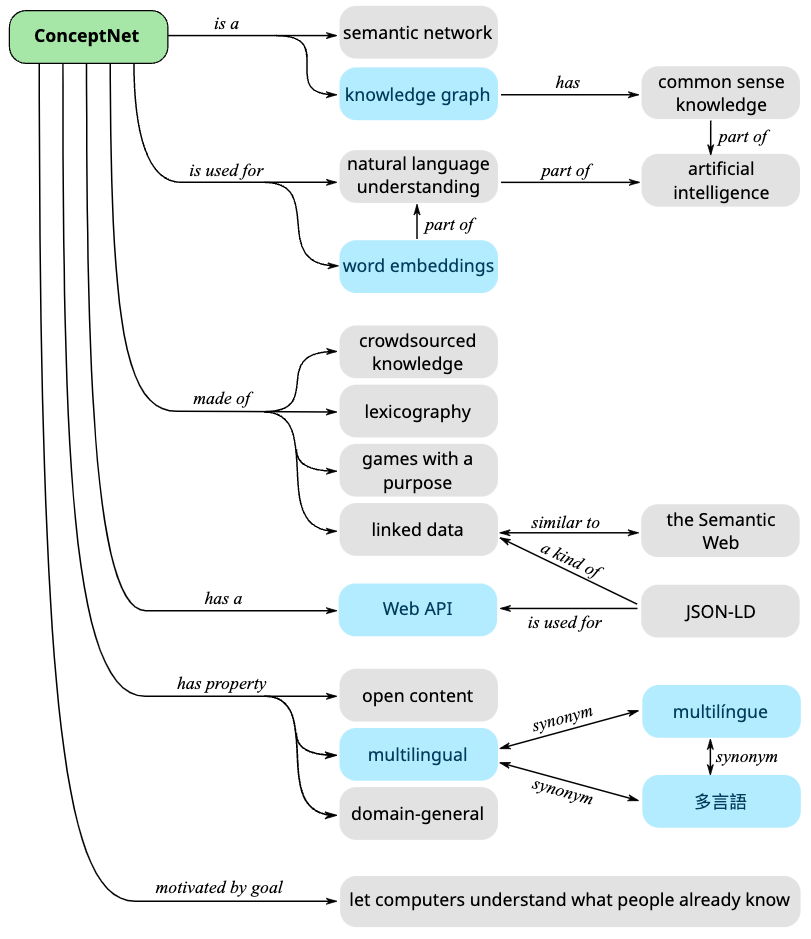}
\caption{ConceptNet Graph-based Knowledge Database}
\label{fig:conceptnet}
\end{figure}

Many facts about the world and logical relationships have been codified and introduced into AI models is through the development of \emph{knowledge graphs}, as illustrated by Figure \ref{fig:conceptnet}. The diagram taken from the website of the ConceptNet open source database \citep{Speer2017,Speer2012}, characterizes some key features about that particular database in its own lexicon. The directed graph consists of triplets of \{subject, relationship, object\} that represent abstract relationships such as ``similar to'' and ``part of'' between pairs of nouns. Much of the database is manually coded through crowdsourcing, an approach that has been found to produce less voluminous but more reliable data than those obtained through web crawling \citep{Davis2017}. ConceptNet and other knowledge graphs are frequently employed in the construction of vector-based embeddings for words, as in BERT or other transfer learning strategies, to enable the incorporation of these logical relationships into network learning approaches. Graph-based representations of knowledge are used in a variety of commercial applications including web search and question answering tasks by Google \citep{Dong2014}.

For queries whose answers already appear as entries in the database, usage of the knowledge graph is straightforward. For cases in which the answer is not identical to an existing entry, strategies for applying knowledge graphs are similar in general to those required for expanding the databases. As \cite{Ji2022} note, some graph-based approaches expand their knowledge bases through logical inference from existing relationships. hThe high rate of exceptions to general logical rules, however, hinders the automation of that task. Consequently, network-based learning approaches are used on their own or supplemented with human verification. These networks take as a starting point embeddings produced from the knowledge graph and employ separate classification approaches to detect relevant nouns, pairs of nouns to link, and relationships among them \citep{Ji2022,Wang2017}. A thorough review of network-based knowledge graph approaches appears in \cite{Ji2022}. Additional reviews discuss the construction and application of the graphs \citep{Davis2017}, the use of knowledge-enriched embeddings \citep{Wang2017}, and a quantitative comparison of approaches \citep{Lin2018}.

Competing approaches for representing semantic knowledge are evaluated annually in the SemEval competition. Tasks include multilinguistic detection of idioms and social cues such as a condescending or sarcastic tone and the parsing of taxonomies that are assumed or implied in context \citep{Emerson2022}. One example question from a prior year describes an event in which people sat down in a sauna and asks the artificial system whether they sat down on a bench or on the floor. The system is required to make this conclusion based upon outside knowledge of how a sauna is typically laid out. Highlights from the 2022 competition are described in Table \ref{tab:semeval}. Selected results (winner, baseline, and in one case, human benchmark) are shown for Task 7 (the winnder for most innovative task) and Task 11 (including the winner for the most innovative model) are shown. Results are shown for the top performer and the baseline for each task. The first of the two tasks is a fill-in-the-blank exercise similar to that performed in the training of BERT. The masked word is not obvious from the sentence, however, and must be inferred from the broader context of nearby sentences or from general knowledge, at times requiring understanding of subtle linguistic patterns such as metonymy. The human benchmark can complete the task with 79.4\% accuracy. A BERT approach without supplemental knowledge performed at 45.7\% accuracy. The winning approach by \cite{Shang2022} is an ensemble strategy that combines BERT-style network forecasting with the ERNIE knowledge graph-trained embeddings \citep{Sun2021}, which performs at 68.9\%.

\begin{table*}[ht!]
\renewcommand{\arraystretch}{1.25}
\centering
\resizebox{\textwidth}{!}{
\begin{tabular}{p{0.27\textwidth}p{0.25\textwidth}p{0.42\textwidth}p{0.10\textwidth}}
Task & Example & Model & Performance \\
\hline
\multirow{3}{*}{\shortstack{Plausible Clarifications\\\citep{Roth2022}}} & \multirow{3}{*}{\shortstack{Check ratings of different salons.\\Visit the salon's website.\\Call \_\_\_ and ask questions.}} & Human benchmark & 79.4\% \\
 &  & X-PuDu - winning submission \citep{Shang2022} & 68.9\% \\
&  & BERT Baseline \citep{Devlin2019} & 45.7\% \\
\hline
\multirow{2}{*}{\shortstack{Multilingual Named Entities\\(MultiCoNER) \citep{Malmasi2022}}} & \multirow{2}{*}{\shortstack{Patrick Gray, former director of the\\Federal Bureau of Investigation}} & DAMO-NLP - winning submission \citep{Wang2020b} & 85.3\% \\
&  & ROBERTa Baseline \citep{Conneau2020} & 47.8\% \\
\hline
\end{tabular}}
\caption{Performance on Selected Semantic Inference Tasks in SemEval-2022 Competition}
    \label{tab:semeval}
\end{table*}

The second of the two tasks with results shown in Table \ref{tab:semeval} is to identify named entities from short snippets of text with little context---where the named entity might be a phrase such as ``Inside Out'' or ``To Kill a Mockingbird'' that also happens to be an artistic work. The learner must separately tag people, locations, corporations, other groups, products, and creative works and must do so in 11 different languages (the multi-language version of the test). The baseline BERT-based approach successfully completes the multilingual version of the task with 47.8\% accuracy. The winning submission employs BERT-style embeddings together with a query-based knowledge retrieval approach, achieving 85.3\% accuracy at the task.

\subsection{Knowledge Extraction} \label{Extraction}

In addition to incorporating domain knowledge and commonsense logic into deep learning systems, a growing body of research works to make neural networks' outputs more easily interpretable. A human might justify the classification of an image by pointing to cues such as objects in Julia Child's kitchen in Figure \ref{fig:julia_labeled} or the mane and face of a lion. Researchers have observed, however, that a ``neural-symbolic gap'' exists, in the sense that neural networks' outputs are generally difficult to distill into simple and explainable logical processes \citep{Besold2017}.

As \cite{Townsend2020} notes, some research aims to derive explanations of this form from black box systems like artificial neural networks. \cite{Craven1995} describe the ex post translation of network-based decisions into tree-based logic. \cite{Guidotti2019, Park2018}, and \cite{Pedreschi2019} develop strategies for considering counterfactuals, generalizing patterns into rules, and constructing text-based justifications for networks' predictions. Work along these lines is highly in demand for application to real-world problems in medicine and finance. At the literature currently stands, however, no dominant tools exist to simply explain deep learning models' projections, and it is a promising area for future research.

In the absence of systematic intuitive explanations that distill network-based knowledge, the deep learning community makes extensive use of techniques for identifying the relative importance of different predictors. \cite{Townsend2020} cover these methods in detail in their review. Some are discussed briefly below, with applications to some simple examples. 

\subsubsection{Shapley Values} \label{Shapley}

One important technique through which machine learning researchers explain the importance of different model predictors is the use of Shapley values. The approach was initially developed by \cite{Shapley1953} for use in the context of game theory. Its more recent use for model attribution in a machine learning context was promoted and popularized by \cite{Lundberg2017}. The approach is ``model agnostic'' and is compatible with settings like those in economics, finance, and medicine in which learning algorithms use diverse sets of predictors, any of which could conceivably be excluded from a given model specification. For a model with $k$ predictors, the Shapley measure of influence for predictor $j$ on a given test case is calculated by measuring its marginal contribution to the final classification---that is, the score if the predictor is included in the model minus the score if it is left out. This score is averaged over the $2^{k-1}$ ways in which subsets of the remaining $k-1$ predictors are excluded from the model.

Shapley values have the advantage that they are highly versatile and can be used to understand any predictive model. Additionally, because the impact is averaged over multiple specifications, the statistic is robust to the many ways in which predictors interact and affect one another's influence. While the original approach is computationally intensive, requiring $2^k$ trained versions of the model, the SHapley Additive exPlanations (SHAP) approach introduced by \cite{Lundberg2017} employs approximations and other efficiency enhancements to make the calculation tractable. It does have the conceptual limitation, however, that its measure of a predictor's influence is based in part upon what its contribution would be in some highly implausible model specifications.

\begin{figure}[h!]
  \centering
  \begin{subfigure}[t]{\columnwidth}
    \centering
    % include second image
    \includegraphics[width=\columnwidth]{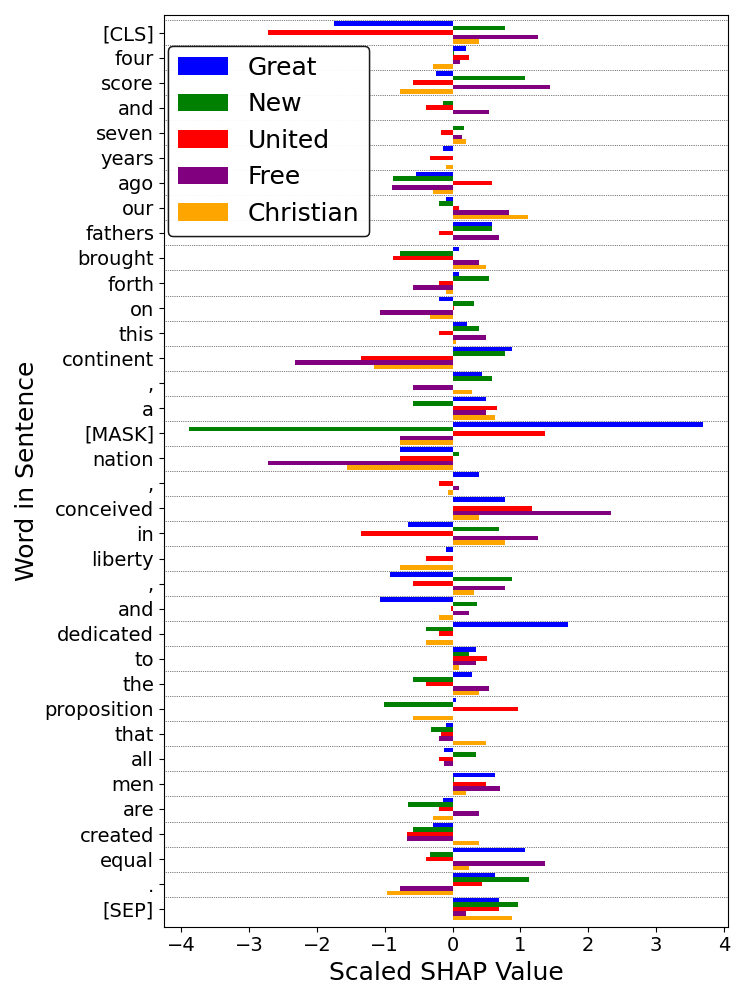}
    \caption{Shapley Values}
    \label{fig:gettysburg_shapley}
  \end{subfigure}
   \vspace{1em} % Add some vertical space between the subfigures
  \begin{subfigure}[t]{\columnwidth}
    \centering
    % include first image
    \includegraphics[width=\columnwidth]{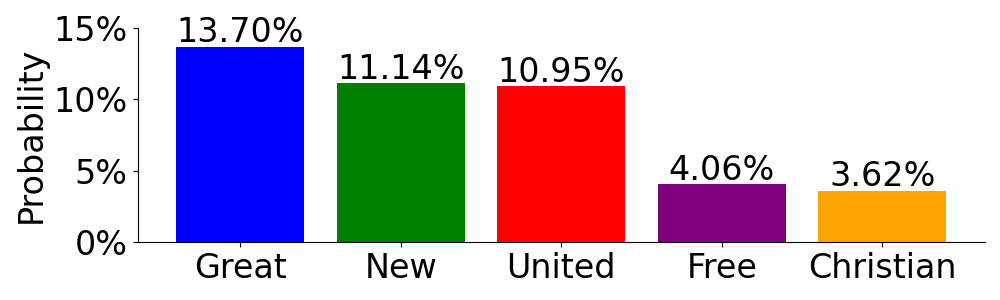}
    \caption{Probabilities for Top Choices}
    \label{fig:gettysburg_probs}
  \end{subfigure}%
\caption{Results from BERT Analysis of First Sentence of Gettysburg Address with ``new'' Masked}
\label{fig:gettysburg}
\end{figure}

Figure \ref{fig:gettysburg} presents Shapley values for one of the example natural language processing problems illustrated in Secion \ref{Task}. The sentence, shown along the vertical axis of Figure \ref{fig:gettysburg_shapley} is the first sentence of \emph{The Gettysburg Address} \citep{Lincoln1863}, with the word ``new'' masked. A BERT model is used to predict the masked word. Figure \ref{fig:gettysburg_probs} shows the model's top five selections and their corresponding probabilities. SHAP values are computed for each of the words in the sentence, the punctuation, and the tokens [CLS], [MASK], and [SEP], which represent the beginning, position of masked word, and end of the sentence, all of which capture elements of the sentence itself and the context in which it appears (in this case as a standalone sentence). A custom gradient-based implementation is used here, with each word's contribution aggregated across all of the hidden layers in the network.

As Figure \ref{fig:gettysburg_probs} shows, the BERT selects ``great'' as the most likely word to appear in place of the [MASK] token. The correct value of ``new'' is the second choice with roughly equal probability to ``united.'' As Figure \ref{fig:gettysburg_shapley} shows, the SHAP values provide some insight into the transformer's decision-making process. The components of the sentence that are most supportive of the top choice of ``great'' are the position of the masked word and ``dedicated,'' and the component that is least supportive of ``great'' is the beginning, which is generally interpreted as the context in which the sentence appears. The word ``new'' is relatively uninformative, being redundant with ``brought forth.'' Relative to the first choice, ``new'' is driven less by other words in the sentence and is more sensitive to the positional terms and punctuation.

\subsubsection{Attribution Heatmaps} \label{Heatmaps}

\begin{figure*}
  \centering
  \begin{subfigure}{0.45\textwidth}
    \centering
    % include first image
    \includegraphics[width=\textwidth]{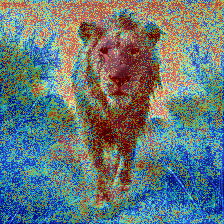}
    \caption{Lion in Natural Habitat - LRP}
    \label{fig:lionhabitat_lrp}
  \end{subfigure}%
  \hfill
  \begin{subfigure}{0.45\textwidth}
    \centering
    % include second image
    \includegraphics[width=\textwidth]{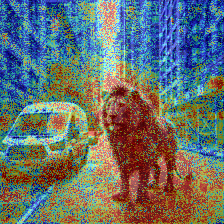}
    \caption{Lion in the City - LRP}
    \label{fig:lioncity_lrp}
  \end{subfigure}
  \begin{subfigure}{0.45\textwidth}
    \centering
    % include third image
    \includegraphics[width=\textwidth]{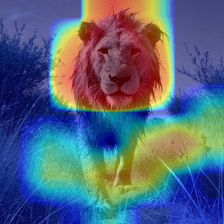}
    \caption{Lion in Natural Habitat - Gradient CAM}
    \label{fig:lionhabitat_grad_cam}
  \end{subfigure}%
  \hfill
  \begin{subfigure}{0.45\textwidth}
    \centering
    % include fourth image
    \includegraphics[width=\textwidth]{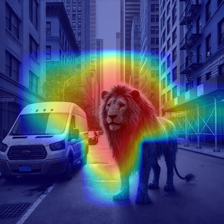}
    \caption{Lion in the City - Gradient CAM}
    \label{fig:lioncity_grad_cam}
  \end{subfigure}
  \caption{Attribution of MobileNet v3 Large Classifications}
  \label{fig:attribution}
\end{figure*}

While SHAP values are sometimes applied to images, some heatmap-based approaches are popular and informative for deep convolution networks in vision-related tasks. Figure \ref{fig:attribution} illustrates two methods for the attribution of network-based features to model outputs. Both are applied to better understand the classifications of the artificially generated lion photos presented in Section \ref{Distribution}. Results are shown for the MobileNet v3 Large network, as that was one for which ``lion'' was predicted with high confidence when the animal was in its natural habitat and ``chow chow'' was predicted, also with high confidence, when the animal was observed in the city. The images help to illustrate what factors led the network to arrive at those conclusions. All four images show pixel-specific heatmaps illustrating what portions of the images are important for the final projections. In all cases, the heatmaps are superimposed on the original images for readability.

The first approach, Layer-wise Relevance Propagation (LRP), introduced by \cite{Bach2015}, considers attribution in a holistic way. For each pixel's contribution to the confidence score for the chosen class is summed over all the layers in the neural network. The LRP thus indicates how much each of the pixels in the original image matters in the model's final classification.

The second approach, gradient-weighted Class Activation Mapping (gradient CAM), was developed by \cite{Selvaraju2017}. It constructs a heatmap at the pixel level based upon the last convolutional layer in the model. The intensity illustrated in the map is the gradient of that pixel on the confidence score for the chosen class. Thus the gradient CAM shows, after the many network steps, what elements in the final processed image are most important for its classification.

As the graphs illustrate, the model is more concentrated on the object of interest in the final convolution layer than overall. Additionally, the many distracting elements in the city photograph inhibit the model's ability to focus on the object of interest. The model's attention is diluted in the city photo as compared to the natural habitat photo.

For the lion in its natural habitat in Figure \ref{fig:lionhabitat_lrp} in the top left, the MobileNet v3 Large network dedicates most of its attention to the upper and central portions of the image, and it is those areas that contribute the most to the model's final classification. The model makes relatively little use of the information from the grassy areas. These low scores are consistent with the background being less important than the foreground in identifying the object in the photograph. As Figure \ref{fig:lionhabitat_grad_cam} below it shows, the focus on the foreground is more pronounced in the final convolution layer. While the earlier layers of analysis use a larger part of the image to identify edges and shapes, by the time the analysis reaches the final convolution layer, the network's attention is concentrated on the parts of the photograph that are most relevant for discerning the class. In the case of the lion in its natural habitat, that area is the animal's face.

On the right-hand side, Figures \ref{fig:lioncity_lrp} and \ref{fig:lioncity_grad_cam} illustrate a similar pattern, but with some important differences. The LRP, which captures influence throughout the layers of the network, shows a wide dispersion of influence across the photograph. While the area containing the lion itself matters the most, the street, van, and buildings all have notable contributions to the final classification. As with the natural habitat photo, moving from LRP to gradient CAM increases the amount of attention that is focused on the subject in the foreground. The level of focus is considerably less concentrated in the city photo than in the natural habitat photo, however. The many city-related elements in that picture appear to be key contributors to its final selection of chow chow as the class for that image.

\begin{figure*}[!t]
  \centering
  \begin{subfigure}[b]{1.0\textwidth}
    \centering
    % include first image
    \includegraphics[width=\textwidth]{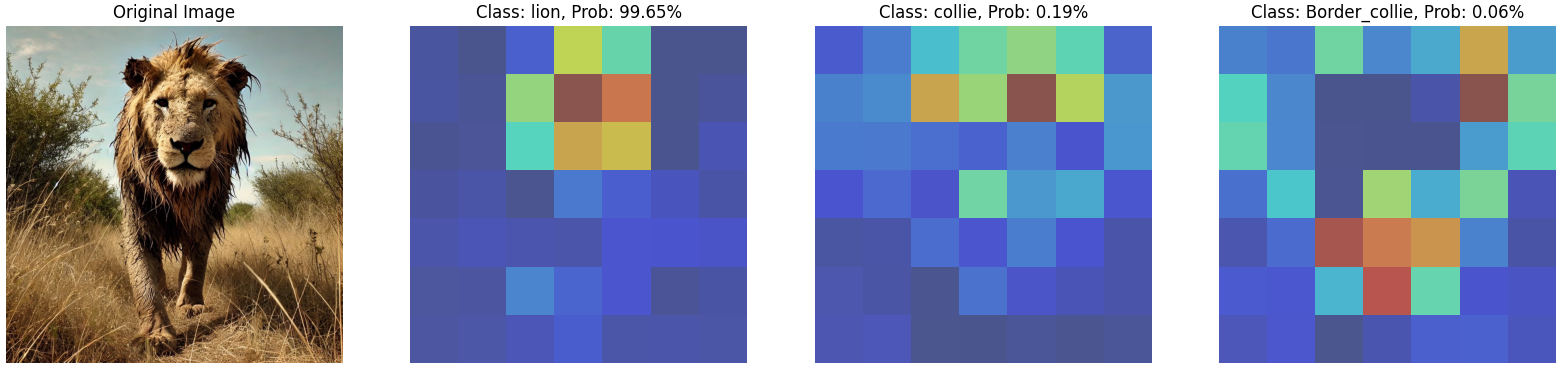}
    \caption{Lion in Natural Habitat}
    \label{fig:lionhabitat_cam_output}
  \end{subfigure}
\vspace{0.5cm} % Add vertical space between the subfigures
  \begin{subfigure}[b]{1.0\textwidth}
    \centering
    % include second
    \includegraphics[width=\textwidth]{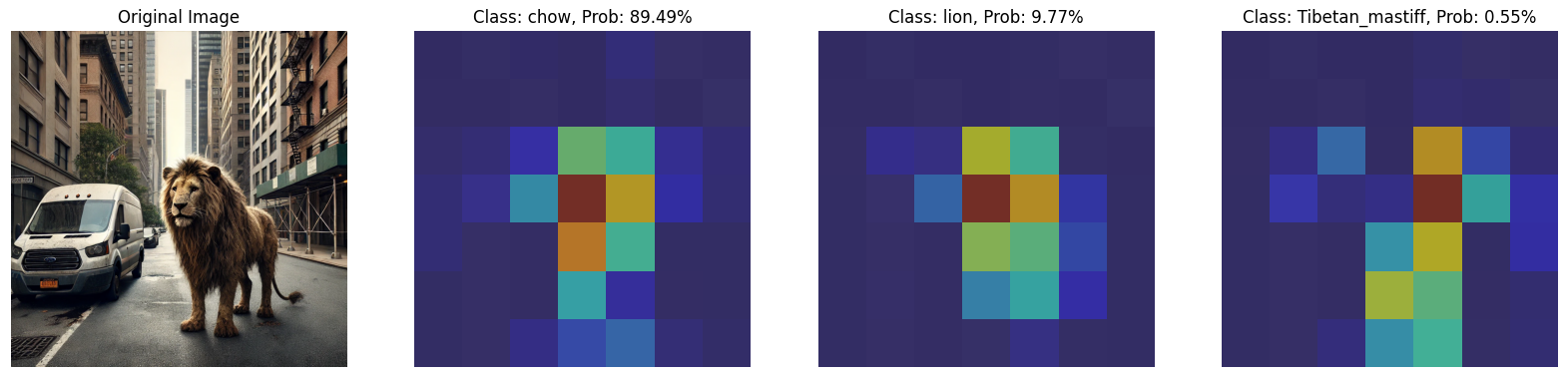}
    \caption{Lion in the City}
    \label{fig:lioncity_cam_output}
  \end{subfigure}
  \caption{Gradient CAMs for Alternate MobileNet v3 Large Classifications}
  \label{fig:alternate}
\end{figure*}

In order to more clearly illustrate the tradeoffs that the classifier makes in its considerations, Figure \ref{fig:alternate} shows gradient CAM representations not only for the chosen class but for the model's second and third choices for each image. These heatmaps help to illustrate what portions of the photograph made the chosen class more likely as compared to which portions mattered for the key alternatives. For both images, the most promising choices besides lion are dog breeds.

For both photographs, the components of the image that contribute most strongly to a classification of lion---shown in the first CAM heatmap in Figure \ref{fig:lionhabitat_cam_output} and the second in Figure \ref{fig:lioncity_cam_output}---are concentrated around the animal's face. As the other heatmaps show, for both images, the legs and surrounding area are more important contributors to the classification of the image as a breed of dog.

\section{Abstraction in Nature} \label{Nature}

As researchers endeavor to improve the ability of artificial networks to generalize and abstract, the structures of animal brains provide a valuable template. One distinguishing feature of biological neural networks is modularity. Neuroscientists have found that animal brains are segmented into well-defined components or modules that are specialized for the handling of specific tasks. In \emph{Drosophia}, for instance, standardized innate responses are processed in the lateral horn, while learned responses and prioritization are handled by a separate component known as the mushroom body \citep{Aso2016,Aso2014b,Devineni2022,Rohlfs2023a}.

\begin{figure*}
\begin{subfigure}{0.30\textwidth}
  \centering
  % include first image
  \includegraphics[width=\textwidth]{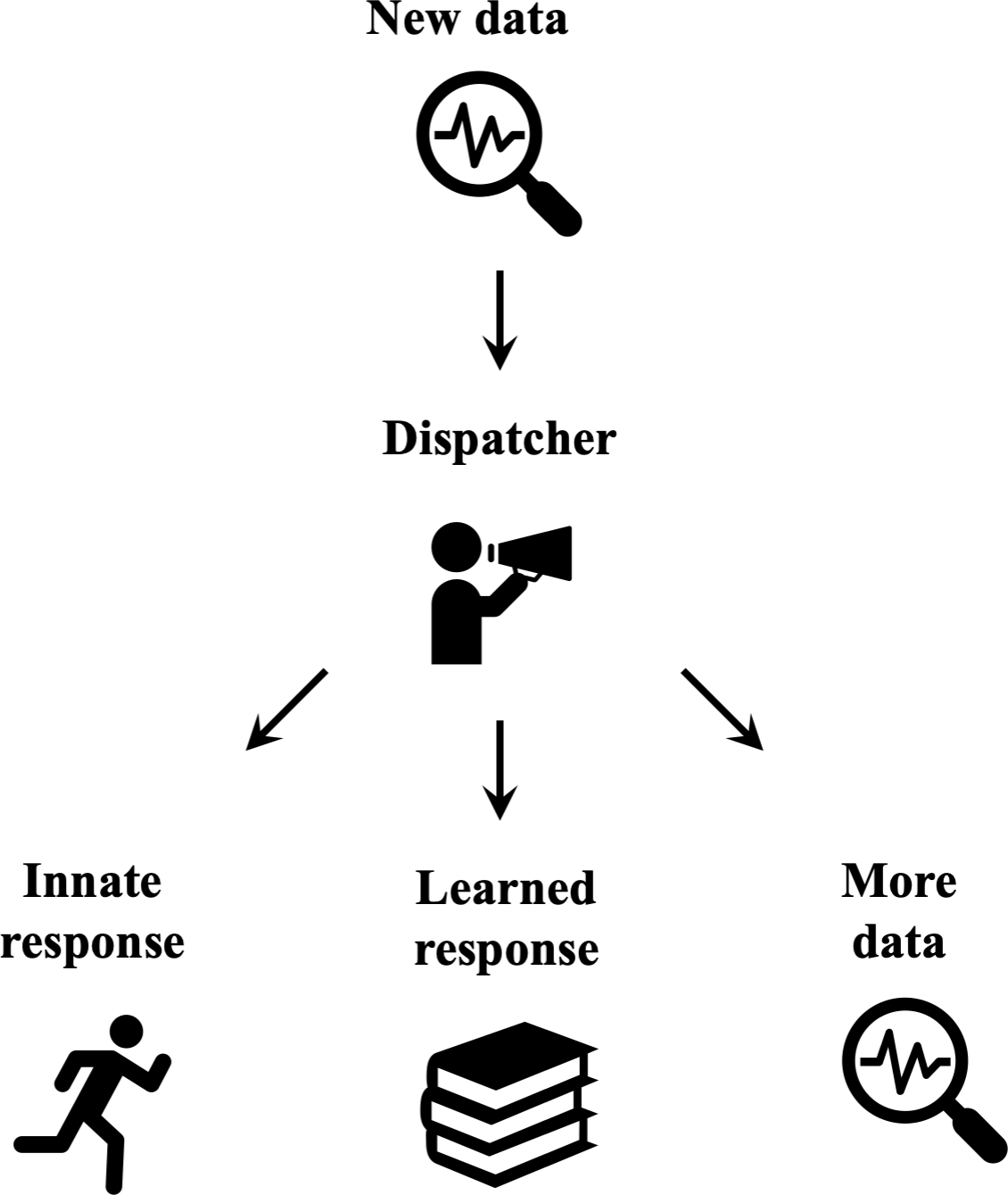}
  \caption{Module-based Routing}
  \label{fig:modules}
\end{subfigure}
\hfill
\begin{subfigure}{0.62\textwidth}
  \centering
  % include first image
  \includegraphics[width=\textwidth]{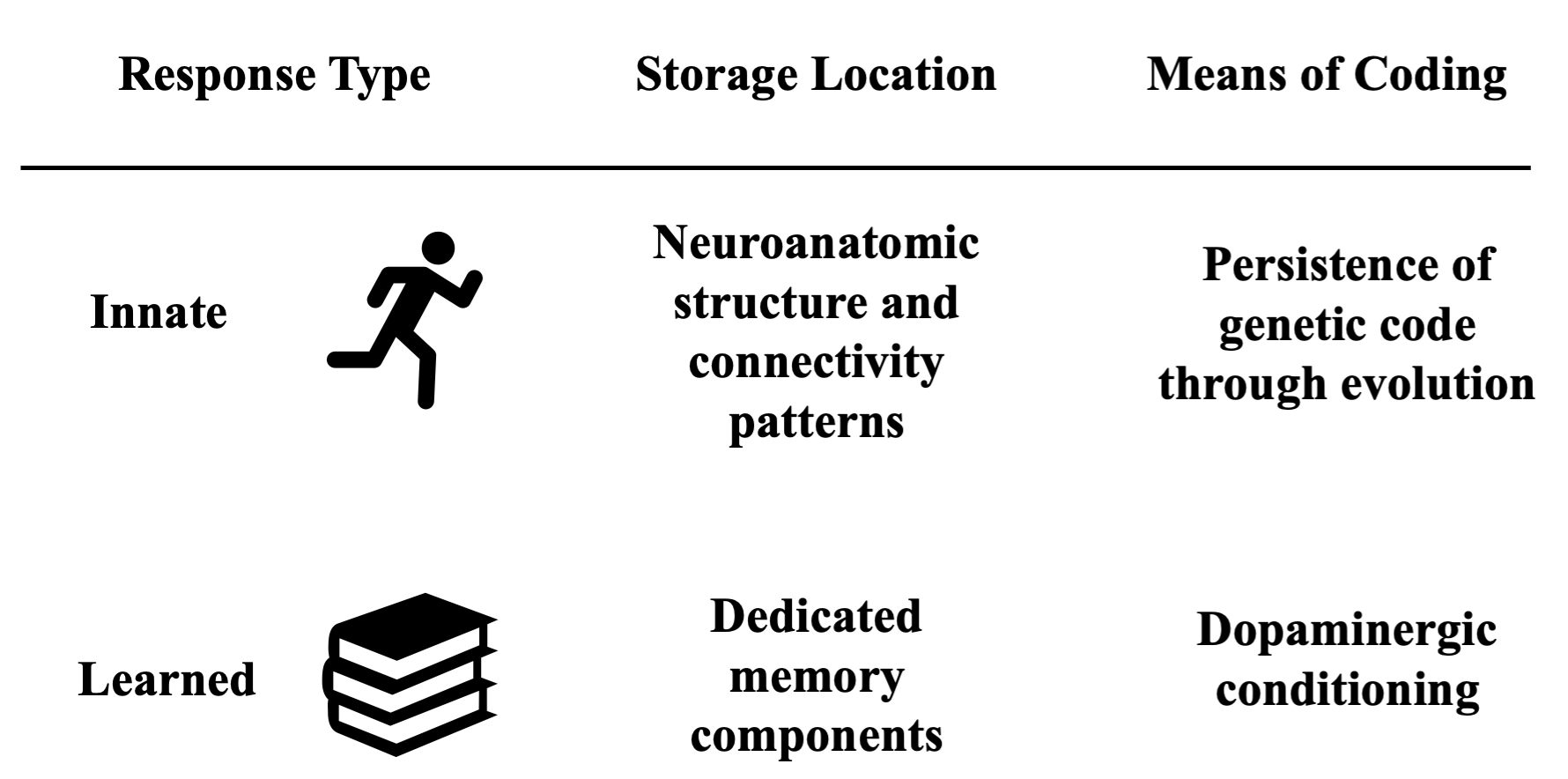}
\vspace{0.13 in}
  \caption{Innate and Learned Responses}
  \label{fig:responses}
\end{subfigure}
\caption{Responses to Stimuli in Animal Brains}
\label{fig:stimuli}
\end{figure*}

This modular structure is illustrated in a stylized form in Figure \ref{fig:modules}, which is a schematic interpretation the structures in animal brains through which data are routed to the appropriate type of response. Each sensory input---described in the diagram as ``new data'' is routed by a dispatcher into one of three channels: innate response, learned response, and more data. The network of evolved pathways, with separate treatment for distinct forms of inputs, can be thought of as the dispatcher. Stimuli that have hard-wired reactions are often handled in a separate component of the brain. Modularity of this form has been observed in a variety of animals, with different parts of the brain activated during abstract tasks than concrete ones in humans \citep{Gilead2014,Wurm2015,Vaidya2021}, and lesions to the prefrontal cortex have been found to inhibit abstract thinking in primates \citep{Mansouri2020}. This processing is often viewed as hierarchical, with successively complex components of the brain handling successively sophisticated problems \citep{Kaiser2010,Kiebel2009,Meunier2009,Meunier2010}.

The organization of the brain into specialized components helps it to maintain consistent performance across contexts. It helps to diversify risks by containing the overall system's vulnerability to any one component's errors. Because each module is used in a variety of applications and is informed by observations from those contexts, the trained structure is relatively insensitive to aberrations in any one domain of knowledge \citep{Sinz2019}. Stability is also maintained in the brain through system-wide limits on the potential influence of new information. \cite{Sadeh2021} describe the brain's process of inhibitory stabilization, whereby a given component's signal expression might be subdued by other parts of the brain that have a broader perspective on the priorities of the overall structure. In addition to this short-term regulation of signal strength, brains perform a process of controlled forgetting of longer-term information that is influenced by the rates at which new neurons are created \citep{Akers2014}. This gradual removal of outdated information enables the system to adjust to general trends that impact relationships and categories in the data without the thrashing from one decision rule to another that could emerge if updating were instantaneous \citep{Richards2017}.

The routing process illustrated in Figure \ref{fig:modules} takes into account the priorities of the organism according to multiple time scales. If recent events call for immediate action---\emph{e.g.}, flight---then chemical signals produce a temporary change in overall priorities at the system level \citep{Devineni2022}. \cite{Savin2014} find that, when a primate moves from one task to another, the change is evident through a shift in activity among large numbers of neurons in the prefrontal cortex---with urgent tasks causing larger numbers of cortical neurons to be repurposed. Also among primates, \cite{Froudistwalsh2020} note that system-wide dopamine levels in the brain impact the performance of working memory---in some cases promoting recurrence among the affected sets of neurons or blocking inhibitory signals from other cells. From a long-term perspective, system-wide priorities are embedded into the structure of the network, with receptors to important stimuli exercising more influence in the network. Connections that appear to reflect and to impact subjective valuations of past and prospective actions are distributed throughout the brain \citep{Lee2012}. \cite{Ji2020} find evidence of circuit-wide activity patterns such as mating, foraging, exploration, exploitation that influence behavioral state for multiple components of the brain in roundworms.

The first two types of responses in Figure \ref{fig:modules}---innate and learned---are fundamentally different in the ways in which they develop and operate. These differences are summarized in Figure \ref{fig:responses}, which describes how innate and learned responses differ in the ways in which the information is coded and stored. The third response, ``more data,'' reverts back to the top of the diagram, so that the responses described in Figure \ref{fig:responses} are the absorbing states of the flow chart.

Like the overall modular design and routing process, innate responses are driven by structural characteristics---the overall layout of the neural network as well as key hard-wired responses---that are stable over the life of the organism. These rules are thus stored in DNA and originate and change slowly through the process of genetic evolution. \cite{Barabasi2020} find that large patterns of synaptic connections---patterns that they call bicliques---are often shared across different animals of the same species. The authors propose that these similarities are innate, and they introduce a model to identify which portions of the connectome are coded by common genes. As \cite{Zador2019} notes, the amount of information stored as innate responses is minuscule compared to the overall capacity of the brain: the human genome can encode at most about a gigabyte of information relative to the 500 terabytes of information contained in the human brain connectome. 

``Learned,'' the second response type in Figure \ref{fig:stimuli}, relates to memory and plasticity, which serve an important role in the modular approach to thinking. In the case of \emph{Drosophila}, \cite{Vogt2014,Vogt2016} find that the structure through which neural pathways group sensory inputs into categories is largely fixed and that a key building block to the formation of memories is the learning of positive and negative associations with those categories. Depending on the flies' experiences, a given odor might attract some flies and repel others. The learned associations that are stored in memory are different, but the mental architecture mapping the stimulus to a response is similar. \cite{Knoblauch2010} characterize the first type of learning---the formation of positive or negative associations---as \emph{synaptic} plasticity. A second and less common type of learning, which \cite{Knoblauch2010} refer to as \emph{structural} plasticity, involves the modification of the categories with which associations are formed. In one example among fruit flies, \cite{Ueno2017} find a form of this plasticity that is a relatively sophisticated process that is only activated if both visual and olfactory sensory signals have been received---so that sensory input from one source serves to verify input from another source.

The grouping of stimuli in memory is a vital element of abstract thought. Neuroscientists find that the amount of abstraction and the level of complexity of these mappings are determined by the importance of such categories to the broader goals of the organism \citep{Cortese2021,Stegmann2020,Timme2016}. Rewards and penalties that are regarded as salient are connected to dopamine signals that drive animals to learn about causal relationships and the world around them. Animals acquire feedback from experience regarding the accuracy of the learned connections between stimuli and rewards or penalties---and the timing and intensity of neurons' dopamine responses update and adjust accordingly \citep{Kahnt2016,Schultz1997,Schultz1998}. A variety of intricate associations are are coded through this dopamine-guided learning process, operating on different time scales and accommodating uncertainty and context-dependence \citep{Dayan2001,Schultz2007}. \cite{Robertson2018} argues that the instability of specific memories serves an important role in this process of abstraction, enabling the translation of specific events into general rules.

\begin{table*}[ht!]
\renewcommand{\arraystretch}{1.25}
\centering
\resizebox{0.95\textwidth}{!}{
\begin{tabular}{p{0.10\textwidth}p{0.10\textwidth}p{0.32\textwidth}p{0.65\textwidth}}
Response Type & Feature & Description & Artificial Analogues \\
\hline
Modularity and Innate Responses & Specialized Components & Different anatomical components specialized to handle specific tasks & Transfer of feature transformations and embeddings (Sections \ref{Domain} to \ref{Scope}), Evolutionary deep learning \citep{Baymurzina2022,Wang2020a,Zhan2022} \\
& System-wide Prioritization & Domain generalization and concept drift (Section \ref{Domain}), Chemical signals regulate entire system behavior based upon context & Inhibitory signals \citep{Cao2018,Seung2018b,Seung2018a}, Multi-objective optimization \citep{Kim2022,Qu2021} \\
Learning and Abstraction & Archetypes & Repeated responses saved to conserve computational resources & Feature embeddings (Sections \ref{Domain} to \ref{Scope}), Knowledge Distillation \citep{Gou2021,Ku2020}, Neural architecture search \citep{Baymurzina2022,Dokeroglu2022,Jaafra2019} \\    
& Synaptic Plasticity & Prior experiences and responses associated with archetypes through conditioning & Feature embeddings (Sections \ref{Domain} to \ref{Scope}), Meta-learning (Section \ref{Task}) \\  
& Structural Plasticity & Representations adapt over time based upon relevance & Meta-learning (Section \ref{Task}), Concept drift and domain generalization (Section \ref{Domain}) \\
& Semantic Knowledge & Use of facts for deductive reasoning & Knowledge graphs (Section \ref{Scope}) \\
Data Collection & Selective Attention & Salient events receive more granular representations and more data collected & Replay and practice \citep{Andrychowicz2017}, Data Augmentation \citep{Shorten2019,Tian2022}, Adversarial training \citep{Chen2020,Frolov2021,Lust2021,Qiu2022}, Self-paced or curriculum learning \citep{Soviany2022} \\
\hline
\end{tabular}}
\caption{Artificial Analogues to Biological Modularity}
    \label{tab:modularity}
\end{table*}

This mapping of sensory inputs helps animal brains to produce archetypal examples that then interact with the experience of sensation. In experiments of vision in primates \citep{DiCarlo2017,Min2020} and listening in humans \citep{Feldman2009}, the categories developed in memory have been found to influence how new stimuli are perceived. Gaps in sensory information are filled in based upon previously encountered examples from the same category, producing a \emph{perceptual magnet effect}, which allows the brain to exploit existing trained skills and mental architecture that are built around these categories. These archetypes also assist in the perception of objects that are partially obscured, are placed in unusual contexts, or have distracting elements. \cite{DiCarlo2017} notes that, for the first 100 milliseconds of object detection, artificial classifiers' facilities rival those of humans and primates. When faced with more difficult cases in which the solution takes additional time, however, the biological brains' advantage becomes more pronounced.

\begin{figure*}
\centering
% include first image
\includegraphics[width=0.75\textwidth]{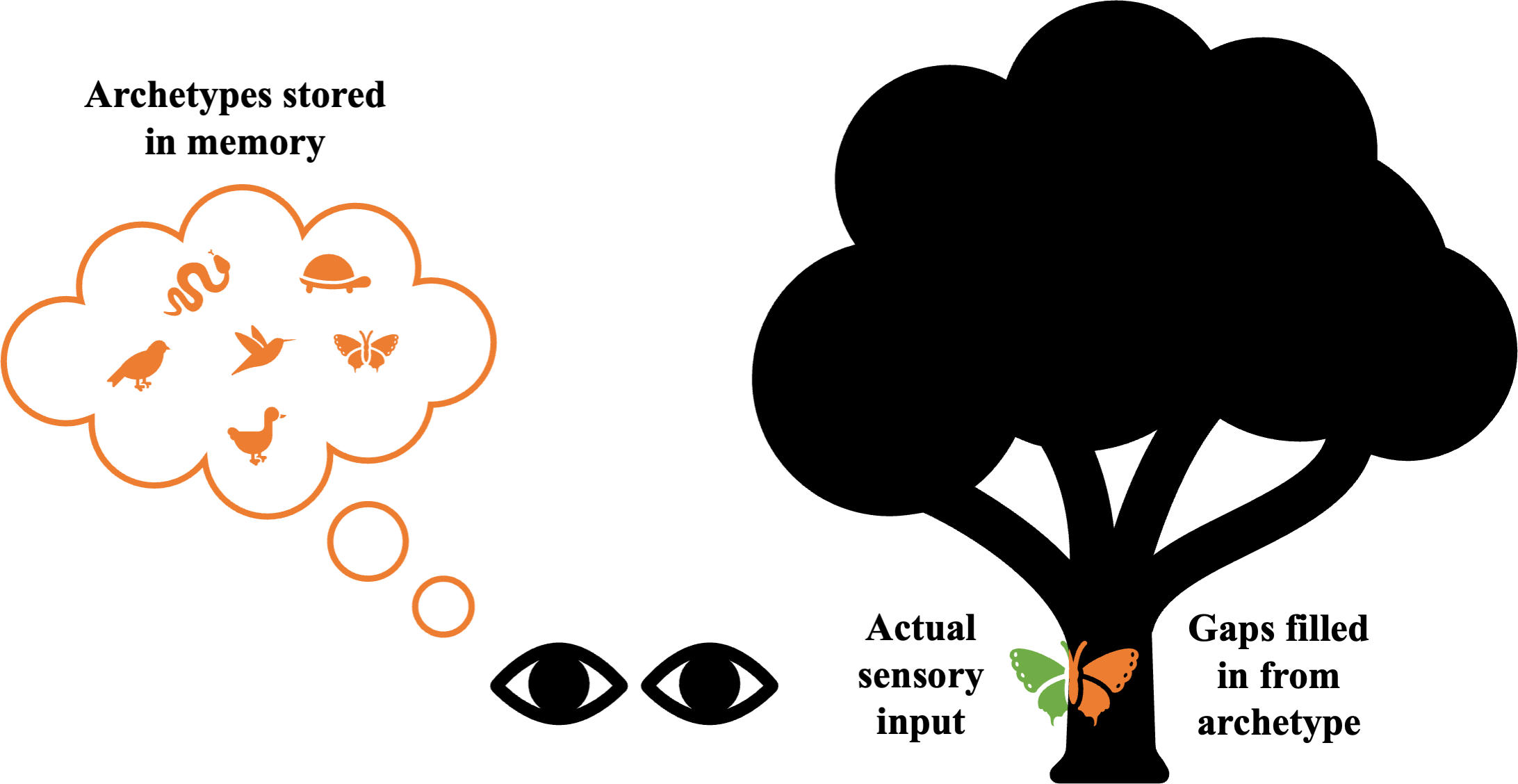}
\caption{Perceptual Magnet Effect}
\label{fig:magnet}
\end{figure*}

This perceptual magnet effect is illustrated in Figure \ref{fig:magnet}. In the image, the viewer perceives the left wing of a butterfly, but the body and right wing are obscured by the tree trunk and consequently unseen. In perceiving the butterfly, the viewer's mind matches the visual data to a category in memory and fills in the missing information based upon an image stored in memory of what a typical case looks like. This form of information processing helps humans to easily solve CAPTCHA-type problems such as those discussed in Section \ref{Domain} and illustrated in Figure \ref{fig:captcha}.

While the brain may initially fill in gaps based upon archetypes, both synaptic and structural plasticity interact with the third response from Figure \ref{fig:modules}, the collection of new data. The pursuit of new data for learning is online and curiosity-driven. Biological systems select cases to learn based upon what is new, deviates from expectation, or fills specific gaps in understanding \citep{Sinz2019}. This targeted training helps animals to limit cognitive effort to cases that provide sufficiently high benefit. A certain training case might offer insights into one chunk of a tasks but not another. Separating the information into chunks helps to identify what parts of a training example are new and to confine relearning and updates to those sub-problems for which the observation is relevant.

Research by \cite{Savin2014} suggests a practical set of principles for the organization of conceptual ``chunks'' in primates. Events are grouped into similar categories in working memory if they are connected to the same external stimulus, resulting motor response, or some combination of the two. \cite{Cohen2021} similarly argue that practice and repetition are performed in animals in order to achieve a consistent motor response. Additionally, research by \cite{Logiaco2019} suggests that animal brains may learn complex motor patterns by breaking them down into manageable ``motor motifs.''

Through this process, the construction of concepts and memories is driven by the same forces that motivate the organism more broadly. Nevertheless, the parts of the brain that are responsible for abstraction behave in many ways as distinct modules that operate on their own time scale. While the learning process is initially motivated by the pursuit of dopaminergic rewards, incentives and learning are distinct in the brain, and knowledge that is acquired in order to satisfy a previous need persists when priorities change \citep{Berridge2012,Berridge2016,Zhang2009}. Hence, ``motivational salience,'' which is malleable and based upon recent situational data, drives an animal's behavior. Its actions are also informed, however, by facts and understanding about the world that have been accumulated over its lifetime.

Once such concepts are established, an essential component of the ability to generalize is the identification of relationships among these categories. As a motivating example, \cite{Amidzic2001} find evidence of ``chunking'' in the neural activity of chess grandmasters. The authors observe that, when playing chess against a computer, grandmasters make less use than novices do of portions of the brain associated with analysis and recent memories, and they make greater use of parts of the brain associated with long-term memory. Thus the key difference in mental activity that the authors observe between experts and novices is not in the experts' use of faster or more powerful processors. The experts' brains employ principles that have been distilled and stored from previously encountered cases and tools for recognizing which principles apply when faced with a new situation.

Table \ref{tab:modularity} briefly summarizes some of the ways in which modularity can be seen in biological systems and analogous strategies used by artificial neural networks. Many of these approaches are discussed earlier in this paper. For artificial approaches that employ elements of modularity but do not relate directly to generalization and the scope of this paper, a few citations are included as pointers for the interested reader.

The first category of response types described in Table \ref{tab:modularity}, ``Modularity and Innate Responses,'' is broken into two features and relate to the role of anatomy in the modular behavior of neural networks. As noted in the discussions of Figure \ref{fig:stimuli} above, the routing of stimuli into innate and conditioned responses occurs through the use of components of the brain that are specialized for those tasks---whereby the innate responses emerge slowly through evolution while conditioned resposes develop over the life of the organism. This distinction, in which different types of responses follow different training processes and timelines, resembles the use of feature transformations in transfer learning models as well as in with pre-trained embeddings such as BERT and knowledge graphs that are developed in a general setting and then deployed for use on specific applications. Experimentation and variation based upon the principle of genetic recombination can also be seen in evolution-inspired processes for optimization of complex systems.

The second feature listed in the table, ``System-wide Prioritization,'' relates to the manner in which the full cognitive apparatus is equipped to respond to large scale events that impact the tradeoffs faced by individual components. Such behavior in artificial neural networks is discussed in Section \ref{Domain} above on domain generalization. Discussions in the literature on multi-objective optimization---and preferred methods for balancing different goals---offer another perspective on this prioritization problem. Additionally, some approaches simulate the strategy used in animal brains in which neurons send inhibitory and excitatory signals to regulate and balance the impacts of stimuli of different types.

The next category of response types, ``Learning and Abstraction,'' involves modularity in the grouping of combinations of stimuli into ideas. The use of archetypes in the brain is mimicked in artificial systems through the forms of generalization described in Sections \ref{Domain} to \ref{Scope}, including the use of feature embeddings. Related work in artificial neural networks that does not explicitly generalize but performs some manner of abstraction involves knowledge distillation---through which smaller networks are trained to exhibit the same behavior as larger and more complex ones. The literature on neural architecture search involves a somewhat different take on abstraction, in which decisions about the structure of the network are assisted through explicit optimization, so that the classifier takes on some of the responsibility that is traditionally handled by the researcher.\footnote{Results from some search-based network classification approaches appear in the discussion of overfitting in Table \ref{tab:sample}.}

``Synaptic plasticity,'' the learning of positive or negative associations with existing archetypes, is achieved in artificial systems both through feature embeddings and meta-learning, both of which take network structures developed in one setting and apply them to other problems. Strategies related to ``structural plasticity,'' the development and refinement of network-based architecture for learning such associations, involve meta-learning as well as transfer- and drift-based domain generalization approaches. Another key form of abstraction, ``Semantic Knowledge,'' relates to knowledge graphs and scope generalization, as discussed in Section \ref{Scope}.

In relation to the third type of biological response from Figure \ref{fig:responses}, the selective collection of additional data, a variety of deep learning approaches augment existing datasets through regularization or adversarial approaches, and a recent area of growth known as self-paced or curriculum learning presents training examples to a network in increasing order of difficulty.

\section{Discussion and Conclusion} \label{Conclusion}

This paper reviews concepts, modeling approaches, and recent findings related to the generalization of insights from neural networks. A taxonomy of forms of generalization is introduced, with a spectrum of increasing levels of abstraction, including generalization across (1) Samples, (2) Distributions, (3) Domains, (4) Tasks, (5) Modalities, and (6) Scopes.

For (1) sample generalization from training to test data, theoretical conditions for generalizability, regularization strategies to improve robustness, and empirical results from computer vision are discussed. Suggestive results indicate that, in the case of ImageNet, popular models exhibit substantial amounts of overfitting, with training accuracy 20 percentage points or more higher than accuracy on a separately provided test sample of images.

Statistical views of (2) distribution generalization are reviewed, including the importance of datasets that make it possible to identify causal relationships and counterfactuals. An empirical example from computer vision is introduced of a lion pictured in an urban environment. Some established neural networks trained on ImageNet incorrectly classify the animal as a dog, highlighting the extent to which network-based training relies in some cases on \emph{confounding variables}, such as the surrounding environment in a photograph. These variables, which may be deep or unobserved, are correlated with target classes in training, and models in some cases learn unstable relationships driven by those predictors that fail to generalize across populations and to counterfactual cases.

Influential frameworks on (3) domain generalization are summarized, including concept drift as well as transfer learning to adapt to a shift in domain. The Photo-Art-Cartoon-Sketch (PACS) dataset is presented as a motivating example. Recent progress is discussed as are a variety of different domain generalization datasets that are used as benchmarks in the literature. The importance of the transfer of learned structures and feature embeddings is highlighted as a key strategy for generalization across domains and higher levels of abstraction.

Recent (4) task generalization breakthroughs are discussed in the areas of few-shot meta-learning---in which the skill of object identification is expanded by introducing new object classes to identify, the multipurpose BERT approach to NLP, and the growing use of transformer-based foundation models that can be applied to many tasks with minimal retraining.

Generalization across (5) modalities remains a relatively unexplored area, but recent innovations include the use and rising popularity of text-based embeddings in image classification and the transfer of a biologically-inspired network across olfactory, visual, and auditory modalities.

Exciting developments in (6) scope generalization involve supplementing deep NLP approaches with embeddings based upon graphical representations of symbolic and semantic knowledge. State-of-the-art performance from the SemEval competition is discussed. Additionally, attribution tools for extracting knowledge from deep learning models are illustrated using examples from NLP and computer vision.

The recent rapid advances that have been seen in different forms of generalization---particularly across domains and tasks---help to underscore the considerable power that deep, network-based predictive systems have demonstrated. Domain generalization, few-shot learning, and the emergence of versatile transformer based foundation models in NLP are possible in part because artificial neural networks learn quickly, generalize well both out-of-sample and out-of-distribution, and perform effectively at a wide range of discriminative and generative tasks. The speed with which these systems have been emerging and improving highights the many directions in which further developments can be expected in the coming years.

Even with such fast and consistent evolution across so many dimensions, the suggestive results introduced in this article on sample and distribution generalization draw attention to some potential directions for future research in which there may be some room for improvement. When applying well-known neural networks to classify ImageNet images, a consistent gap is apparent between the training data and a separately provided test sample. Additionally, half of those models failed to correctly recognize a lion when it was moved from its natural habitat into an urban environment. Both of these suggestive results pertain to just one dataset and application. Given, however, the considerable degree to which diverse applications rely upon backbone networks such as these, the possibility that the stability and robustness of their forecasts could be improved upon is a topic worthy of further investigation.

In addition to these patterns and findings that concern specific ways in which models models are generalized, this article provides insights about abstraction in neural networks more generally and how biological and artificial approaches to these tasks are related. The later part of the survey explores neuroscience literature on how biological brains perform generalization and abstraction tasks, focusing on their modular structure as well as the dopamine-driven organization of learned ideas into conceptual ``chunks.''

The widespread adoption of transformer-based foundation models is one important way in which AI systems are moving in the direction of this modular approach to intelligence. Tasks like recognizing objects in images or understanding text are effectively standalone. These basic problems are effectively solved, and researchers working on specialized offshoots of that work no longer have to reinvent the wheel. Each time that marginal improvements are made to those solutions, they are swiftly deployed, and that learning is leveraged at a global scale.

As foundational models are increasingly used as modules of this form, some elements could help them to move closer to the parsimonious and efficient architecture of brains. It is desirable, for instance, to make the modules themselves highly stable and robust to generalization across many populations and domains. Training the systems with diverse sets of unusual and counterfactual cases may help in those dimensions.

Additionally, brains process information hierarchically, with simple and common cases handled immediately and complex and rare cases taking additional effort. Artificial systems could in principle adopt a similar structure, breaking up the larger models into submodules that use routing rules to handle new cases. Research in few-shot learning shows that very small submodules can produce accurate projections. Designs of this form could help to improve total system stability, by preventing aberrations from exercising outsized influence on the overall system. At the same time, having submodules that work with small numbers of rare cases would enable plasticity-like updating at a decentralized level to events such as concept drift. Such updating could use incoming data to fill in gaps in knowledge, like in biological brains' curiosity-driven learning.

The movements in AI toward multimodal analysis and generalization across modalities also point toward a convergence between artificial and biological systems. Evidence from the neuroscience literature suggests that some important forms of learning and memory are multimodal, with internal representations applying to abstract entities rather than looks, sounds, or other specific sensory indicators. Publicly available foundation models are becoming the memory modules of the world's AI systems. Systems that make greater use of multimodal inputs and recognize connections among them make help to move those memory modules closer to the concept-based thinking of animals and humans.

An additional feature of biological systems that may be used fruitfully in AI systems is perceptual magnet-style inference from limited information. Researchers have found that synthetic data such as machine-generated captions on images \citep{Vasu2024} and wins and losses from self-play of games \citep{Silver2017} can substantially improve a model's performance in real-world situations. Work in AI has approached this topic by feeding simulated data directly into learning-based systems as training data. In biological brains, similar tasks are performed automatically through the use of archetypes stored in the brain. One promising area of research could be to find ways in which artificial neural networks could make such connections immediately without those extra steps.

In the reverse direction, some of the questions that arise in deep learning suggest possible avenues for neuroscience research. While brains are known to exercise selective, curiosity-driven attention, understanding the process through which new data are considered or ignored and the importance of dopamine in driving this process may help to inform biologists' understandings of neuroplasticity. Additionally, research into how brains divide into modules---which tasks are handled by which components and how each of them update according to new sensory input---could shed light on how learning occurs at different levels and how brains make the transition from the specific to the abstract. Additionally, some of the tools that deep learning researchers use for knowledge extraction and attribution have the potential to provide insights to the increasingly complex computational models that neuroscientists are developing to represent brain activity.

As the neural network community continues to explore and progress, we can expect artificial systems to continue to become more stable and robust and to achieve broader and more consistent levels of abstraction. The experiences seen in applying artificial neural networks at a spectrum of different levels of generality---and the insights that the field of neuroscience offers into analogous biological processes---both provide valuable blueprints of potential paths forward.

\bibliographystyle{cas-model2-names}
\bibliography{generality}%

\end{document}